%% file: iclr2027_conference.tex
\documentclass{article}
\usepackage{iclr2027_conference,times}

\input{math_commands.tex}

\usepackage{hyperref}
\usepackage{url}

\usepackage[utf8]{inputenc}
\usepackage[T1]{fontenc}
\usepackage{booktabs}  
\usepackage{amsfonts}
\usepackage{nicefrac}
\usepackage{microtype}
\usepackage{xcolor}
\usepackage{graphicx}
\usepackage{listings}

\usepackage{enumitem}
\usepackage{amsmath}
\usepackage{multirow}
\usepackage{float}
\usepackage{rotating}
\usepackage{amssymb, amsthm}

\theoremstyle{plain}
\newtheorem{lemma}{Lemma}
\newtheorem{proposition}[lemma]{Proposition}
\theoremstyle{definition}
\newtheorem{definition}[lemma]{Definition}
\newtheorem{example}[lemma]{Example}
\newtheorem*{axm}{Axiom}       
\newtheorem*{thmT}{Theorem}    
\theoremstyle{remark}
\newtheorem{remark}[lemma]{Remark}

\newlength{\cellw}
\newcommand{\ci}[2]{\makebox[\cellw][l]{#1\,(#2)}}
\newcommand{\dg}{\makebox[\cellw][c]{100.0$^{\dag}$}}

\title{In Search of Lost Time: Temporal Competence in Large Language Models}

\author{Ryan Liu$^{1}$, David Lee$^{1,2*}$, Icey Siyi Ai$^{3*}$, Marwa Abdulhai$^{1,4}$, \\[0.1em]
\textbf{Rachit Dubey$^{5}$, Thomas L. Griffiths$^{1,6}$}\\[0.5em]
\normalfont
$^{1}$Department of Computer Science, Princeton University\\
$^{2}$Department of Operations Research and Financial Engineering, Princeton University\\
$^{3}$Department of Industrial Engineering and Operations Research, Columbia University\\
$^{4}$Department of EECS, University of California, Berkeley\\
$^{5}$Department of Communication, University of California, Los Angeles\\
$^{6}$Department of Psychology, Princeton University\\[0.55em]
$^{*}$Equal contribution\\[0.1em]
Correspondence to: \href{mailto:ryanliu@princeton.edu}{\text{ryanliu@princeton.edu}}
\vspace{-3pt}
}

\iclrfinalcopy 
\begin{document}
\maketitle
\lhead{Preprint. Under review.}

\input{Sections/0_abstract}
\input{Sections/1_introduction}
\input{Sections/2_related_work}
\input{Sections/3_exp1}
\input{Sections/4_exp2}
\input{Sections/6_exp4}
\input{Sections/5_exp3}
\input{Sections/6_discussion}

\subsubsection*{Acknowledgments}
This work was supported by ONR MURI grant N00014-24-1-2748 and the Cooperative AI PhD Fellowship. The authors would like to thank Rafael M. Batista for his insights and feedback. 

\subsection*{AI use statement}

In this work, we used generative AI tools for assistance with:

\textbf{Experiments.} Writing experimental code, running the code, sanity checking the experimental design and setup, finding datasets, suggesting and implementing some analyses, running initial analyses.

\textbf{Writing.} Writing initial drafts for portions of the paper, shortening text, creating figures, creating tables, and generating examples.

\textbf{Ideation.} Looking for related work and deriving the axiomatic framework and proofs.

\vspace{3pt}
We have not used generative AI tools for:

\textbf{Important decision making.} Determining the research questions to study, designing experiments, editing experiment designs, editing the final draft, and numerous sanity checks. 

\textbf{Manual inspection.} The authors manually inspected prompts and experimental stimuli before running experiments. All theoretical proofs were manually verified. 

No end-to-end research agents were used---all AI interactions were back-and-forth with human oversight. All writing and experimental designs were carefully reviewed and iterated upon by the author team. The authors take sole responsibility for the final content of the work, such as text, claims, and results---including those produced with the assistance of generative AI. 


\subsection*{Reproducibility statement}

We will publish all code on a publicly accessible GitHub repository. These will include instructions to reproduce the results in the experiments in our paper. The authors will also be available via email in case additional details are required to reproduce findings. 


\bibliography{bibtex}
\bibliographystyle{iclr2027_conference}

\appendix
\input{Appendices/exp1}
\input{Appendices/exp2}
\input{Appendices/exp4}

\input{Appendices/exp3}

\end{document}

%% file: math_commands.tex
\usepackage{amsmath,amsfonts,bm}

\def\eqref#1{equation~\ref{#1}}

\def\1{\bm{1}}

\DeclareMathAlphabet{\mathsfit}{\encodingdefault}{\sfdefault}{m}{sl}
\SetMathAlphabet{\mathsfit}{bold}{\encodingdefault}{\sfdefault}{bx}{n}



%% file: Sections/0_abstract.tex
\begin{abstract} 
Today's large language models (LLMs) increasingly interact with time: AI agents schedule tasks, coordinate with people and other AI systems, and even decide how long to think. 
Yet LLMs only ever encounter time secondhand---in descriptions rather than experience. Does this hinder their reasoning and decisions that involve time?
Our paper traces this question from a model's knowledge to its actions through four successive evaluations. 
We find that models know the rules of time, which can be read, but struggle to predict durations, which must be lived; 
while they are highly accurate (92.0--97.5\%) on a collection of questions built from the theorems of various time-related literatures, 
they underestimate their own completion times by up to 99.5\% and fail to identify which tasks will take them longer to run. 
And while they are better at estimating durations of human tasks, their duration judgments change based on how an event is worded, magnifying similar human biases by up to 1.3$\times$. 
Moving from estimation to action, we find that some LLMs disregard time costs completely when answering user queries, ignoring a 1,200-fold increase in wait time entirely.
Together, our experiments reveal new challenges posed to LLMs as they navigate a dimension they cannot directly perceive, 
and highlight how their blind spots can even be simple cognitive abilities that humans take for granted. 
%
%
\end{abstract}

%% file: Sections/1_introduction.tex
\section{Introduction}


\begin{quote}
\begin{small}
In theory one is aware that the earth revolves, but in practice one does not perceive it, the ground upon which one treads seems not to move, and one can live undisturbed. So it is with Time in one's life. And to make its flight perceptible novelists are obliged, by wildly accelerating the beat of the pendulum, to transport the reader in a couple of minutes over ten, or twenty, or even thirty years. 
\begin{flushright}  
\vspace{-14pt}
-- Marcel Proust, {\em In Search of Lost Time}
\end{flushright}
\end{small}
\end{quote}

Time is a ubiquitous aspect of human life. We treat it as both a resource we allocate to tasks~\citep{becker1965theory, perlow1999famine} and a shared dimension along which we coordinate our lives~\citep{zerubavel1981hidden, orlikowski2002time}. 
However, this dimension is alien to large language models (LLMs). While their training data reflects the passage of time in timestamps~\citep{dhingra-etal-2022-time} and human descriptions~\citep{zhou2020temporal, pustejovsky2003timebank}, these are all second-hand accounts. 
In practice, language models are largely stuck in turn-based mode, only acting after user inputs~\citep{ouyang2022training, thoppilan2022lamda}. 
Autonomous agent systems such as OpenClaw go beyond this paradigm, but still tie LLM activity to scheduled intervals~\citep{openclaw_heartbeat_2026, packer2023memgpt}, and even `proactive' agents are only triggered by external clocks or events~\citep{lu2025proactive, deng2025proactive}.
Between invocations, LLMs are stateless---the passage of time reaches them as just another timestamp in context. Language models thus never experience time directly, in training or deployment, leading to the question of whether they truly understand time or if their conceptualization is jagged~\citep{karpathy2024jagged, dell2023navigating, collins2025evaluating}. 

Early evidence suggests that models can struggle with time. Gemini 3, without web access before release, refused to accept it was 2025 and accused the researcher of fabricating evidence~\citep{karpathy2025gemini3}. 
Claude 3.5 Sonnet failed to stock a simulated vending machine because it mistakenly believed its orders arrived before they had~\citep{backlund2025vendingbench}. 
And coding agents are anecdotally known to produce human-scale estimates of their completion times, forecasting days or weeks for work that they finish in minutes~\citep{starkinsider2026claudecode}. 
These cases show that temporal errors are not confined to one type or setting, but instead point to a broad class of failures that merit systematic study. 

Moreover, today's AI systems need to reason competently about time. Agents are given more complex tasks with longer runtimes~\citep{kwa2025measuring, jimenez2024swebench} and are put in multi-agent paradigms that benefit from proper scheduling and time-sensitive coordination~\citep{wu2024autogen, hong2024metagpt, fourney2024magentic}. 
Leading platforms have also introduced paradigms where models cowork with humans~\citep{anthropic2026cowork, openai2026frontier, microsoft2026copilot}, requiring not just an understanding of the LLM's own task durations but also those of human collaborators. 
%
Across these domains, failures to navigate time can reduce efficiency, increase running costs, and frustrate users. 

Despite its importance, measuring whether a model understands time is not straightforward. 
Prior work at this intersection consists of different benchmarks: temporal QA~\citep{chen2021timeqa, tan2023tempreason}, common sense~\citep{zhou2019mctaco}, date arithmetic~\citep{fatemi2025TestOfTime}, and planning~\citep{zheng2024naturalplan}. 
However, it is hard to know if failure in one setting predicts failure in another, or whether two problems share a common cause.
Similarly, the study of time itself is fragmented across literatures, so there is no collection of ground truths; findings about time are scattered across knowledge representation~\citep{allen1983maintaining, allen1985commonsense}, scheduling~\citep{allahverdi2008survey}, cognitive psychology~\citep{newell1981practice, monsell2003task, boksem2008mental}, economics~\citep{arrow1962learning}, and more.
Finally, temporal failures are hard to localize. An inefficient schedule could come from misjudged durations, competing priorities, or an LLM not even knowing the rules of time; a holistic evaluation must separate these failure modes to identify the root cause.
%
%

To address these challenges, our paper builds a systematic evaluation of LLMs' temporal competence across four levels. 
First, we test models' understandings of the \textbf{qualitative rules} (Exp 1) that govern time. We build a collection of axioms synthesizing the various aforementioned literatures, unite their findings under a shared ontology, and develop a series of vignettes to test the rules of time. 
Next, we ask if LLMs are \textbf{quantitatively calibrated} (Exp 2), examining whether they are able to attach accurate time magnitudes to human and agent task durations, including their own, as well as their responses to user chats. 
We compare these outputs to real completion times and human estimates. %
Third, we ask if LLMs \textbf{replicate human temporal biases} (Exp 2 \& 3), testing for systematic deviations in their duration estimates associated with the planning fallacy or how an event is described. 
Finally, we ask how models {\bf trade-off time against accuracy in decisions} (Exp 4). Our experiments force models to make a choice between saving time and increasing accuracy, allowing us to measure how much LLMs value time relative to other goals when completing a user query. 
We do this both when models need to estimate a duration and when that duration is provided, allowing us to break down the relative effects of estimation failure and models' inherent trade-off preferences. 

%% file: Sections/2_related_work.tex
\vspace{-3pt}
\section{Related work}
\label{sec:related_work}
\vspace{-1pt}

\textbf{Time perception in humans.} Human time estimation has been studied extensively in cognitive psychology \citep{fraisse1984,zakay1997,block1997,matthews2016,eagleman2008}. A central finding is that experienced time diverges from clock time, with differences depending on how an interval is described~\citep{tonietto2019, donnelly2022, tu2014, labianca2005}, whether time is judged prospectively or retrospectively \citep{block1997,martinelli2022}, and memory or arousal \citep{poynter1989,brown2007,stetson2007}. Studies have found that people treat time as a limited computational resource \citep{inbar2011decision, lieder2020resource}, where people constantly monitor and maintain estimates of completion time \citep{buehler1994exploring, allman2014time, dubey2026aha}, and incorporate this resource when making decisions~\citep{ariely2001timely, payne1988adaptive}.  When estimating how long tasks will take, people sometimes exhibit the ``planning fallacy'': For a task, they predict its duration to be shorter than how long the task took in the past~\citep{kahneman1979intuitive, buehler1994exploring, buehler2010planning}. In Section~\ref{sec:exp2} and \ref{sec:exp4}, we investigate LLMs' estimates of durations and their biases. Then in Section~\ref{sec:exp3}, we follow the literature and incorporate time estimates into decisions. 

\textbf{Time \& language models.}
Prior work has studied whether LLMs reason about time correctly, including benchmarks for temporal ordering, arithmetic, and duration~\citep{wang2024tram, fatemi2025TestOfTime}. Papers also document blind spots from training cutoffs~\citep{wallat2024}, show that agents can ignore time between turns~\citep{cheng2026}, but also report humanlike temporal cognition emerging without explicit training~\citep{li2026}. Others have measured LLM time preferences using intertemporal choice, finding that models are less patient than humans and that their preferences shift with language and elicitation format~\citep{goli2024, ross2024, mazeika2025, rios-sialer2026}. Existing work typically evaluates one aspect of competence, but misses the dependencies and structure across aspects that characterizes temporal competence as a whole. 
In contrast, our paper builds a holistic understanding---from rules to estimates to decisions. 

\textbf{Representing time.} 
Prior work in AI has used formalizations of domains such as intuitive physics to test whether generative models hold a coherent world model~\citep{hayes1979naive, vafa2024world}. We take a similar approach, joining different literatures that each offer formulisms for how time intervals should be represented in order to test LLMs' temporal knowledge: Knowledge representation in AI provides interval algebras that fix how events can be ordered and composed~\citep{allen1983maintaining}; scheduling theory in operations research characterizes when tasks can be parallelized, sequenced, or must wait on one another~\citep{allahverdi2008survey}; cognitive psychology and economics describe how durations fall with practice~\citep{newell1981practice, arrow1962learning}; and organizational psychology describes how they rise with fatigue and recover with rest~\citep{meijman1998workload}. We unite these formalisms into a single ontology in Section~\ref{sec:exp1}.


%% file: Sections/3_exp1.tex
\section{Exp 1: Evaluating consistency with an axiomatization of time}
\label{sec:exp1}

\subsection{Axioms and theorems}
\label{sec:exp1_axioms}

A natural way to test structural knowledge in LLMs is to curate a set of rules in a domain and check whether models follow them. 
But research on time duration is scattered across many fields, so any curated list of rules must answer the question, \emph{why these rules and not others?} 
To address this, we use the qualitative reasoning method~\citep{hayes1979naive, allen1983maintaining}: we build a domain theory for task duration, featuring an ontology that defines what components are in scope (i.e., a person's state and tasks) and a set of axioms that act upon those components (i.e., how the duration of a task depends on a person's state). 
Axioms are selected from various literatures and serve as premises of the theory. We justify our theory by showing that other established principles about time can be derived from it, including the effects of practice, context switching costs, and fatigue. 
An objection to coverage should then question the ontology or axioms, which are few and inspectable, rather than an open-ended list.

\textbf{State.} 
Our central assumption is that the duration of a task depends on the state of the individual completing it. We define state to include three components: (1) the person's \emph{skill} at the task family, which affects the speed of base work; (2) a \emph{setup} register that records the most recent task family, which influences if starting a new task requires a context switch; and (3) \emph{fatigue}, which slows all work. 
We call two tasks \emph{related} if work on one can raise skill at the other, and \emph{foreign} otherwise; task families group related tasks, and tasks not sharing a family are foreign.
Our theory allows three kinds of actions to change one's state: same-family work, foreign work, and rest. 

\textbf{Axioms.} These five axioms are our chosen assumptions that define how actions can edit an individual's state, labeled A1--A5. 
We provide an intuitive example for each axiom and theorem in Appendix~\ref{app:exp1_axiom_examples}.
\begin{enumerate}[topsep=-2pt, itemsep=1pt, label=\textbf{A\arabic*.}]
\item\textbf{Learning}: Working on a task can increase skill for that task family, but cannot decrease it. Higher skill never slows the base rate of work~\citep{newell1981practice, arrow1962learning}. 
\item\textbf{Decay}: Time away from a task family can decrease its skill. The magnitude of the decay weakly increases with more time away \citep{arthur1998factors, bailey1989forgetting}. 
\item\textbf{Screening}: The duration of a task is base work plus a fatigue cost plus a switching penalty, paid only if the previous task was foreign~\citep{monsell2003task, altmann2002memory}.
\item\textbf{Loading}: Every task weakly adds fatigue~\citep{meijman1998workload, pencavel2015productivity, warm2008vigilance}. 
\item\textbf{Recovery}: Rest weakly lowers fatigue. The magnitude of the recovery is non-decreasing in the
length of rest~\citep{bechtold1984rest, meijman1998workload}.
\end{enumerate}

\textbf{Theorems.} Five principles across different literatures are entailed by the above axioms, labeled T1--T5.
\textbf{T1 (no free transfer)}: foreign work never speeds up a task~\citep[e.g.,][]{thorndike1901influence, rogers1995costs}. 
\textbf{T2 (interruption overhead)}: a task interrupted by foreign work takes longer, increasing with each additional interruption~\citep[e.g.,][]{gillie1989interruptions, altmann2002memory}.
\textbf{T3~(rest vs.~rust)}: interrupting a task with rest trades fatigue recovery against skill decay~\citep[e.g.,][]{ammons1947acquisition, adams1961second}.
\textbf{T4 (batching dominance)}: with negligible fatigue and decay, batching work (XXYY) dominates interleaving~\citep[XYXY; e.g.,][]{jersild1927mental, liaee1997families}. 
\textbf{T5 (overlap)}: filling a wait during an external process with foreign work pays off only if the time overlap exceeds the switching penalty plus the recovery the wait provides~\citep[e.g.,][]{barnes1980motion, johnson1954optimal}. 
We provide derivations of the theorems from the axioms, along with detailed literature connections in Appendix~\ref{app:proofs}. 

A key quality of the ontology is our assumption that duration depends on a person's state generates the \textit{entire} phenomena space. Without it, T1--T4 become equalities and T5 is reduced to simple arithmetic (Appendix~\ref{app:degeneracy}).

\vspace{-3pt}
\subsection{Experiment design}
\vspace{-1pt}
\label{sec:exp1_design}
We apply the axioms and theorems to test LLMs by constructing a series of vignettes, with questions that have answers \textit{entailed} by the axioms. 
Thus, our ground truths in this section do not rely on human judgments or their biases.
Since most axioms only specify relative duration, the easiest way to measure if models are aligned with them is to ask questions comparing two durations. 
To do this, we use counterfactuals: \textit{if a person had done $X'$ instead of $X$, would action $Y$ take longer or shorter?}



\textbf{Vignettes.} 
Each vignette describes a person completing a set of tasks in a specified order. 
Vignettes are generated by templates, which include a task domain and a specific aspect of an axiom or theorem to be tested. 
%
%
All vignettes are manually examined and edited to ensure quality and coherence.

%
Our vignette domains span the four top-level branches of the O*NET Generalized Work Activities taxonomy~\citep{onet_content_model, peterson2001onet}, with two domains per branch. 
Domains in the same branch are chosen to be clearly related,
making it clear that within-branch tasks are related and cross-branch tasks are foreign. 
We picked domains by screening candidates for five criteria: repeatable units, a meaningful setup, folk concepts of fatigue and rustiness, individual work, and absence of safety norms that could trigger model refusals, yielding the chosen list in Table~\ref{tab:domains}. 
\input{Tables/exp1_domains}


\textbf{Questions.}
Each vignette contains a character's plan of action (\textit{``She is going to\ldots''}), as well as relevant context.  
Questions pose a counterfactual that modifies either the action or context (\textit{``If instead she\ldots''}), and asks how the duration of the action would change. 
Questions have four answer options: 
(a)~Longer, (b)~Shorter, (c)~About the same, (d)~Could go either way. 
Each counterfactual changes the character's state in a way that meaningfully affects the action's duration, so no item has (c) as the correct answer. Ground truth (a) or (b) indicates that a directional inequality can be derived from the axioms. 
%
Option (d) is the correct answer for deliberately under-specified vignettes. 
Each question has a flipped orientation, where reality and the counterfactual antecedent are swapped, and the ground truth is flipped as well. 

\textbf{Prompts.} Vignettes are presented single-turn in individual context windows, along with the question, answer options, and response guidelines (see Appendix~\ref{app:exp1_prompts}). 
No system prompt is given. Option ordering is fixed, and a pilot with shuffled options found no meaningful differences (Appendix~\ref{app:shuffle}). 

\textbf{Models and sampling.} We test GPT-5.6~Sol, Claude Opus~5, and
Gemini 3.7~Flash at their
default reasoning levels and temperatures, with a limit of 2000 tokens for GPT and Claude and 5000 for Gemini. Each model answered each vignette 30 times, yielding 6{,}720 responses per model. 
Claude returned refusals in 14.2\% of responses, which we remove from our analyses (see Appendix~\ref{app:coverage}).

\textbf{Metrics.} We report mean accuracy over 30 runs, averaged over each axiom or theorem, with 95\% intervals from a cluster bootstrap (resampled over 10{,}000 draws). We also measure pairwise coherence, the rate that answers of flipped orientations of a vignette mirror each other (a$\leftrightarrow$b, c$\leftrightarrow$c, d$\leftrightarrow$d). 

\vspace{-1pt}
\subsection{Results}
\label{sec:exp1_results}
\vspace{-1pt}

\input{Tables/exp1_accuracies}

\textbf{Models generally succeed at axiom- and theorem-based tests.}
Claude Opus 5, GPT-5.6 Sol, and Gemini 3.7 Flash achieve high averages of 92.0\%, 97.5\%, and 96.3\% respectively across axioms and theorems (see Table~\ref{tab:family-accuracy}). All models scored 100\% on learning (A1) and decay (A2). 

\textbf{However, models also show individual differences.} Claude performed best on axioms but worst on theorems, even though the theorems can be derived from said axioms. 
In rest vs.~rust, Claude (67.1\%) is correct when the effects of either rust or rest clearly outweigh the other; but when the comparison could go either way, Claude assumes that the time increase from rust dominates the decrease from rest. 
In 27.7\% of A4 (loading) responses, Gemini states that a morning of mentally demanding work could instead prime a person to be faster on an unrelated mental task. 
No axiom or theorem was difficult for all models, and no model was uniformly best, suggesting that temporal knowledge is not acquired as a block, and matching how the theory itself builds from many literatures.

\textbf{Models are coherent across flipped questions.} As questions are generated in flipped pairs, we can measure if models' answers are stable across flips. We match every sample of a question with every sample of its flipped version, and mark a pair as coherent if the two answers are mirrored. 
Coherence is high: 95.0\% (92.1--97.4) for Claude, 95.4\% (93.0--97.3) for GPT, and 97.8\% (95.7--99.4) for Gemini. Using the same base rates for answers (a)--(d) yields random baselines of 38--45\%. Thus, errors reflect coherent but mistaken beliefs rather than noise.

%% file: Tables/exp1_domains.tex
\vspace{-0.47cm}
\begin{table}[H]
\centering
\caption{Vignette domains grouped by O*NET Generalized Work Activities branch.}
\vspace{5pt}
\small
\setlength{\tabcolsep}{4pt}
\begin{tabular}{@{}llll@{}}
\toprule
\textbf{Information Input} & \textbf{Mental Processes} & \textbf{Work Output} & \textbf{Interacting w/ Others} \\
\midrule
\vspace{1pt}
Summarize sources     & Math problems      & Brick wall   & Social media posts    \\
Fact-check claims     & Logic questions     & Brick steps  & Reply to comments      \\
\bottomrule
\end{tabular}

\label{tab:domains}
\end{table}
\vspace{-0.41cm}

%% file: Tables/exp1_accuracies.tex
\begin{table}[!t]
\centering
\small
\settowidth{\cellw}{\phantom{0}99.9\,(99.9--100.0)}
\vspace{-17pt}
\caption{Axiom and theorem accuracy across vignettes (\%), averaged over all domains and responses. 
}
\vspace{7pt}
\label{tab:family-accuracy}
\begin{tabular}{@{}l@{\hspace{1.8em}}l@{\hspace{1.4em}}l@{\hspace{1.4em}}l@{}}
\toprule
Axiom / Theorem & \phantom{00}Claude Opus~5 & \phantom{000}GPT-5.6 Sol & \phantom{0}Gemini 3.7 Flash \\
\midrule
Learning   & \dg & \dg & \dg \\
Decay      & \dg & \dg & \dg \\
Screening  & \ci{\phantom{0}99.2}{98.1--100.0} & \ci{\phantom{0}90.0}{82.0--96.1} & \ci{\phantom{0}95.2}{88.3--99.8} \\
Loading    & \ci{\phantom{0}99.3}{97.4--100.0} & \ci{\phantom{0}92.5}{86.9--97.3} & \ci{\phantom{0}72.3}{50.2--91.5} \\
Recovery   & \dg & \ci{\phantom{0}99.4}{98.3--100.0} & \dg \\
\midrule
Interruption count & \ci{\phantom{0}74.1}{59.3--87.4} & \ci{\phantom{0}97.9}{95.6--99.6} & \dg \\
Rest vs.\ rust     & \ci{\phantom{0}67.1}{54.2--79.5} & \ci{\phantom{0}99.9}{99.5--100.0} & \dg \\
Batching           & \dg                              & \ci{\phantom{0}99.8}{99.0--100.0} & \dg \\
Overlap            & \ci{\phantom{0}88.5}{80.8--95.0} & \ci{\phantom{0}98.1}{96.5--99.3} & \ci{\phantom{0}99.0}{97.4--99.9} \\
\midrule
Overall & \ci{\phantom{0}92.0}{89.7--94.2} & \ci{\phantom{0}97.5}{96.4--98.4} & \ci{\phantom{0}96.3}{93.8--98.5} \\
\bottomrule
\addlinespace[2pt]
\multicolumn{4}{@{}l@{}}{\scriptsize $^{\dag}$Items answered identically across all 30 runs; the bootstrap interval is degenerate.}
\vspace{-17pt}
\end{tabular}
\end{table}

%% file: Sections/4_exp2.tex
\vspace{-3pt}
\section{Exp 2: Can LLMs make competent estimates of time?}
\vspace{-1pt}
\label{sec:exp2}

While the previous section evaluated whether LLMs can compare task durations, competent temporal reasoning also requires reasoning about magnitudes. To test this, we ask LLMs to estimate the completion time of humans, themselves as scaffolded agents, and other LLM agents on technical agentic tasks. We also test models' abilities to estimate their own response times to everyday user queries. These settings separate LLMs' general knowledge on task durations from knowledge of particular actors' capabilities, while self-estimation tests for a LLM analogue of the planning fallacy.

\textbf{Setup.} 
We use software and research tasks from the Human-Calibrated Autonomy Software Tasks (HCAST) benchmark developed by Model Evaluation and Threat Research~\citep[METR;][]{rein2025hcast, kwa2025measuring}. 
Of these, 16 have publicly released completion times for humans and various LLM actors, which we use as our ground truth. 
Given the complete task instructions and a description of the target actor, models provide a point estimate in minutes and 50\%, 80\%, and 95\% uncertainty intervals for completion time conditioned on task success. 
For each task, actor, and LLM conducting estimations, we elicit estimates and intervals 30 independent times.

To also test LLMs' time estimation abilities in ordinary user interactions, we take 500 user requests from WildChat-1M~\citep{zhao2024wildchat} and compare models' estimates of their own response times against real runtimes. User requests are filtered and then sampled uniformly randomly (see Appendix~\ref{app:wildchat-sampling}). For each request, we prompt Claude Opus 5, GPT-5.6 Sol, and Gemini 3.7 Flash to (1)~predict how long it would take to answer the request, including round-trip API latency, and (2)~respond to the original request verbatim, for which we record the observed response time. Prompts are sent in independent context windows and repeated five times per model and request, yielding 7,500 total trials. 

\textbf{Metrics.}
METR only provides the geometric mean duration of successful human attempts.
For ease of comparison, we use this measure for all ground truth labels for each task and actor. 
Reference-containment is the proportion of tasks whose reference falls in the estimated interval. 
From LLMs, we elicit estimates of individual completion times, which have wider uncertainty intervals and thus make it easier to contain the ground truth. Therefore, our reference-containment results are an upper bound for models' true capabilities. 
We aggregate 30 repeated estimates by taking the median point estimate and median interval bounds. 
Point-estimate bias is the median estimate-to-reference ratio across tasks, with values below and above one indicating under- and over-estimation. 

In the WildChat domain, for each model $M$ and request $i$, we measure the median estimated duration $\hat t_{M,i}$ and median observed duration $t_{M,i}$ across repetitions. 
Define the error ratio $r_{M,i} = \hat t_{M,i}/t_{M,i}$, where values below and above one indicate under- and over-estimation. 
For each model $M$, we measure its \emph{overall bias} as the geometric mean of error ratios across requests,
$\exp(\frac{1}{N_M}\sum_i \log r_{M,i})$,
where $N_M$ is the number of requests with valid medians for $M$. 
We also measure \emph{overall absolute error} as the geometric mean of error ratio magnitudes, 
$\exp(\frac{1}{N_M}\sum_i \left|\log r_{M,i}\right|),$
which equals one for perfect estimates and increases with both over- and under-estimation. We report the proportions of valid requests satisfying $\tfrac{1}{2} \leq r_{M,i} \leq 2$ and $\tfrac{1}{4} \leq r_{M,i} \leq 4$, referred to as within-$2\times$ and within-$4\times$ accuracy, respectively, as well as Spearman's rank correlation between $\hat t_{M,i}$ and $t_{M,i}$.

\vspace{-3pt}
\subsection{Estimating human task duration}
\vspace{-1pt}
We first ask Claude Opus 5, GPT-5.6 Sol, and Gemini 3.7 Flash to estimate completion times and uncertainty intervals for a skilled professional 
working independently without AI assistance on HCAST tasks, matching the participants' description in the data collected by METR (full prompt provided in Appendix~\ref{app:human-estimation-prompt}). We compare these with real human completion time data for those tasks. 

\begin{figure*}[t]
    \centering
    \includegraphics[width=\textwidth, trim={0 1.15cm 0 1.84cm},
        clip]{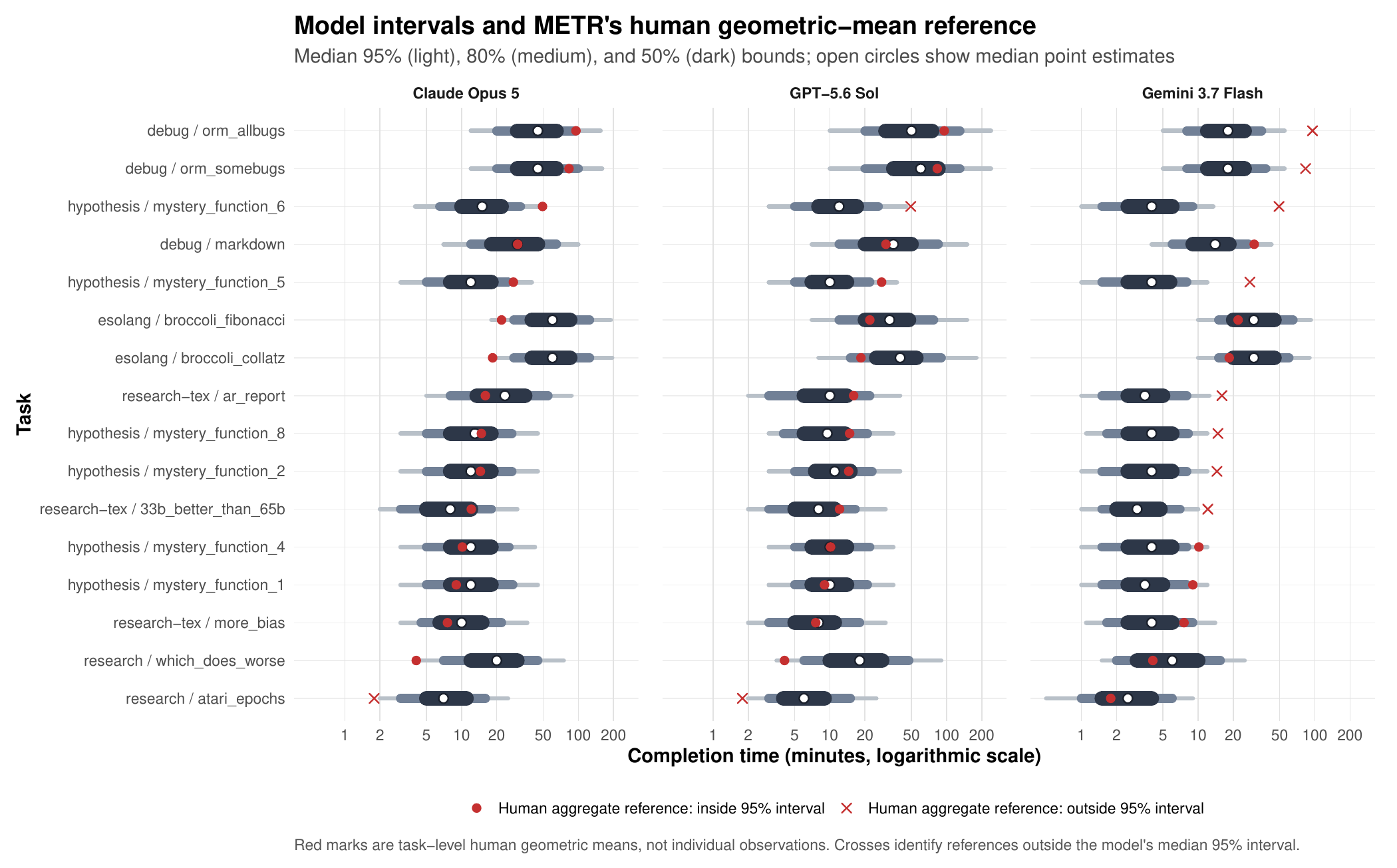}
    \caption{\textbf{LLM estimates of human task-completion times}, shown on log scale.
    White circles show median point estimates across 30 repetitions. Dark, medium, and light intervals show median 50\%, 80\%, and 95\% uncertainty bounds. Red markers show HCAST's geometric-mean human completion-times; crosses indicate references outside the model's 95\% interval.}
    \label{fig:human-time-estimation}
\end{figure*}


\textbf{Models' uncertainty intervals contain the true human-duration references to differing degrees.}
For 95\% uncertainty intervals, reference-containment is 93.8\% for Claude Opus 5, 87.5\% for GPT-5.6 Sol, and only 50.0\% for Gemini 3.7 Flash. For 50\% uncertainty intervals, containment is 43.8\% for Claude and GPT, but only 18.8\% for Gemini~(Figure~\ref{fig:human-time-estimation}; full results in Appendix Table~\ref{tab:human-reference-containment}). 

\textbf{Models moderately track differences in human task duration but vary in overall bias.}
Spearman's rank correlations between estimated and real human durations are $\rho=0.71$ for GPT, $0.68$ for Claude, and $0.58$ for Gemini (Figure~\ref{fig:human-time-estimation}, red vs.~white dots). At the median, Claude overestimates completion time by 9\%, while GPT and Gemini underestimate it by 13\% and 67\%, respectively. 

\textbf{Models underestimate the variation in human task durations.}
The standard deviation of log human completion-time references is 1.03, compared with 0.71–0.86 for model estimates. Models tend to overestimate the shortest tasks and underestimate the longest ones (Figure~\ref{fig:human-time-estimation}). 

\vspace{-3pt}
\subsection{Estimating their own task duration}
\vspace{-1pt}

Next, we have models estimate their own agentic runtimes on HCAST tasks, and check whether models exhibit an analogue of the planning fallacy bias. 
We use the newest models from each family with available METR runtimes: Claude Opus 4.6, GPT-5.2, and Gemini 3 Pro. The first two estimate their own configurations directly; because Gemini 3 Pro is no longer available, we use Gemini 3.1 Pro Preview as a proxy. 
Complete prompts and agent configurations are provided in Appendix~\ref{app:self-estimation-prompts}. 

\begin{figure*}[t]
    \centering
    \includegraphics[width=\textwidth, trim={0 1.15cm 1.2cm 1.84cm},
        clip]{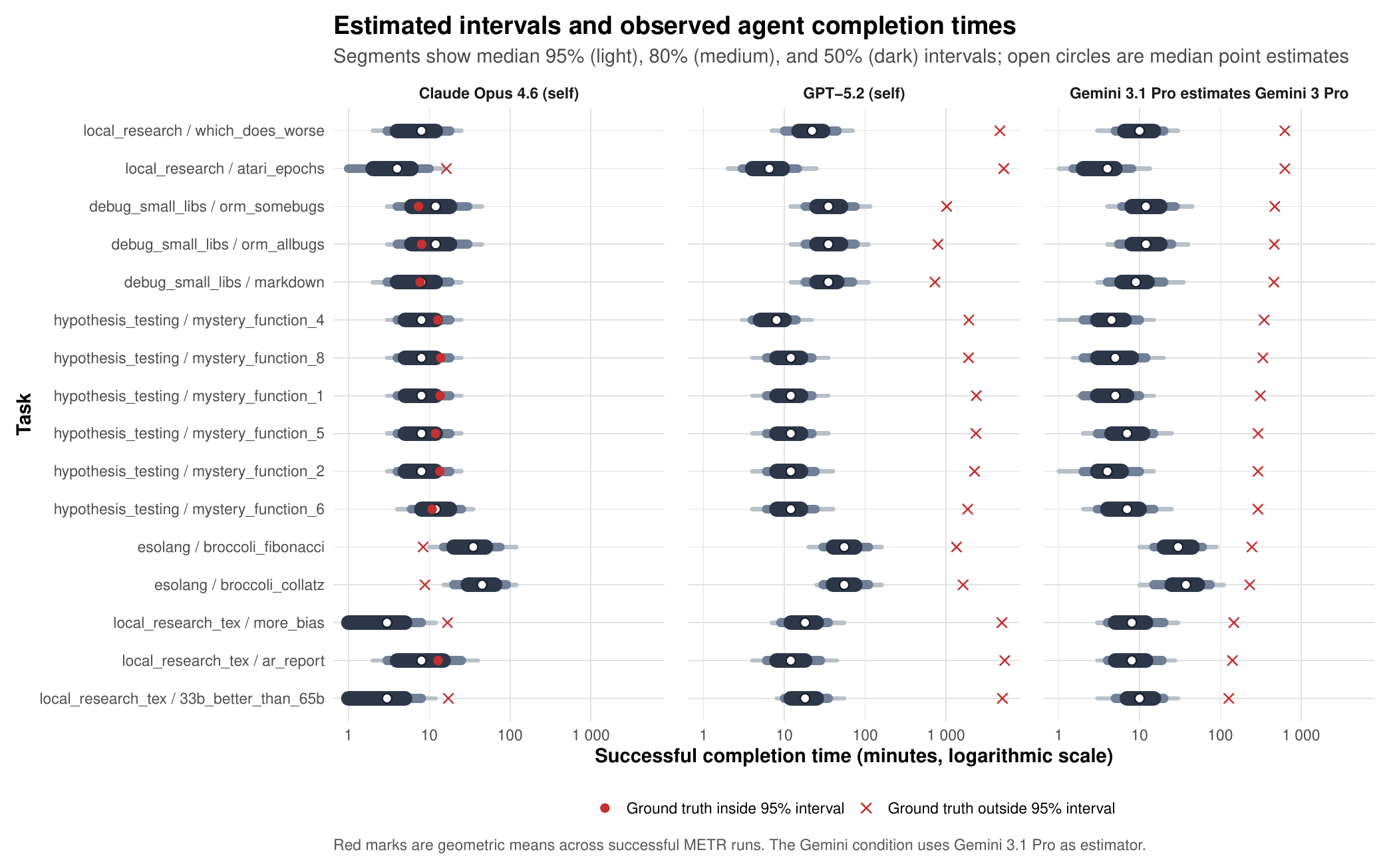}
    \vspace{-16pt}
    \caption{\textbf{Model estimates of their own agent completion times.}
    Markers and intervals follow Figure~\ref{fig:human-time-estimation}. Red markers show geometric-mean completion times across successful METR runs.}
    \label{fig:self-time-estimation}
    \vspace{-7pt}
\end{figure*}

\textbf{Models substantially underestimate the duration of their scaffolded execution.} At the median, Claude Opus 4.6, GPT-5.2, and the Gemini proxy underestimate their observed durations by 37\%, 99.5\%, and 97.6\%, respectively (Figure~\ref{fig:self-time-estimation}), 
consistent with a model analogue of the planning fallacy.

\textbf{Self-estimates also fail to track which tasks require longer agent runs.} Spearman rank correlations between estimated and observed durations are negative for Claude Opus 4.6 ($\rho=-0.81$), GPT-5.2 ($\rho=-0.59$), and Gemini ($\rho=-0.18$). Thus, errors cannot be explained by a constant scaling bias alone: models also fail to anticipate how their execution time will vary across tasks.

\textbf{Wider uncertainty intervals don't account for these errors.} Even at the 95\% level, GPT-5.2 and Gemini contain no ground truth times, while Claude contains 10/15 (full results in Appendix Table~\ref{tab:self-reference-containment}).

\vspace{-3pt}
\subsection{Estimating other LLM agents}
\vspace{-1pt}
Third, we evaluate whether LLMs can estimate the completion times of other agents, an increasingly important capability in multi-agent paradigms. 
We let Claude Opus 5, GPT-5.6 Sol, and Gemini 3.7 Flash estimate the runtimes of all 20 agents from the METR dataset, which use models ranging from GPT-4 to Claude Opus 4.6. 
Each estimator model receives the agent's model identity, available reasoning-configuration information, and scaffold description. Only agent-task pairs with at least one successful run are tested. 
By testing a range of agents with different capabilities, we check whether models distinguish among agent configurations or instead apply a generic prior over agent runtime. 
Full details and prompts are in Appendix~\ref{app:cross-model-prompt}.

\textbf{Models also greatly underestimate the task durations of other agents.}
We find that across 269 agent-task pairs, median error ratios are 0.111 for Claude, 0.075 for GPT, and 0.020 for Gemini. 
This means that observed completion times were approximately $9\times$, $13\times$, and $49\times$ longer than the estimates provided by the three models. 
Even models' 95\% intervals only contained the reference in 39.0\%, 33.5\%, and 14.1\% of cases. 
The most severe underestimation occurs for o1, o3, GPT-5, GPT-5.1-Codex-Max, GPT-5.2, GPT-5.3-Codex, and Gemini 3 Pro: for every estimator, the median ratio is below 0.03 and the median 95\% intervals contain none of the references.
Complete results are in Appendix~\ref{app:cross-agent-results} and Appendix Table~\ref{tab:agent-reference-coverage}.

\subsection{Estimating response times in user interactions}
Lastly, we check whether models' underestimation of task times extends to daily user interactions. We test this using a random sample of 500 user queries from WildChat-1M~\citep{zhao2024wildchat}. Models are provided with a user query and asked to estimate the time it would take to respond to the user, including network latency. These values are then compared with actual model response runtimes. 

\textbf{All three models underestimate their response latency on average, some models more than others.}
The overall bias is 0.94 for Claude Opus 5, 0.70 for GPT-5.6 Sol, and 0.50 for Gemini 3.7 Flash, where values closer to 1 are better. However, not every request is underestimated: Claude, in particular, overestimates some requests while underestimating others.

\textbf{GPT provides the most accurate point estimates, while GPT and Gemini best preserve the ordering of request durations.}
Overall absolute errors are $1.61\times$ for GPT, $1.96\times$ for Claude, and $2.04\times$ for Gemini; 73.4\%, 52.5\%, and 46.2\% of their estimates are within-$2\times$ accuracy. Spearman correlation between the estimates and ground truth are $0.86$ for GPT, $0.82$ for Gemini, and $0.61$ for Claude.
Complete prompts and results are provided in Appendices~\ref{app:wildchat-estimation-prompt} and~\ref{app:wildchat-results}.


Across these experiments, underestimation is most pronounced when models predict the runtime of scaffolded LLM agents, including themselves.
The WildChat results show that underestimation also occurs for ordinary API responses, but with substantially smaller effect sizes. Our results suggest that models may distinguish relatively fast from slow requests without reliably predicting their absolute duration, limiting their usefulness for scheduling, coordination, and other time-sensitive applications. 

%% file: Sections/6_exp4.tex
\vspace{-3pt}
\section{Exp 3: Do LLMs judge time durations in the same way as people?}
\vspace{-1pt}
\label{sec:exp4}

Previous experiments asked whether LLMs order and estimate durations accurately. However, exact measures of time are often different from human judgments: in addition to the event itself, humans are also influenced by how an event is described. 
In our third experiment, we measure whether LLMs' temporal estimates are aligned with people, which could help facilitate human-AI collaboration.

\textbf{Setup.} 
We compare models' and people's judgments on different sentences describing the same event, and measure how sentence features change these judgments.
We use the UDS-T dataset~\citep{vashishtha2019finegrained}, in which 765 crowdworkers assigned 163,338 duration labels to 32,302 events in English web text. 
Annotators read two contiguous sentences with an event word highlighted and chose its likely duration from an 11-point ordinal scale with a confidence rating. We sample 5K events uniformly at random from the 25,192 events with at least three human labels (Appendix~\ref{app:exp3_sampling}). 

We also test models using the 20 stimulus pairs available in~\citet{wittenberg2017quick}, which asks participants to estimate durations of events in sentences, where the same event is written using either a simple verb or a light-verb construction, to observe how phrasing changes the judgment of time. 
Stimuli are separated into three event classes: punctive count (e.g., ``kiss''), durative count (e.g., ``lecture''), and durative mass (e.g., ``advise''), upon which human judgment biases differ. 

\textbf{Models \& metrics.} We test Claude Opus~5, GPT-5.6~Sol, Gemini~3.7~Flash, and Qwen2.5-32B-Instruct, presenting each event as in the human study and receiving back either a value on the 11-point scale or a time judgment from models. 
Prompts are provided in Appendix~\ref{app:duration-prompts}. 
For API models, we repeat queries five times on a random subset of 1000 events to check the variance of individual responses. We also use this subset to check variation across three prompt rephrasings (Appendix~\ref{app:exp3_robustness}). 

We report Spearman correlation between each model's outputs and mean human labels, and compare how often two runs of a model or two humans chose the same point value for each question. 
To measure what drives each judgment, we fit the same regression to human and model labels with ten sentence features (five of the predicate and five of its context; Appendix~\ref{app:duration-detail}). 
Let $\beta^{H}_k$ and $\beta^{M}_k$ denote the effects of feature $k$ on human and model judgments. The \emph{sensitivity gain} $g = \frac{\sum_k \beta^{H}_k\,\beta^{M}_k}{\sum_k (\beta^{H}_k)^2}$ measures how strongly the model responds to features relative to humans, computed as the least-squares slope of model effects on human effects, where $g=1$ indicates human-level. We report the sensitivity gain over all features and by feature group, with 95\% bootstrapped intervals.

\vspace{-3pt}
\subsection{Results}
\vspace{-1pt}
\label{sec:exp4_results}

\textbf{Models agree with human consensus more closely than individual annotators.} Claude Opus~5, GPT-5.6 Sol, and Gemini 3.7 Flash reach Spearman correlations of 0.62--0.69 with the mean human label, against 0.49 for a single held-out annotator; Qwen2.5-32B falls below this reference (0.43). Claude and Gemini rate events about half a scale step longer than human annotators, while Qwen2.5-32B rates events to be a full step shorter (Appendix Table~\ref{tab:exp4-duration}).

\begin{figure*}[t]
\centering
\vspace{-0.5cm}
\includegraphics[width=0.65\textwidth]{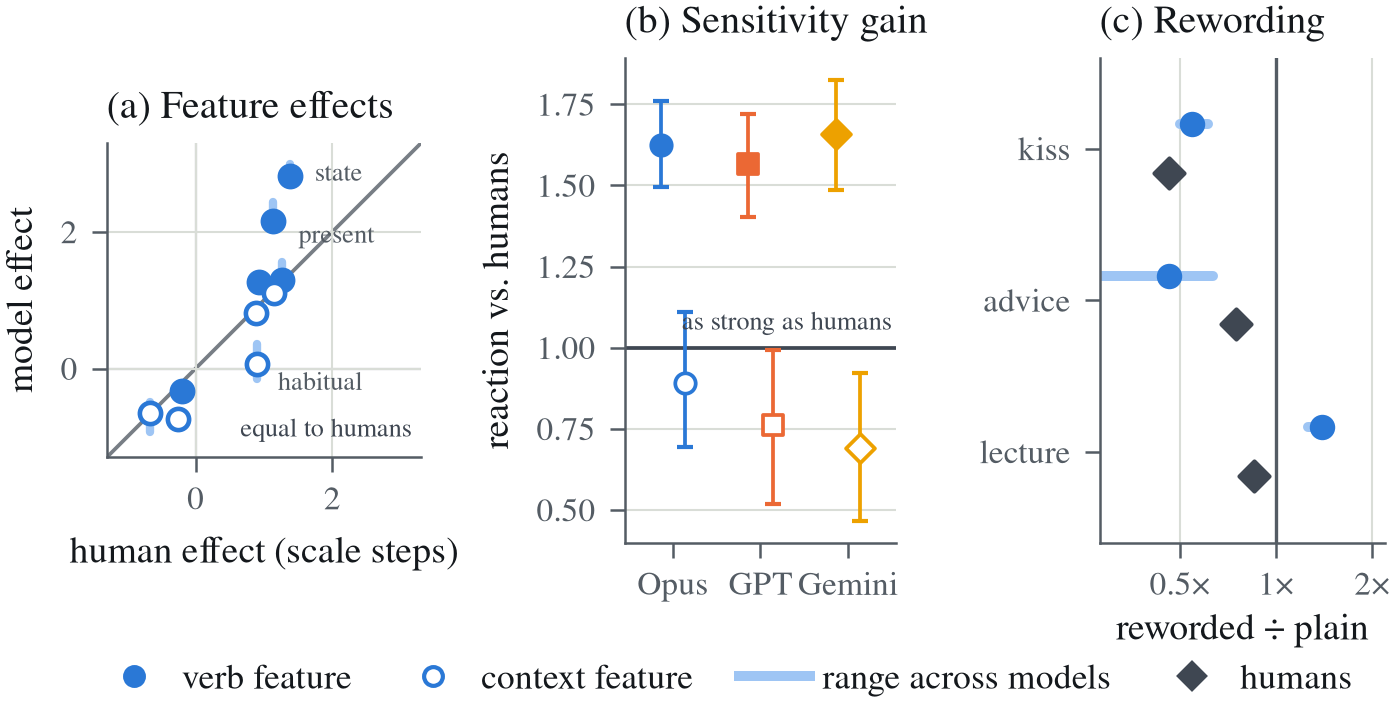}
\vspace{-8pt}
\caption{\textbf{LLMs' duration judgments react to wording differently from humans.} They are more sensitive to verb features and less sensitive to context features. (a) Human vs model effects per sentence feature; the diagonal is human-level sensitivity. (b) Sensitivity gain by feature group, with 95\% intervals. (c) The ratio of light-verb over simple-verb duration estimates for \citet{wittenberg2017quick} stimuli, one example per event class.}
\label{fig:exp4-duration}
\vspace{-0.5cm}
\end{figure*}

\textbf{Models react to wording more strongly than people, and weight wording differently.} Pooled over the ten features, the API models' sensitivity gain is 1.27--1.35, with every 95\% CI above 1, indicating a stronger reaction to how the event is described. This is caused by the features of the predicate, which moves LLMs' judgments 1.6--$1.7\times$ as much as it moves people, while context features only move them 0.7--$0.9\times$ as much as people (Figure~\ref{fig:exp4-duration}b). The pattern is stable across five runs and three prompt phrasings, and recurs on the minimal pairs of \citet{wittenberg2017quick}. However, on these pairs models reverse the human effect for durative count events (e.g., ``lecture''; Figure~\ref{fig:exp4-duration}c). Per-feature effects, robustness and the minimal-pair results are in Appendices~\ref{app:duration-detail}--\ref{app:duration-wl}.

\textbf{Models reproduce human judgments with less variation.} Two human subjects assigned the same category to an event only 29\% of the time, whereas two runs of a model agreed 80--92\% of the time. 

Overall, models match the average person's duration estimates, but struggle to capture the variation between individuals. And while they mirror certain trends in how people judge time, they can be misaligned when it comes to superficial details such as how an event is phrased. 

%% file: Sections/5_exp3.tex
\section{Exp 4: Do LLMs make competent decisions with time?}
\label{sec:exp3}
In Experiment 2 and 3, we examined how LLMs judge durations. Next, we ask how they make decisions that contain time durations as part of a trade-off, borrowing ideas from the literatures of resource rationality~\citep{lieder2020resource, gershman2015computational} and intertemporal choice~\citep{frederick2002time, green2004discounting}. This allows us to measure how much they value time relative to other traits such as accuracy. 
We measure LLMs' choices both when they must estimate the durations themselves (Section~6.1) and when those durations are provided (Section~6.2). Comparing the two allows us to separate whether a failure originates from misestimation or decision making. 

\textbf{Setup.} To test models, we construct decisions in the domain of tool use. Models complete a task and can either use a tool to increase accuracy at the cost of time, or solve the task without it. These simple decisions are naturalistic and very common in agentic workflows, and allow us to manipulate tool times to produce salient trade-offs.
%
Tools are chosen such that they offer plausible accuracy benefits but are not necessary to complete the task. In our tasks, a model chooses between two actions:
\begin{enumerate}[label=\Alph*., topsep=1pt, itemsep=0pt]
    \item Query the tool once, and use the output to obtain a final answer. 
    \item Answer directly. This avoids the time cost of the tool, but likely reduces accuracy.
\end{enumerate}
Using prompting, we elicit each model's estimated probability of answering correctly with and without the tool, and define the perceived benefit of tool use as
$\Delta p = p_{\mathrm{tool}} - p_{\mathrm{no\text{-}tool}}$.
Models almost always expect tool use to improve accuracy, with $\Delta p > 0$ in 99.77\% of trials, $\Delta p = 0$ in 0.22\%, and $\Delta p < 0$ in 0.01\%.

Each tool use decision trades off two quantities: the perceived accuracy benefit from tool use $\Delta p$, and the extra time required for tool use $\Delta t$. We measure how these dimensions affect model choices under two conditions. 
In Section~6.1, we provide only the tool's running time $\Tilde{t}$, so the model must first implicitly estimate how long each option (tool vs.~no tool) takes in order to reason about the time difference $\Delta t$. 
In Section~6.2, we provide the time previously taken for the model to do the same question under both options, yielding $\Delta t$ via direct comparison.

\textbf{Tasks.} We use tasks from two domains: solving polynomial roots and answering time-sensitive factual questions.
Polynomials are sourced from the DeepMind Mathematics generative dataset~\citep{saxton2019analysing}. We generate 30 cubic functions equally split between easy, medium, and hard difficulties as defined by the dataset (see Appendix~\ref{app:exp3_example_tasks} for examples). 
The form of equations was chosen via a pilot such that models' estimates of $p_{\text{tool}}$ and $p_{\text{no-tool}}$ were both frequently less than 100\%. 
Models are provided with a calculator tool that evaluates numerical expressions but cannot solve equations.
%

For time-sensitive factual questions, we use the ``fast-changing'' subset of FreshQA \citep{vu-etal-2024-freshllms}, consisting of questions whose answers change frequently (from daily to yearly). We randomly sample 30 questions and update the ground truth on the same day of data collection (08/27/26; examples in Appendix~\ref{app:exp3_example_tasks}). 
Models are provided with a widely used commercial search tool, Tavily \citep{tavily}, and can either rely on their internal knowledge or query the web for current information. 

\textbf{Models and sampling.} We evaluate GPT-5.6~Sol, Claude Opus~5, and Gemini~3.7~Flash in their default and no-reasoning settings, at each model's default temperature. Since Gemini lacks a no-reasoning setting, we use minimal reasoning instead and refer to it as ``no reasoning'' for simplicity.

\textbf{Delays.} We test seven tool runtimes, $\Tilde{t} \in \{1,5,10,20,30,180,1200\}$ seconds. For Section~6.2, past completion times are provided by recording the duration of tasks in Section~6.1. For both sections' experiments, each task and delay pair is repeated $10$ times per model and reasoning setting.

\textbf{Metrics.} For each task and stated tool runtime $\Tilde{t}$, we measure tool use as the fraction of retained trials that query the tool, then average these rates equally across tasks with retained trials. Separately for each model, reasoning setting, domain, and section (6.1 or 6.2), we fit $\bar{u}(\Tilde{t}) = a + b\log_{10}\Tilde{t}$ by unweighted least squares over the seven runtime values, where $\bar{u}(\Tilde{t})$ is mean tool use expressed as a percentage. The slope $b$ measures time sensitivity in percentage points per $10\times$ increase in $\Tilde{t}$, and $R^2$ describes the fit to these seven means.
\begin{figure*}[t]
\centering
\includegraphics[width=0.9\textwidth, trim={0 0.09cm 0 0.03cm},
    clip]{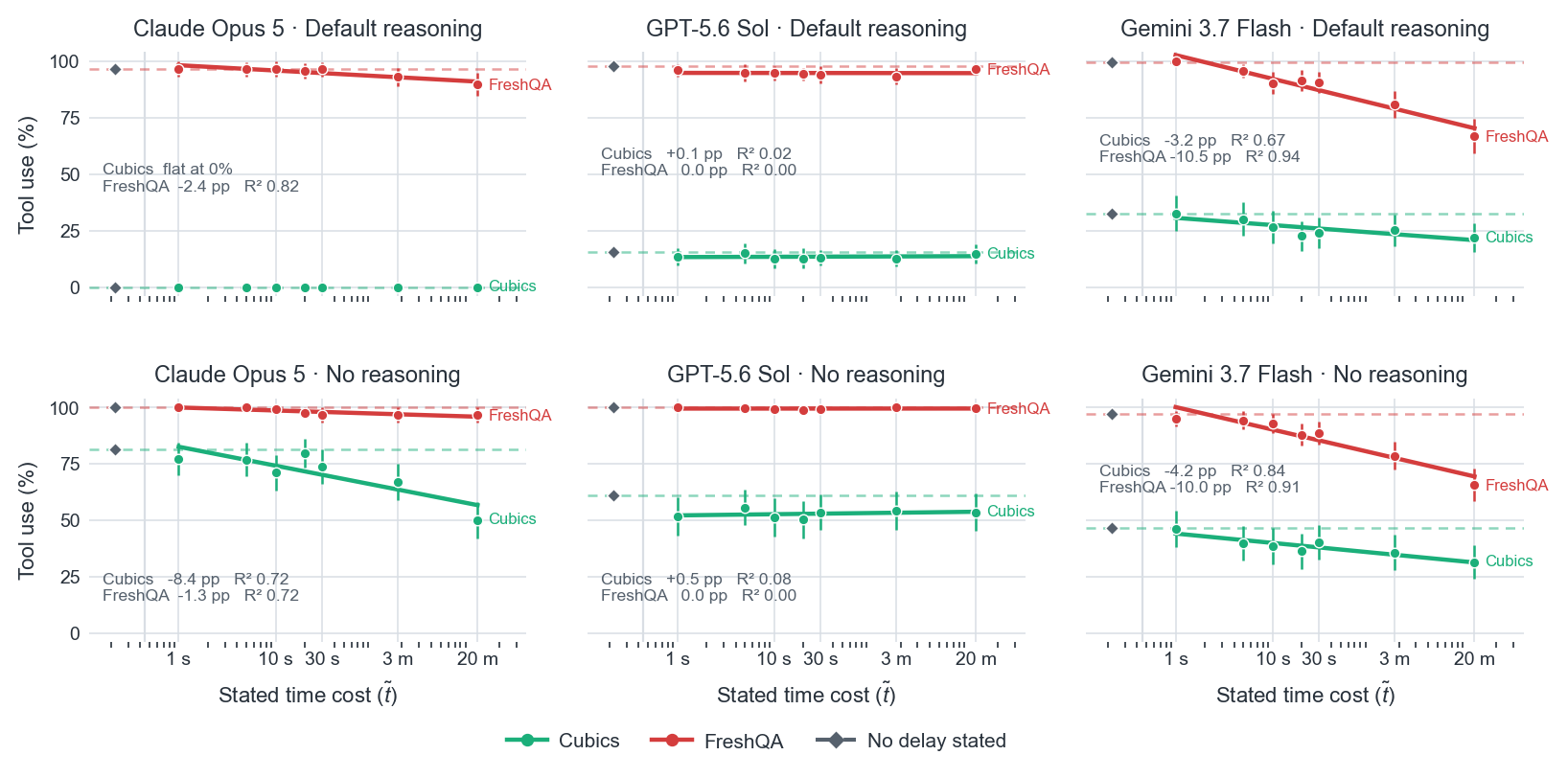}
\vspace{-10pt}
\caption{\textbf{Sensitivity to time cost varies across models.} Points indicate mean rates that the model chooses the tool across 30 tasks. Lines are least-squares fits in $\log_{10}\Tilde{t}$. Gray diamonds and dashed lines represent the base rates of tool use with no stated delay; this baseline is excluded from the fits. }
\label{fig:tool-use-delay1}
\vspace{-17pt}
\end{figure*}

\subsection{LLMs' decisions when incorporating estimation accuracy}
\vspace{-1pt}
We first study whether models decide to use a tool when only the tool runtime is stated, and completion times with and without the tool are left for the model to infer. Thus, a model's choices reflect both the accuracy of its duration estimates and the value it assigns to increased accuracy.

\textbf{Prompts.} Prior to decisions, we ask each model to estimate its probability of answering correctly with ($p_{\text{tool}}$) and without ($p_{\text{no-tool}}$) the tool. Then, in a new context window, the model gets the task, tool description, and tool runtime $\Tilde{t}$, and chooses whether to use the tool or answer directly. If the model chooses to use the tool, the tool output is returned to the model before it responds with an answer. The model is only allowed to query the tool once for each trial. See full prompts in Appendix~\ref{app:exp3_prompts}.

In addition to the seven listed tool durations, we include a baseline condition with no stated delay. 
We exclude no-reasoning trials that visibly expose reasoning in model responses (1.0\% of all trials; Appendix~\ref{app:exp4-exclusions}). 
We also investigate the quality of $\Tilde{t}$ as a proxy of $\Delta t$, and find that even with the tool runtime removed, solving without the tool takes less time than with the tool, meaning that $\Tilde{t}$ is a consistent underestimate of the true $\Delta t$~(Appendix~\ref{app:reasoning-duration}). 
We confirm models continue to misestimate their own task completion times by separately asking them to estimate how long they would take to complete each task with and without the tool (Prompt in Appendix~\ref{app:exp4-prompts-time}).

\subsubsection{Results}

\textbf{Models reduce tool use with increasing time costs, but this tendency varies across LLMs.} As tool time increases from 1~s to 20~min, for cubic functions Gemini reduces tool use by $10.7\%$ and $14.8\%$ with and without reasoning and Claude decreases by $27.0\%$ without reasoning and doesn't use tools with reasoning (Figure~\ref{fig:tool-use-delay1}). However, GPT has negligible time sensitivity, increasing rates by $1.3\%$ and $2.0\%$ instead.
For search, GPT changes by only $+0.3\%$ and $-0.3\%$ with and without reasoning, while Claude decreases by $7.0\%$ and $3.3\%$. Gemini is most sensitive, decreasing by 33.3\% and 29.3\%, respectively. Full results are in Appendix Table~\ref{tab:app-toolbydelay}. In the no stated delay baseline, tool use rates were not significantly different from rates in $\Tilde{t}=1$ and $\Tilde{t}=5$ seconds conditions. 

\textbf{Models forgo perceived accuracy benefits linearly with log runtime.} In the cases where models' tool use is sensitive to $\Tilde{t}$, least-squares fits show that tool usage declines approximately linearly with log time, with high $R^2$ values ranging from $0.67$--$0.94$; (Figure~\ref{fig:tool-use-delay1}). For example, per $10\times$ increase in tool runtime, Gemini tool use rates on search decline by 10.5\% and 10.0\% with and without reasoning. Within-task linear regressions yield 95\% intervals for the slope (pp) that exclude zero for all four Gemini conditions and Claude without reasoning on cubics. 

\textbf{Tool use positively correlated with expected accuracy gains.} Models used tools more frequently when they expected a larger improvement in their chances of answering correctly (Spearman $\rho = 0.31$--$0.91$ across models and reasoning levels; Appendix~\ref{app:benefit-choice}). Descriptive fits also suggest an accuracy--time trade-off in some conditions. For longer tool delays, a larger perceived accuracy gain yields a comparable rate of tool use to decisions where delays are shorter, consistent with said trade-off.

\textbf{Models generally overestimate their completion times without tools.} This was especially the case on cubics, with median overestimates of 5--12$\times$ for GPT, 1.5--4$\times$ for Gemini, and 12--43$\times$ for Claude, exceeding 100$\times$ on a tenth of Claude's no-reasoning questions (Appendix~\ref{app:exp4-time-estimates}). 
Their estimates for runtimes with tool use are generally more accurate. However, this leads to a misconception that for short tool times completion with the tool is faster (59--99\% of cubic estimates at $\Tilde{t} = 1$~s, whereas actual tool use completion times are faster in at most 1 of 30 questions per condition). 
As $\tilde{t}$ grows, the tool runtime outweighs the apparent time saving, and by $\tilde{t} = 3$~min almost no model estimates the tool to be quicker~(Appendix Table~\ref{tab:believed-faster}).
While these misestimations are directionally opposite from Section~\ref{sec:exp2}, they match the pattern that models fail at estimating their own completion times. 

\subsection{Provided Completion Times}
\vspace{-1pt}

In Section~6.1, a model's choice reflects both how long it expects each action to take and how it weighs time against accuracy. Given the estimation errors documented in Section~4 and in Section~6.1, the observed trends are difficult to attribute to either component alone. To address this, we localize models' decision making by removing the reliance on estimation. Specifically, we repeat the experiment while explicitly providing completion times for both actions, taken from each model's past attempts. 

\textbf{Prompts.} 
We use the same decision prompt as in Section~6.1, but the model is told how long both options took in its own past attempts at the same question (See prompt in Appendix~\ref{app:exp4-prompts-provided}). Each duration provided is the median time taken of the model's calls on that option in Section~6.1's trials. If the model never took an option on a question, we get the duration of a separate forced run instead (Appendix~\ref{app:exp4-provided-times}). 
Since real tool durations do not match the fixed tool delay $\Tilde{t}$, we adjust for this difference in the provided with-tool completion times, as if the tool took duration $\Tilde{t}$ to run. 
As in Section~6.1, we exclude trials whose output visibly exposes reasoning (0.5\% of trials; Appendix~\ref{app:exp4-exclusions}).

\begin{figure*}[t]
\centering
\includegraphics[width=0.9\textwidth, trim={0 0.09cm 0 0.03cm},
    clip]{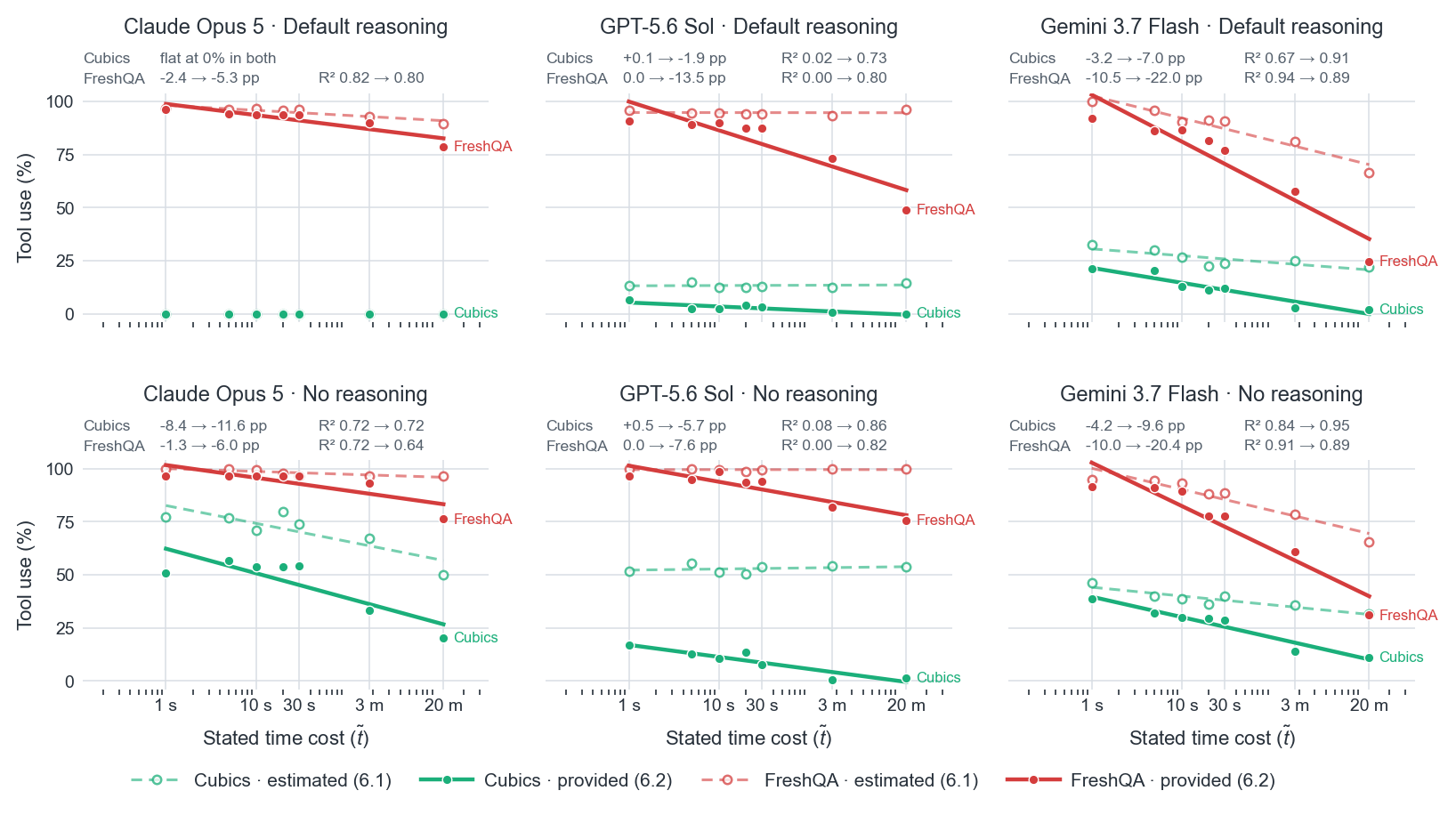}
\vspace{-8pt}
\caption{\textbf{Across settings and models, sensitivity to time cost increases when full completion durations are provided.} Points show mean tool use across 30 tasks. Lines are least-squares fits in $\log_{10}\Tilde{t}$. Hollow points and dashed lines mark durations estimated by the model (Section~6.1); filled points and solid lines mark true durations provided in the prompt (Section~6.2).}
\label{fig:tool-use-delay2}
\vspace{-7pt}
\end{figure*}

\subsubsection{Results}
\textbf{Models respond more strongly to delay when completion times are provided.} When told how long each action takes, all three models reduce tool use more steeply as $\tilde{t}$ grows (Figure~\ref{fig:tool-use-delay2}). 
GPT changes from nearly flat lines to negative slopes between $-1.9\%$ and $-13.5\%$ per $10\times$ increase. The largest change for GPT is on search with reasoning, where its tool use at $\Tilde{t}=20$~min falls from 96\% to 49\%. Gemini's slopes roughly double ($-10.5\% \rightarrow -22.0\%$ on search with reasoning), and Claude's steepen on search ($-2.4\% \rightarrow -5.3\%$ and $-1.3\% \rightarrow -6.0\%$ with and without reasoning) and on cubics without reasoning ($-8.4\% \rightarrow -11.6\%$).

\textbf{This change is especially pronounced on difficult math problems at long delays.}
On hard cubic function tasks at 20~min, tool use drops from $91\% \rightarrow 4\%$ for GPT without reasoning, and from $56\% \rightarrow 6\%$ for Gemini with reasoning and $78\% \rightarrow 31\%$ without (Appendix Figure~\ref{fig:difficulty-provided}). 
%

\textbf{Tool use remains positively associated with elicited accuracy gains.} Across questions, models continue to use tools more frequently when their accuracy estimates indicate a larger benefit (Spearman $\rho = 0.61$--$0.83$ across models and reasoning levels; Appendix Table~\ref{tab:results-provided}). Likewise, Figure~\ref{fig:tool-use-delay2} exhibits the same approximately linear relationship with log time. Descriptive fits further suggest that at longer stated delays, models require a larger perceived accuracy gain before choosing the tool, with this trade-off being more pronounced than in Section~6.1 (Appendix~\ref{app:exp4-provided-benefit}). With durations provided, models no longer overestimate their completion times without tools, and we see a corresponding increase in sensitivity to tool times relative to perceived accuracy benefits.

%% file: Sections/6_discussion.tex
\vspace{-5pt}
\section{Discussion}
\vspace{-3pt}
\label{sec:discussion}

Time is a fundamental part of human experience, but not directly accessible to language models. This invites the question of how these models represent and reason about time. We conducted a systematic evaluation of these capabilities, beginning with qualitative principles, moving to quantitative estimation, humanlike biases, and finally assessing decision-making. We found that LLMs are mostly accurate in reasoning about basic elements of time, but falter on tasks that rely on experience: estimating how long they will take to complete tasks, and making decisions that build upon these estimations. In the remainder of the paper we consider limitations of our analyses and future directions.

\vspace{-4pt}
\subsection{Limitations and future work}
\vspace{-3pt}



While our evaluation studies different aspects of time, we do not comprehensively test all aspects~\citep[e.g., date arithmetic;][]{tan2023tempreason}. However, our four-level evaluation paradigm, starting from axioms and ending at decision making, is to our knowledge the most holistic evaluation to date. 

Another limitation is in the expressiveness of the axioms used for Exp 1. 
For instance, our current theory has a binary setup register that immediately removes setup when starting a foreign task. This goes against some findings that switching costs grow with the length of interruption~\citep{altmann2002memory, monk2008interruption}. 
However, we show that to prove T4 (batching dominance), switching costs need to reset rather than accumulate after returning to a task, justifying this choice (see Appendix~\ref{app:graded}). 
%
%
Design choices like these are imperative for concise abstractions, and at points where literatures disagree, we try to make the best judgment for the theory. Counterexamples that break our theory serve instead as the next layer upon which more detailed abstractions can be built. 

For Exp 2, the limited available METR data constrained both task coverage and task information supplied to the model. Executing agents use task resources that are non-public and thus unavailable to estimators, so some errors may reflect missing cues about task difficulty rather than temporal-estimation failures alone. To investigate this, we collect as much task and environment information as we could reconstruct from public materials, and repeat the self-estimation experiment. Underestimation either persisted or became worse, suggesting that models' failures do not stem from lack of information (see Appendix~\ref{app:self-estimation-context}). However, future work should still evaluate estimates within complete agent environments when possible, and also cover broader task domains.

For Exp 3, we compare how LLMs estimate time versus humans on written descriptions of events. However, most existing human studies of duration judgment use video or other perceptual stimuli, and these cannot be directly extended to LLMs because models process other modalities differently than people do (e.g., as sampled frames rather than continuous experience). Hence, we restrict our analysis of Exp 3 to text, where humans and models receive the same input. Our results show how the description of an event affects models' duration judgments, but not whether models share human biases when judging durations they observe.

For Exp 4, we only studied one type of decision, whether to query a tool with a stated delay, across two task domains. While our design isolates an accuracy--time trade-off, real-world agents often face a wider range of time-sensitive decisions, including how long to reason before answering, how to schedule and batch subtasks, whether to wait for a slow external process or pursue alternatives, and when to provide updates to users. An important direction for future work is to examine whether models apply the temporal principles they correctly identify in Exp 1 to these more complex decisions.


Beyond the types of decisions agents face, an interesting next question is how human preferences can influence models' priorities between time, accuracy, and other desiderata. Due to the multifaceted nature of human preferences, we view the question of how an LLM adapts and aligns to users' time preferences as a valuable follow-up direction. 



%% file: Appendices/exp1.tex
\section{Details for Experiment 1: evaluating models' structural knowledge of time}
\input{Appendices/exp1/derivations}
\input{Appendices/exp1/axiom_examples}
\input{Appendices/exp1/names_exp1}
%
\input{Appendices/exp1/prompts_exp1}
%
\input{Appendices/exp1/order_shuffle_exp1}
%
\input{Appendices/exp1/token_limits_and_refusals}

%% file: Appendices/exp1/derivations.tex
\providecommand{\States}{\mathcal{S}}
\providecommand{\Acts}{\mathcal{A}}
\providecommand{\Fams}{\mathcal{F}}
\providecommand{\del}{\delta}
\providecommand{\dur}{\tau}
\providecommand{\Tt}{T}
\providecommand{\famof}{\mathrm{fam}}
\providecommand{\setup}{\sigma}
\providecommand{\skill}{\kappa}
\providecommand{\load}{\phi}
\providecommand{\pen}{\pi}
\providecommand{\mspan}{M}
\providecommand{\seq}{\mathbin{;}}
\providecommand{\indic}[1]{\ensuremath{\mathbf{1}\!\left[#1\right]}}

\subsection{Axioms, Theorems, and their derivations}
\label{app:derivations}

In this section, we provide the formal counterpart of the state model in Section~\ref{sec:exp1_axioms}. This covers axioms A1--A5, theorems T1--T5, and how the theorems can be proven from said axioms. 
In the main text, each axiom/theorem is stated as a directional claim, which we formalize as an inequality over a small transition system. We also show why the requirements within vignettes exist in order to make their ground truths derivable, and provide a countermodel showing the structural assumption that batching dominance relies upon.

\subsubsection{Formal setup}
\label{app:setting}

\paragraph{Actions, families, and states.}
Let $\Acts$ be a set of task types (\emph{actions}) and $\Fams$ a set of skill \emph{families}, with the labeling $\famof : \Acts \to \Fams$ fixed by
each vignette. 
A state is $s = (w, \skill, \setup, \load)$: a world $w$; a skill profile of the character $\skill : \Fams \to
\mathbb{R}_{\ge 0}$; a setup register $\setup \in \Fams \cup
\{\bot\}$ containing the family of the most recent task (initially $\bot$);
and a fatigue level $\load \ge 0$. Two primitive functions act upon
the theory: a transition function $\del : \Acts \times \States \to \States$ and a duration function $\dur : \Acts \times \States \to \mathbb{R}_{\ge 0}$. We denote $\skill_F(s)$ as the character's skill at family $F$ in state $s$.

\begin{definition}[Related and foreign work]
\label{def:foreign}
In our theory, work on task $a$ can raise one's skill at task $b$, i.e., $\skill_{\famof(b)}(\del(a, s)) >
\skill_{\famof(b)}(s)$ for some $s$. Two tasks are \emph{related} if
either task can raise skill at the other task, and \emph{foreign} otherwise.
Skill families group mutually related tasks, where work on each task within the family can raise skill in all other tasks in the family. 
Tasks not sharing a family are foreign. %
By definition, for foreign tasks $a, b$ and every $s$,
$\skill_{\famof(b)}(\del(a, s)) \le \skill_{\famof(b)}(s)$. That is, performing a foreign task can never raise skill.
\end{definition}

\paragraph{Events.} Three kinds of action can affect state: same-family
work (with respect to the previous task), foreign work, and rest. Rest is when no tasks are being conducted, and is represented by idle family $\{w_d\}_{d \ge 0}$.
 
\begin{definition}[Well-definedness, state sufficiency]
$\del$ and $\dur$ are only functions of $a$ and $s$. Past action trajectories that reach the same state are interchangeable in their future actions and durations.
\end{definition}
\begin{definition}[Existence, rest] 
Rests can be of any length: for every $d \ge 0$, there is an idle action $w_d$ with $\dur(w_d, s) = d$. 
Idling for time $d$ takes duration $d$: $\dur(w_d, s) = d$. 
Concatenating rests is the same as resting for their total duration: $w_d \seq w_e \equiv w_{d+e}$.
Rest does not change the world: $\del(w_d, \cdot)$
keeps the same $w$ and $\setup$, and its effects on $\skill$ and $\load$ are only the ones stated in A2 and A5.
\end{definition}
\begin{definition}[Existence, splits]
A non-atomic task $a$ can split into two tasks $(a_1, a_2)$. These two tasks belong to
$a$'s family, and if performed subsequently in the correct order, yield the same transition function and duration: $\del(a_1 \seq a_2, s) = \del(a, s)$ and
$\Tt(a_1 \seq a_2, s) = \dur(a, s)$. 
\end{definition}
\begin{definition}[Existence, external processes]
An external process $p$ has duration $d_p$ and does not require the agent to attend to it during its duration. 
The agent launches it ($\ell_p$) and later collects it
($c_p$). $c_p$ cannot occur unless $d_p$ has passed since $\ell_p$. 
\end{definition}

\paragraph{Composite duration.}
For a sequence of actions $\alpha = a_1 \seq \cdots \seq a_n$, the duration of the total sequence is defined recursively: $\Tt(\alpha, s) = \sum_{i} \dur\big(a_i,\; \del(a_1 \seq \cdots \seq a_{i-1}, s)\big)$. Thus, we have the following lemma: 

\begin{lemma}[Composition]
\label{lem:composition}
$\Tt(\alpha \seq \beta, s) = \Tt(\alpha, s) + \Tt(\beta, \del(\alpha, s))$.
\end{lemma}
\begin{proof}
We can prove this by inducting on the size of $\beta$, where we unfold the last action
$\del(\alpha \seq \beta', s) = \del(\beta', \del(\alpha, s))$. Following from Definition 2, past events do not affect the current state or future transitions. 
\end{proof}

\subsubsection{Axioms}
\label{app:axioms}
 
Throughout this section, we write $F = \famof(a)$ for the family of the current action.
 
\begin{axm}[A1, learning]
When performing task $a$, it is possible (but not always the case) for learning; thus the skill in that family is non-decreasing: 
$\skill_F(\del(a, s)) \ge \skill_F(s)$. Moreover, higher skill never slows base work.
\end{axm}
 
\begin{axm}[A2, decay]
Time away from skill family $F$ weakly erodes $\skill_F$, regardless of if it is filled by rest $w_d$ or foreign work $a'$: $\skill_F(\del(w_d, s)) \le \skill_F(s)$ and $\skill_F(\del(a', s)) \le \skill_F(s)$ for $\famof(a') \ne
F$. Decay is non-decreasing with duration.
\end{axm}
 
\begin{axm}[A3, screening]
The duration of a task $a$ in skill family $F$ is its base cost $b_a$ plus a fatigue term $h$ plus a switching cost $\pi_F$ if the task does not match the current setup. 
\[
\dur(a, s) \;=\; b_a\big(\skill_F(s)\big) \;+\; h\big(\load(s)\big)
\;+\; \pen_F \cdot \indic{\setup(s) \ne F},
\]
The base-work function $b_a$
is non-increasing in $\skill_F$, $h$ is non-decreasing (fatigue slows all work), and the switching cost $\pen_F > 0$~\citep{rogers1995costs, monsell2003task}. Performing task $a$ sets the setup register $\setup :=
F$.
 
\end{axm}
\begin{axm}[A4, loading]
Fatigue increases when performing all tasks. 
$\load(\del(a, s)) \ge \load(s)$ unless $a$ is idling. We denote the fatigue toll of task $a$ as
$\Delta_a(s) = \load(\del(a,s)) - \load(s) \ge 0$. 
\end{axm}
 
\begin{axm}[A5, recovery]
When idling, fatigue drops. 
$\load(\del(w_d, s)) \le \load(s)$, non-increasingly in $d$.
\end{axm}
 
\begin{remark}[Limitation: fatigue as a single resource]
Our theory treats fatigue as a single resource that every task increases. However, in reality restoration can be attained by qualitatively different work, for instance doing mental tasks to rest the physical body. We consider this outside the scope of our theory; To address this, in the vignettes we build where this difference matters, we explicitly state that 1) the character is well-rested, and 2) that tasks are not tiring for them to complete, in order to ensure that this detail does not affect the tests generated by these axioms. 
\end{remark}
 
To illustrate the coverage of our axioms over the theory space, we show that our axioms cover how each type of event \{same-family work, foreign work, rest\} affect a component of the theory \{skill, setup, fatigue\}. 
 
\begin{center}
\small
\begin{tabular}{@{}llll@{}}
\toprule
Component & Same-family work & Foreign work & Rest \\
\midrule
$\skill_F$ (skill) & never lowers (A1) & never raises (Def.~\ref{def:foreign}); erodes (A2) & erodes (A2) \\
$\setup$ (setup) & set to $F$ (A3) & set to foreign (A3) & preserved (Def.~3) \\
$\load$ (fatigue) & raises (A4) & raises (A4) & lowers (A5) \\
\bottomrule
\end{tabular}
\end{center}
 
\paragraph{What the axioms leave open.}
Commutativity is not guaranteed; our theory does not entail that $\Tt(a \seq b, s) = \Tt(b \seq a, s)$, nor does it negate it. 
All axioms except A3 are directional claims, but the magnitudes of those directions can vary. 
Because the rate of skill decay and the gains from reducing fatigue are not specified beyond monotonicity, a long break away from a task may help or hurt completion times. We cover this in Theorem T3 (rest vs.~rust).
 
\subsubsection{Special conditions for theorems T4 and T5}
\label{app:stipulations}
We use additional common assumptions, which are part of the space that our axioms deliberately leave open. 
 
\textbf{Condition} (C1, negligible fatigue)\textbf{.} All tasks raise fatigue by a negligible amount. 
 
\textbf{Condition} (C2, negligible decay)\textbf{.} Over the course of the vignette, skills decay by a negligible amount.

\textbf{For our proofs}, these conditions are used to prove Theorems T4 and T5. 

\textbf{For our vignettes}, a condition is stated if when left unspecified, under our axioms the gold answers are not the only plausible answer for the question. This is independent of whether the corresponding proof uses the condition. Both conditions are used in prompts corresponding to listings 7, 11, 12, and 13. Only condition C2 is used in listings 5 and 6.

\subsubsection{Derivations}
\label{app:proofs}
 
\begin{thmT}[\textbf{T1, no free transfer}]
\label{thm:t1}
If $a$ is foreign to $b$, then the time taken to do $b$ after $a$ is at least as long as the time it takes to do $b$ directly. 
$\dur(b, \del(a, s)) \ge \dur(b, s)$.
\end{thmT}
 
\paragraph{Connection to literature.} Theorem T1 splits into two cases, when the setup register contains a foreign family ($\setup(s) \neq \famof(b)$) and when it contains a related family ($\setup(s) = \famof(b)$). These two cases are covered by two different bodies of research. 

The first body of work corresponding to $\setup(s) \neq \famof(b)$ finds that practice at a task will provide no transfer to a dissimilar one. 
Early work over a century ago in experimental psychology found no transfer between spelling and multiplication~\citep{thorndike1901influence}.
This style of finding has continued across various skill domains up to recent years: 
a meta analysis between chess, music, and working-memory instruction have found that transfer to academic and cognitive skills rarely occurs, with a further decrease in effect sizes as experimental controls improve~\citep{sala2017far}, 
and a review of the brain-training literature found that gains appeared on trained tasks, less on closely related ones, but these gains reduced to very little on distantly related tasks~\citep{simons2016brain}.
The key takeaway here is that \textbf{performing an unrelated task does not help, but it also doesn't necessarily hurt}. 

The second body of work corresponding to $\setup(s) = \famof(b)$ is the concept of \emph{switch costs} in the task-switching literature. This is defined as the difference between switch and repeat trials, i.e., $b$ after $a$ vs.~$b$ after $b$~\citep{rogers1995costs, monsell2003task}. 
The switch cost persists as a residual cost even when the participant has had time to prepare. 
In this setting, \textbf{the switch cost actively hurts the performance duration of task $b$. }

In our proof of Theorem T1 below, we show that \textbf{our axioms entail this key difference across the two literatures above}, unifying them. Skill transfer research covers the cold-start comparison (foreign work does not help), and task switching covers the warmed-up comparison (foreign work strictly hurts), and we will demonstrate that the setup register $\setup(s)$ reproduces the boundary between them.
\vspace{-10pt}
\begin{proof}
Consider when A3 is applied to the action $b$ on both sides of the equation.
 
For the first term $b_b$ (base work), since $a$ is foreign it never raises task $b$'s skill $\skill_{\famof(b)}$ (Definition~\ref{def:foreign}).
Furthermore, performing $a$ weakly erodes $\skill_{\famof(b)}$
(A2). Then by A1, lower skill never speeds up base work (A1), so the base cost term weakly rises. 
 
For the second term $h$ (fatigue cost), performing $a$ weakly
raises load $\load$ (A4), and $h$ is non-decreasing in $\load$. 
 
Finally, consider the third term $\pen$ (switching penalty).
Performing $a$ changes the setup register to $\famof(a) \ne \famof(b)$, which either leaves $b$'s penalty status unchanged (if previously foreign) or increases it (if previously related). 

Combining the three terms, in both cases of the switching penalty, all three terms move weakly upward, so there is no free transfer. 
\end{proof}
\vspace{-6pt}
\textbf{This split exactly matches the two sets of findings from literature mentioned above.}
Specifically, when the previous task was related to $b$ ($\setup(s) = \famof(b)$), the third term strictly increases from $0$ to $\pen_{\famof(b)} > 0$ (A3), and the inequality is strict. And when $\setup(s) \neq \famof(b)$, the inequality is loose. 
\vspace{6pt}
 
\begin{thmT}[\textbf{T2, interruption overhead}]
\label{thm:t2}
Let $(a_1, a_2)$ be a split of $a$ (following Definition 4) and let $\beta$ be a non-idle foreign task. 
When $\beta$ interrupts $a$ in the form of $a_1 \seq \beta \seq a_2$, the time attributable to $a$ is greater than $\dur(a, s)$, specifically by at least $\pen_{\famof(a)}$. This scales linearly, so with $k$ foreign interruptions the cost is at least $k\cdot\pen_{\famof(a)}$.
\end{thmT}

\paragraph{Connection to literature.} 
When $a_2$ resumes after $\beta$, the setup register is at $\famof(\beta)$, so by A3, $a_2$ has an additional switching cost back to $\famof(a)$. 
Two literatures use different names for this cost: the aforementioned \emph{switch cost} in task-switching~\citep{rogers1995costs, monsell2003task}, and the \emph{resumption lag} of interruption research---which is the time from the end of an interruption to the first action of the next task, attributed to re-establishing a suspended goal~\citep{altmann2002memory}.

In some studies, the resumption lag increases with the duration of the interrupting task~\citep{monk2008interruption}, but not in others~\citep{gillie1989interruptions}. This is captured within our theory using the flat setup cost (A3) along with a scaling decay cost (A2). 
Our theory adopts the fixed cost model, while leaving any additionally accruing costs to decay (A2), which is monotone to the duration of time away from the task. 
In the literature, resumption cost also scales with the complexity of the interrupter task~\citep{gillie1989interruptions, monk2008interruption}. 
Our theory covers this via the fatigue term (A4). 

However, the literature describes the scaling as due to working-memory displacement rather than fatigue, which is a direction of refinement that we discuss in Section~\ref{app:graded}.
Another effect not reflected in our theory is how sometimes, interruptions \emph{similar} to the primary task are more disruptive~\citep{gillie1989interruptions}. Instead, a same-family task is treated as not affecting the setup at all. However, Theorem T2 does not specify anything for same-family interrupters, and our vignettes for T2 use foreign interrupters to avoid this case. 
Finally, studies have shown that subjects tend to compensate for the time lost during interruptions by working faster afterwards, at the cost of stress and effort~\citep{mark2008cost}. 
Due to the history-free property of our model, this effect is not supported by our current theory. 

\begin{proof}
Define state $s_1 = \del(a_1, s)$. 
Without interruptions, $a_2$ proceeds with $\setup(s_1) =
\famof(a)$ and pays no switching cost (A3). 

In the interruption case, $a_2$ runs in state 
$s_2 = \del(\beta, s_1)$. The last action was $\beta$, which is foreign to $a$, and thus also foreign to both $a_1$ and $a_2$ (Definition 4). The last action $\beta$ also sets the setup register to $\famof(\beta)$, and so when completing task $a_2$ we need to pay an extra $\pen_{\famof(a)}$ for the setup term (A3). 

Next, we show that with the additional task $\beta$, the other cost terms in Axiom A3 (base cost and fatigue) do not reduce. 
Base cost $b_{a_2}$ depends only on skill $\skill_F$ and performing foreign task $\beta$ cannot increase skill (Definition 1) and can weakly erode skill (A2), therefore $b_{a_2}$ weakly rises. 
Similarly, after performing $\beta$ fatigue can only be equal or higher (A4). 
Since the total time is preserved among splits, i.e., $\Tt(a_1 \seq a_2, s) = \dur(a, s)$ (Definition~4), the total increase in completion time is at least $\pen_{\famof(a)}$. 

Finally, consider the case where there are $k$ foreign interrupters. 
These interruptions can each be added one after the other from left to right, where the right portion of the previous split is then split again, and any time cost increases are charged to that split. Then, by simple induction, with $k$ total interruptions we will have at least $k\cdot\pen_{\famof(a)}$ extra time cost. 
\end{proof}

\begin{thmT}[\textbf{T3, rest versus rust}]
\label{thm:t3}
Let $(a_1, a_2)$ be a split of $a$. Consider the case where $a$ is interrupted by a rest action $w_d$.
Let the state $s'$ be after $w_d$ but before $a_2$, i.e., $s' = \del(w_d, \del(a_1, s))$. 
Then the time difference of $a_2$ compared between resting after $a_1$ and not resting, i.e., 
$\dur(a_2, s') - \dur(a_2, \del(a_1, s))$ is:
\[
\underbrace{b_{a_2}(\skill'_{\famof(a)}) - b_{a_2}(\skill_{\famof(a)})}_{\ge 0
\ \text{(decay, A2 via A1)}}
\;+\;
\underbrace{h(\load') - h(\load)}_{\le 0 \ \text{(recovery, A5)}},
\]
with both terms monotone in $d$. The sign (+/-) depends on the relation of the two terms, and is not entailed by the axioms.
\end{thmT}

\paragraph{Connection to literature.} 
T3 is the theorem where two opposing forces clash---the benefit from resting to reduce fatigue versus the decay of skill. 
The motor-learning literature has observed and named these two opposing forces, and have measured them and found them independent. 
The term governing fatigue recovery is \emph{reminiscence}: the improvement of one's performance after a rest without intervening practice~\citep{ammons1947rotary, ammons1947acquisition}. This is explained as dissipation of a fatigue-esque inhibition that accumulates under continuous work, matching A4 and A5~\citep{hull1943principles}. 
The term covering decay is the \emph{warm-up decrement}, also termed the second facet of forgetting~\citep{adams1961second}, where performance is worsened immediately after a rest but quickly recovers---an effect separate from actual learning.

These two forces are independent in the literature~\citep{adams1961second}, and one paper illustrates how outcomes of rest can be two-sided: a study in motor learning~\citep{catalano1978effect} found that when performing a continuous task and then resting, rest \emph{improved} performance, but the \emph{opposite} happened for a discrete task where fatigue is presumably lower due to time between each subtask. 
Similarly, a review of breaks in
industrial and transport domains found regular breaks effective for
maintaining performance, but that there is little evidence on the optimal length of said break~\citep{tucker2003impact}. 
Our theory's opposing forces of A2 and A5 capture the flexibility of sign in these past findings. 

We address one potential refinement in Section~\ref{app:graded}---currently, the theory assigns all rest-induced loss
to skill decay (A2). However, warm-up decrement can also be attributable to the decay of the setup under idling. Instead, in our theory has a stricter definition of rest that preserves $\setup$ (Definition~3). 
Despite this, at the sign level both count as rust, so T3 stands regardless. 
Our three vignettes for rest vs.~rust fix the ground truth sign by specifying skill and fatigue levels beforehand, where either skill decay is negligible, fatigue restoration is negligible, or both are in play; these fix the comparative relationship for the duration of remaining work.

\begin{proof}
Consider the two cases, one where $a_2$ immediately follows $a_1$, and the other where they are interrupted by rest $w_d$. 
As $w_d$ preserves the setup register as $\famof(a)$, in both cases there is no resumption penalty. 
Because rest weakly erodes skill (A2), $\skill' \le \skill$, and duration is non-increasing in skill, so the first term $b_{a_2}(\skill'_{\famof(a)}) - b_{a_2}(\skill_{\famof(a)})$ is non-negative. 

On the other hand, rest reduces fatigue (A5), so $\load' \le \load$. Since duration is non-decreasing in fatigue load, the second term $h(\load') - h(\load)$ is non-positive. As one term is non-negative and one is non-positive, but the magnitudes are not fixed, the sign is not guaranteed. 
\end{proof}

\begin{lemma}[Block decomposition]
\label{lem:blocks}
Let there be several task families that are pairwise foreign with each other. 
Each family has a sequence of tasks to be completed. 
Consider a plan $\alpha$ that interleaves task sequences from each family without rest, e.g., for two families $X$ and $Y$, $XXYY$, $XYXY$, and $XYYX$ are all such plans. 
We call a maximal run of consecutive tasks of family $F$ an \emph{$F$-block}. 
Under conditions C1 (negligible fatigue) and C2 (negligible decay), the total time of $\alpha$ is:
\begin{align*}
\Tt(\alpha, s_0) \;=\;
\underbrace{\sum_{a \in \alpha} b_a(\skill_{\famof(a)} \text{ at } a)}_{\text{order-invariant}}
\;+\;
\underbrace{|\alpha| \cdot h(\load_0)}_{\text{order-invariant}}
\\[2pt]
{}+ \sum_{F} \pen_F \cdot \#\{\text{$F$-blocks of $\alpha$ not entered
with } \setup = F\},
\end{align*}
The key point is order invariance---given identical task families and sequences, the order in which they are interleaved does not matter for the first two terms; 
interleaving affects total duration only through the number of $F$-blocks.
\end{lemma}
\begin{proof}
By A3, $\Tt(\alpha, s_0)$ expands into a sum of base-work terms, fatigue terms, and penalty terms, and group like terms. 
Because C2 (negligible decay) is active, each task's skill is only affected by practice from predecessors in its own family's sequence (A1). 
However, because each family's tasks are completed in the same order, in every interleaving the skill gain is identical, so the total sum of base work is order invariant. 
Because C1 (negligible fatigue) is also active, and rest never occurs, $\load$ stays at the initial $\load_0$ and the sum of fatigue costs is $|\alpha| \cdot
h(\load_0)$ across every interleaving. 

Finally, for the penalty sum terms, the first task in every $F$-block sets $\setup := F$, so all tasks after it keep the same $\setup$, so none of the later items in a block pay a switching cost. 
However, the first task in a block always follows a foreign task, as if it was preceded by a non-foreign task, the block would no longer be the maximal sequence of like-family actions. Thus, the first task always pays switching cost $\pen_F$. 
Across all terms, this becomes $\sum_{F} \pen_F \cdot \#\{\text{$F$-blocks of $\alpha$ not entered
with } \setup = F\}$. 
The specification of ``not entered with $\setup = F$'' only applies at the start of $\alpha$, where the first task could either benefit from a preceding task of the same family or not. 
\end{proof}
 
\begin{thmT}[\textbf{T4, batching dominance}]
\label{thm:t4}
Under conditions C1 (negligible fatigue) and C2 (negligible decay), for a pair of foreign families $X$ and $Y$, and a plan $\alpha$ without any rest actions, the
batched order (i.e., one block per family; $XXYY$) takes less time than every interleaving of the same set of tasks (e.g., $XYXY$).
\end{thmT}

\paragraph{Connection to literature.}
The core of batching dominance is the number of same family ``blocks'' of tasks that do not incur a setup penalty. 
Two independent literatures have used this quantity directly. 
First is the founding study of the task-switching paradigm performed by Jersild in 1927~\citep{jersild1927mental}. Jersild timed subjects on their completion of separate tasks one set at a time versus alternating item-by-item. 
He compared times across conditions and found that a majority of tasks took excess time when alternating, matching Theorem T4. 

The other literature is scheduling with job families from operations research (OR), where a cost is charged on each change of family~\citep{liaee1997families, monma1989complexity}. 
The \emph{group technology assumption} common in the literature is a standard restriction that requires each family's tasks to be processed as one contiguous sequence~\citep{liaee1997families, allahverdi2008survey, potts2000batching}, matching the result of the batching theorem. %
Then, each batch is an $F$-block and the literature's total setup time---the switching costs summed over batches---is the lemma's third term.

\begin{proof}
From Lemma~\ref{lem:blocks}, the ordering of the tasks only affects the number of foreign setup switches; the base and fatigue costs of the actions are otherwise identical. 
Batching has one $F$-block per family, so not counting whether the first task requires setup, there is only one switching cost. 
For all other interleavings, the minimum is that some family is split into two or more blocks, meaning that there are at least two switching costs, away from that family and then back. Because these costs are strictly positive ($\pen_F > 0$; A3), batching tasks takes strictly less time. 
\end{proof}

\begin{thmT}[\textbf{T5, overlap}]
\label{thm:t5}
Let $p$ be an external process (see Definition 5) with duration $d$, and let its collection be a task within family $F$. 
Let $A$ be a task foreign to family $F$.
Define the makespan $\mspan$ of a plan $\alpha$ as the wall-clock time it takes from the start of its first task to the end of its last task. 

Under conditions C1 and C2 (negligible fatigue and decay), moving task $A$ from after the collect action to before it (i.e., during the time it takes the external process to complete) changes makespan by:
\[
\Delta\mspan \;=\; \pen_F \;-\; \min(d, \dur'_A) \;+\; R,
\]
where $\dur'_A$ is the duration of $A$ when performed during the wait, and $R \geq 0$ is the recovery that resting over the full wait would provide. 
Filling the wait is better iff the overlap
gained is greater than the sum of the resumption penalty and $R$. 
\end{thmT}

\paragraph{Connection to literature.} 
The three terms of T5 come from three distinct literatures. 
The overlap gain $-\min(d,\dur_A)$ is an concept in classic industrial engineering. The man--machine chart puts the human operator and the machine on the same time scale to identify human idle time while the machine runs, and the standard approach was to fill that time with other work~\citep{barnes1980motion}. 
The chart also shows why maximal time gain is limited: if the filler task lasts beyond when the machine finishes, the machine instead waits for the operator. The time saved is the shorter of these two durations. 
The same idea scaled up to many jobs and machines is the basis of flow-shop scheduling in OR. 
To minimize the makespan, the overlap between jobs is maximized by choosing the order in which jobs are done~\citep[Johnson's rule;][]{johnson1954optimal}.

The resumption penalty $\pen_F$ belongs to the same literature as discussed in Theorem T2. 
The $R$ term parallels the literature associated with T3: waiting for the process to finish is a rest break, and rest can help improve fatigue~\citep{tucker2003impact}, but filling the wait raises fatigue instead. Thus, the fatigue term is the recovery the wait would have bought (A5). 
Empirically, studies have demonstrated that wait times can be used effectively. 
A study from user-centered design in the human-computer interaction literature shows how filling waits with micro-learning can help users learn foreign languages~\citep{cai2015wait}. 

Two aspects are not tracked in our model. 
First, filling a wait requires keeping some attention on when a process has finished, which can be demanding on memory and can also fail~\citep[i.e., forgetting;][]{dismukes2010remembrance}.
Our paradigm assumes that collecting is scheduled and monitoring is free. 
Second, Theorem T5 assumes that $A$ is a foreign task. If the filled task were part of the same family, the term instead becomes $\Delta\mspan = -\min(d,\dur_A) + R$, which becomes a rest vs.~rust (T3) problem. 
Our vignettes for Theorem T5 choose external processes that do not need to be attended to (e.g., letting mortar sit), and the filler tasks are selected from foreign families. We vary the length of $\dur_A$ compared to $d$ for coverage. 
\begin{proof}
Consider two plans $P_1$ and $P_2$, where
$P_1 = \ell_p \seq w_d \seq c_p \seq A$ and $P_2 = \ell_p \seq A \seq w_{d'} \seq c_p$.

$\ell_p$ is the launch action for external process $p$, and $c_p$ is its collect action. $w_d$ is an idle task. $A$ is a foreign task to family $F$, which contains $\ell_p$ and $c_p$. $d'$ is the remaining wait after completing $A$ but before process $p$ finishes:
$d' = \max(0, d - \dur_A')$. 

Since condition C2 (negligible decay) is active, skill in $\famof(A)$ does not change until $A$ is started. Thus, the base work for task $A$, $b_A\big(\skill_{\famof(A)}(s)\big)$, is the same across the two plans. 
Similarly, the collect action $c_p$ always happens after the launch action $\ell_p$, so their base costs are also the same. 

Next, we consider setup. We can safely ignore any switching costs from before the plan to $\ell_p$ since those will be equal across plans. 
In $P_1$, the setup register is not affected by $w_d$ (Definition 3). As $\ell_p$ and $c_p$ belong to the same task family, the only switching cost paid is the transition from $c_p$ to $A$, costing $\pen_{\famof(A)}$ since $A$ is foreign. 
In $P_2$, $A$ is conducted after $\ell_p$ and switches the setup register to $\famof(A)$, costing the same amount. Then, when conducting $c_p$, whether there was a rest action $w_d$ with positive duration or not, we need to pay an additional $\pen_{F}$. 

Finally, consider fatigue. Since C1 (negligible fatigue) is active, the $\load$ can only fall during rest $w_d$ (A5). 
Let $\load$ be the load after $\ell_p$, $\load'$ be the load after rest $w_d$, and let $\load''$ be the load after the partial rest $w_{d'}$, with $d' \leq d$. 
By A5, $\load' \le \load'' \le \load$.
Then fatigue costs change at two parts: In $P_1$, $A$ runs at load $\load'$, but in $P_2$ it runs at $\load$. 
In $P_1$, $c_p$ runs at $\load'$ but in $P_2$ it runs at $\load''$. These two differences contribute to make $R = [h(\load) - h(\load')] + [h(\load'') - h(\load')]$.

Now consider the total makespans $M_1$ and $M_2$. 
\begin{align*}
    M_1 &= \dur_\ell + d + \dur_c + \dur_A \\
    &= \dur_\ell + d + b_c + h(\load') + 0 + b_A + h(\load') + \pen_{\famof(A)} \\
    &= \dur_\ell + d + b_c + h(\load') + b_A + h(\load') + \pen_{\famof(A)} \\
    M_2 &= \dur_\ell + \dur_A' + d' + \dur_c' \\
    &= \dur_\ell + \max(d, \dur_A') + \dur_c' \\
    &= \dur_\ell + \max(d, b_A + h(\load) + \pen_{\famof(A)}) + b_c + h(\load'') + \pen_{F}
\end{align*}

Using $\max(d, \dur_A') - d - \dur_A' = -\min(d, \dur_A')$,
\begin{align*}
M_2 - M_1 &= \big(\max(d, \dur_A') - d - \dur_A'\big) + (\dur_A' - \dur_A)
   + (\dur_c' - \dur_c) \\
&= -\min(d, \dur_A') + (h(\load) - h(\load'))
   + \pen_{F} + (h(\load'') - h(\load')) \\
&= \pen_{F} - \min(d, \dur_A') + R
\end{align*}
$R$ is non-negative because $h$ is non-decreasing (A3) and
$\load' \le \load'' \le \load$. 
Filling the wait is therefore beneficial when
\[
\min(d, \dur_A') \;>\; \pen_{F} + R,
\]
i.e., the time overlapped during the external process is greater than the sum of the resumption penalty and the recovery that the wait provides.
\end{proof}
 
\subsubsection{Degeneracy}
\label{app:degeneracy}
 
\begin{proposition}[Theorems depend on state]
\label{prop:degenerate}
As part of our construction, if we remove state, i.e., $\dur(a, s) = \dur(a)$ for all $s$, then the time taken for all tasks $\Tt$ is now additive and order-invariant.
Our theorems T1--T4 collapse to equalities, and T5 reduces to the simple identity $\Delta\mspan = -\min(d, \dur_A)$.
\end{proposition}
\begin{proof}
With state-free terms, order no longer matters when computing time taken. Thus, the sum $\Tt$ is simply permutation-invariant. 
Switching penalties cannot be expressed because the previous task's family is not recorded. Fatigue cost is at most constant, and skill decay and recovery cannot be expressed. 

T1: Skill change cannot affect duration and thus T1 simply becomes $\dur(b, \del(a, s)) = \dur(b, s)$.

T2: There is no setup register and fatigue costs are constant, so $\dur(a_2, \del(b, \del(a_1, s))) = \dur(a_2, \del(a_1, s))$, and thus $\dur(a_1, s) + \dur(a_2, \del(b, \del(a_1, s))) = \dur(a_1, s) + \dur(a_2, \del(a_1, s)) = \dur(a, s)$. 

T3: Waiting no longer recovers fatigue since it is at most constant. The rest vs.~rust trade-off $\dur(a_2, \del(w_d, \del(a_1, s))) - \dur(a_2, \del(a_1, s)) = \dur(a_2, s) - \dur(a_2, s) = 0$. 

T4: All costs of the same task are equal, so batching yields no advantage: \\
$\Tt(x_1;x_2;y_1;y_2, s) = \dur(x_1, s) + \dur(x_2, s) + \dur(y_1, s) + \dur(y_2, s) = \Tt(x_1;y_1;x_2;y_2, s)$. 

T5: $\pen_F = R = 0$, so only the simple $\min$ computation remains, and overlap can never hurt anymore: \\
$M_2 - M_1 = (\dur_\ell + \max(d, \dur_A') + \dur_c') - (\dur_\ell + d + \dur_c + \dur_A) = \max(d, \dur_A) - d - \dur_A = -\min(d, \dur_A)$.
\end{proof}
\textbf{This is the core justification of our theory: state-dependence is what allows our theory to express the various findings in literature, generating the entire phenomena space.} Without it, only simple scheduling arithmetic remains, which a simple duration lookup table model suffices to predict. 
 
\subsubsection{The register idealization and a countermodel}
\label{app:graded}
 
A3 makes the setup register able to only hold a single family, and the removal of that family is immediate after another task starts. 
Instead, an alternate view is that the register decays with foreign work, and is also restored gradually. This matches empirical findings where
resumption costs grow with the length and the level of demand that the interrupting task has~\citep{altmann2002memory, monk2008interruption}. 
Doing this would also make Theorem T2 (interruption overhead) lose its lower bound of one switching penalty, and only have overhead be positive. 
However, we show below using a countermodel that T4 (batching dominance) relies on having a way to reset switching costs after returning to a family, such as the current register structure, and not just any model with cross-family switching costs.
 
\begin{example}[Cumulative interference]
\label{ex:cumulative}
Consider the countermodel with the following properties.
Let the state $s$ track two counters, $(\#X, \#Y)$, consisting of the number of completed tasks for families $X$ and $Y$.
Let $\dur(X, s) = 1 + (\#Y)^2$ and $\dur(Y, s) = 1 + (\#X)^2$. Let there be no fatigue and skill terms. 
Definition~2 holds; A1 holds (repeating $X$ does not impact $\#Y$); T1 holds (each foreign task strictly raises the counter and thus the penalty). 

However, in the batched condition, $X_1 X_2 Y_1 Y_2$ costs $1 + 1 + 5 + 5 = 12$, while in the interleaved condition, $X_1 Y_1 X_2 Y_2$ costs $1 + 2 + 2 + 5 = 10$.
Here, interleaving strictly wins. The reason is that when foreign harm accumulates quadratically, the victim tasks are best run early. 
This is exactly where the countermodel violates A3---switching costs are not routed through a register that an own-family task resets, which is exactly what the proof for T4 needs.
\end{example}

%% file: Appendices/exp1/axiom_examples.tex
\subsection{Grounded Examples of Axioms and Theorems}
\label{app:exp1_axiom_examples}

The axioms and theorems of Section~\ref{sec:exp1_axioms} are stated abstractly, over families, state components, and events. 
To make their content concrete, we restate each one alongside a worked example in a single domain, where a woman named Kathryn is laying a brick garden wall, a unit of work being one row. 
The same family task we use is laying brick steps, and the foreign task is working through logic puzzles. 
Examples here are illustrative and are not vignettes used in the evaluation; see Appendix~\ref{app:exp1_prompts} for the vignettes.

The axiom/theorem is first stated in \textbf{bold}. 
Then, a brief 1-sentence summary is provided in \textit{italics}. 
Lastly, an example is given within the setting of a brick-laying primary task and a puzzle secondary task with character Kathryn. 

\subsubsection{Axiom grounded examples}

\paragraph{A1 (learning).} \textit{Same-family work never lowers skill for that family.} If Kathryn laid a row of brick steps yesterday, that would not reduce her skill at tasks in the same family, such as laying a brick wall today. 
\textit{Higher skill never slows base work.} If Kathryn practiced laying brick last week and got better at it, she would be no slower at laying a row today than she was before.

\paragraph{A2 (decay).} \textit{Time away from a family weakly erodes its skill, increasing with length.} If Kathryn has not laid any brick in two years, she is no more skilled at it than she was when she stopped, and she is at most as skilled as she would have been had she only stopped for two weeks.

\paragraph{A3 (screening).} \textit{The duration of a task is its base work, plus a fatigue cost, plus a resumption penalty paid only if the setup register holds a foreign family.} The time Kathryn takes to lay the next row is the time the row itself takes, plus whatever she loses to being tired, plus---if the last thing she did was a logic puzzle---the time she spends re-reading the guide string, remembering where she was, and getting back into the work. Had she instead just come off laying a step, that last term is not paid. 
\textit{Performing a task sets the register to its own family.} Once she has laid the first row after the puzzle, the penalty is behind her: the second row does not pay it again.

\paragraph{A4 (loading).} \textit{Every task weakly raises fatigue.} Laying a row of brick leaves Kathryn no less tired than she was before she laid it, and the same is true of a logic puzzle. No task she can perform leaves her fresher than it found her.

\paragraph{A5 (recovery).} \textit{Rest weakly lowers fatigue, no less with its length.} Sitting down for ten minutes leaves Kathryn no more tired than she was when she sat down, and sitting down for an hour leaves her no more tired than ten minutes would have.

\subsubsection{Theorem grounded examples}

\paragraph{T1 (no free transfer).} \textit{Foreign work never speeds up a task.} If Kathryn breaks off the wall to work on a logic puzzle, the remaining rows do not go faster than they would have. The puzzle leaves her skill at brickwork untouched (it is foreign), clears her setup so that she must get back into the wall (A3), and leaves her no less tired (A4). Every channel through which it could act either does nothing or works against her.

\paragraph{T2 (interruption overhead).} \textit{A task interrupted by foreign work takes at least one resumption penalty longer, and each additional interruption adds at least one more.} If Kathryn stops midway through the wall to do a puzzle, the wall takes longer by at least the time it costs her to get back into it. If she stops for a puzzle three separate times, she pays that cost three times over.

\paragraph{T3 (rest versus rust).} \textit{Interrupting a task with rest pays no penalty but trades fatigue recovery against skill decay; the sign is open.} If Kathryn stops laying brick and rests, she does not pay a resumption penalty, because rest does not put a foreign family in her setup register. She does recover (A5), which speeds the remaining rows, and she does lose skill (A2), which slows them. Which effect wins is not settled by the axioms: for a ten-minute break by a tired bricklayer, recovery plainly dominates; for a two-year layoff, decay does. Vignettes in this family therefore stipulate how tired she is and how perishable the skill is, or else deliberately leave both open.

\paragraph{T4 (batching dominance).} \textit{With negligible fatigue and decay, batching foreign work dominates interleaving.} Suppose Kathryn has two rows and two puzzles to get through. Laying both rows and then doing both puzzles (XXYY) crosses between families once. Alternating between them (XYXY) crosses three times, and each crossing costs her a resumption penalty. So long as getting back into either kind of work costs something, the batched order finishes no later, and finishes strictly earlier whenever any penalty is positive.

\paragraph{T5 (overlap).} \textit{Filling a wait on an external process with foreign work pays off only if the time overlapped exceeds the resumption penalty plus the recovery the wait provides.} Kathryn cannot lay the upper rows until the mortar below has set, a wait she does not control. She could spend that wait on a logic puzzle. Doing so buys her whatever puzzle work now overlaps the setting time, but costs her the penalty of getting back into the wall afterwards, plus the recovery she would have had if she had spent the wait resting. When the mortar sets in two minutes, filling the wait is likely not worth it; when it sets in an hour, it likely is.

%% file: Appendices/exp1/names_exp1.tex
\subsection{Vignette character names}
\label{app:exp1_names}
Names are drawn at random from a fixed pool of common US first names and functions purely as a rotation variable. We summed Social Security Administration national birth counts from the \textit{Baby Names from Social Security Card Applications - National Data} dataset~\citep{ssa_babynames} for each (name, sex) pair over birth years 1980--2000, took the 200 most frequent names of each sex, and drew five names per sex by simple random sampling without replacement under a fixed seed, for ten names in total. Aside of pronouns, no other element of the vignette varies with the name, and names are not used as a factor in our analyses; the draw script, the sampled pools, and the resulting name list will be included in the replication materials.

%% file: Appendices/exp1/prompts_exp1.tex

\subsection{Prompts}
\label{app:exp1_prompts}

Every item is rendered into a single user message with no system prompt and no
few-shot example: the header \texttt{Question:}, the vignette, the question,
the four options in fixed order, and the answer-guideline block, which states
the granularity convention the golds were derived under and the response
format. 
The guidelines differ between vignettes based on the entailed gold answer from the theory: wording is different based on whether strictly greater/lesser ($>$ or $<$) is entailed, or if equality is also possible under the derivations ($\geq$ or $\leq$). 
The listings below give all fourteen base vignettes in the
brick-wall domain, the set from which the remaining seven domains were
ported: five axiom families (A3 in two variants) and four theorem families
(T3 rest vs.~rust and T5 overlap in three variants each). Items in the other domains and in the
flipped orientation are identical in form and differ only in the vignette and
question text.

\textbf{Axiom tier.}

\begin{lstlisting}[
    basicstyle=\ttfamily\footnotesize,
    breaklines=true,
    breakindent=0pt,
    breakautoindent=false,
    breakatwhitespace=true,
    columns=flexible,
    keepspaces=false,
    frame=single,
    caption={A1, learning: practice quantity manipulated, skill level left
    unstated. Item \texttt{A1-001}, gold (b).},
    label={lst:prompt-A1}
]
Question:
Kathryn has a brick garden wall to lay, three rows high. She has only laid brick a few times before.

If instead she had laid brick plenty of times before, would the three rows take longer, shorter, or about the same? Assume that she isn't tired at all in either setting.

(a) Longer
(b) Shorter
(c) About the same
(d) Could go either way

Answer guidelines:
Treat 'longer' and 'shorter' as including the possibility of equality. Answer 'about the same' only if it would make no difference at all. Answer in the following format, do not include anything else:
Answer: a single letter
Explanation: 1 to 2 sentences explaining why you chose that answer.
\end{lstlisting}

\begin{lstlisting}[
    basicstyle=\ttfamily\footnotesize,
    breaklines=true,
    breakindent=0pt,
    breakautoindent=false,
    breakatwhitespace=true,
    columns=flexible,
    keepspaces=false,
    frame=single,
    caption={A2, decay: gap since last practice manipulated. Item
    \texttt{A2-001}, gold (a).},
    label={lst:prompt-A2}
]
Question:
Mariah has a brick garden wall to lay, three rows high. She last laid brick a week ago.

If instead she had last laid brick five years ago, would the three rows take longer, shorter, or about the same? Assume that she isn't tired at all in either setting.

(a) Longer
(b) Shorter
(c) About the same
(d) Could go either way

Answer guidelines:
Treat 'longer' and 'shorter' as including the possibility of equality. Answer 'about the same' only if it would make no difference at all. Answer in the following format, do not include anything else:
Answer: a single letter
Explanation: 1 to 2 sentences explaining why you chose that answer.
\end{lstlisting}

\begin{lstlisting}[
    basicstyle=\ttfamily\footnotesize,
    breaklines=true,
    breakindent=0pt,
    breakautoindent=false,
    breakatwhitespace=true,
    columns=flexible,
    keepspaces=false,
    frame=single,
    caption={A3, screening, interruption-vs-continue variant: a foreign
    interruption against uninterrupted work. Item \texttt{A3-001}, gold
    (a).},
    label={lst:prompt-A3-interrupt}
]
Question:
Tiffany is partway through laying a brick garden wall and has three rows left. She isn't tired at all. She has just finished a row; the mortar for the next one is mixed and she knows exactly where the next brick goes. She is about to start on the next row.

If instead she stops now for a while to work on a logic puzzle before coming back to finish the last three rows, will the three remaining rows take longer, shorter, or about the same?

(a) Longer
(b) Shorter
(c) About the same
(d) Could go either way

Answer guidelines:
Count any difference, however small, as longer or shorter. Answer 'about the same' only if it would make no difference at all. Answer in the following format, do not include anything else:
Answer: a single letter
Explanation: 1 to 2 sentences explaining why you chose that answer.
\end{lstlisting}

\begin{lstlisting}[
    basicstyle=\ttfamily\footnotesize,
    breaklines=true,
    breakindent=0pt,
    breakautoindent=false,
    breakatwhitespace=true,
    columns=flexible,
    keepspaces=false,
    frame=single,
    caption={A3, screening, class-comparison variant: a related interrupter
    against a foreign one, gap length held fixed and the two interrupters
    stipulated equally tiring. Item \texttt{A3-002}, gold (a).},
    label={lst:prompt-A3-class}
]
Question:
Tiffany is partway through laying a brick garden wall and has three rows left. She isn't tired at all. She has just finished a row; the mortar for the next one is mixed and she knows exactly where the next brick goes. Before starting the next row, she is going to spend a few minutes adding bricks to the garden steps nearby.

If instead she spends those few minutes working on a logic puzzle, assuming the garden steps and the logic puzzle tire her out equally, will the three remaining rows afterwards take longer, shorter, or about the same?

(a) Longer
(b) Shorter
(c) About the same
(d) Could go either way

Answer guidelines:
Count any difference, however small, as longer or shorter. Answer 'about the same' only if it would make no difference at all. Answer in the following format, do not include anything else:
Answer: a single letter
Explanation: 1 to 2 sentences explaining why you chose that answer.
\end{lstlisting}

\begin{lstlisting}[
    basicstyle=\ttfamily\footnotesize,
    breaklines=true,
    breakindent=0pt,
    breakautoindent=false,
    breakatwhitespace=true,
    columns=flexible,
    keepspaces=false,
    frame=single,
    caption={A4, loading: prior load on the morning manipulated, fatigue left
    as the manipulated output, and decay over the morning stipulated absent.
    Item \texttt{A4-001}, gold (a).},
    label={lst:prompt-A4}
]
Question:
Cody has a brick garden wall to lay, three rows high, this afternoon. He has spent the morning relaxing at home.

If instead he had spent the morning helping a friend move furniture, would the three rows take longer, shorter, or about the same? Assume that he doesn't get less skilled at laying brick just by spending a morning on something else.

(a) Longer
(b) Shorter
(c) About the same
(d) Could go either way

Answer guidelines:
Treat 'longer' and 'shorter' as including the possibility of equality. Answer 'about the same' only if it would make no difference at all. Answer in the following format, do not include anything else:
Answer: a single letter
Explanation: 1 to 2 sentences explaining why you chose that answer.
\end{lstlisting}

\begin{lstlisting}[
    basicstyle=\ttfamily\footnotesize,
    breaklines=true,
    breakindent=0pt,
    breakautoindent=false,
    breakatwhitespace=true,
    columns=flexible,
    keepspaces=false,
    frame=single,
    caption={A5, recovery: an idle rest at high fatigue, with the rest itself
    excluded from the measured quantity and decay during the rest stipulated
    absent. Item \texttt{A5-001}, gold (b).},
    label={lst:prompt-A5}
]
Question:
Kathryn is partway through laying a brick garden wall and has three rows left. She is extremely tired. She has just finished a row and is about to start on the next one.

If instead she sits down and rests for fifteen minutes before starting the next row, not counting the time she spends resting, will the three remaining rows take longer, shorter, or about the same? Assume the fifteen-minute break would not make her any less skilled at laying brick.

(a) Longer
(b) Shorter
(c) About the same
(d) Could go either way

Answer guidelines:
Treat 'longer' and 'shorter' as including the possibility of equality. Answer 'about the same' only if it would make no difference at all. Answer in the following format, do not include anything else:
Answer: a single letter
Explanation: 1 to 2 sentences explaining why you chose that answer.
\end{lstlisting}

\textbf{Theorem tier.}

\begin{lstlisting}[
    basicstyle=\ttfamily\footnotesize,
    breaklines=true,
    breakindent=0pt,
    breakautoindent=false,
    breakatwhitespace=true,
    columns=flexible,
    keepspaces=false,
    frame=single,
    caption={T2, interruption overhead: $k$ re-entries pay $k$ switching
    costs, total interruption time held fixed, with decay and fatigue
    stipulated absent. Item \texttt{T2-001}, gold (a).},
    label={lst:prompt-T2}
]
Question:
Henry is partway through laying a brick garden wall and has a few rows left. He isn't tired at all. He has just finished a row; the mortar for the next one is mixed and he knows exactly where the next brick goes. While laying the last few rows, he is going to stop once in the middle, for six minutes, to work on a logic puzzle.

If instead he stops twice along the way, for three minutes each time, to work on the puzzle, will the remaining rows take longer, shorter, or about the same? Assume none of the stops would make him less skilled at laying brick, and that neither the bricklaying nor the puzzle solving take long enough to tire him at all.

(a) Longer
(b) Shorter
(c) About the same
(d) Could go either way

Answer guidelines:
Count any difference, however small, as longer or shorter. Answer 'about the same' only if it would make no difference at all. Answer in the following format, do not include anything else:
Answer: a single letter
Explanation: 1 to 2 sentences explaining why you chose that answer.
\end{lstlisting}

\begin{lstlisting}[
    basicstyle=\ttfamily\footnotesize,
    breaklines=true,
    breakindent=0pt,
    breakautoindent=false,
    breakatwhitespace=true,
    columns=flexible,
    keepspaces=false,
    frame=single,
    caption={T3, rest vs.~rust, sign forced toward rest: recovery and decay
    are both live, and a calibration sentence bounds three months of decay
    below the cost of mild fatigue, so recovery from extreme fatigue
    dominates. Item \texttt{T3-001}, gold (b).},
    label={lst:prompt-T3-rest}
]
Question:
Katrina has a brick garden wall to lay for a neighbor, three rows high, and can do it either this Saturday or on a Saturday three months from now. She has laid brick plenty of times before, and knows that even a three month break from laying brick slows her down less than being a little tired does. She is in the middle of an exhausting few weeks at work and will be extremely tired this Saturday; she knows she won't be tired at all on the Saturday in three months. She plans to lay the wall this Saturday.

If instead she waits and lays it on the Saturday in three months, will the three rows take longer, shorter, or about the same?

(a) Longer
(b) Shorter
(c) About the same
(d) Could go either way

Answer guidelines:
Treat 'longer' and 'shorter' as including the possibility of equality. Answer 'about the same' only if it would make no difference at all. Answer in the following format, do not include anything else:
Answer: a single letter
Explanation: 1 to 2 sentences explaining why you chose that answer.
\end{lstlisting}

\begin{lstlisting}[
    basicstyle=\ttfamily\footnotesize,
    breaklines=true,
    breakindent=0pt,
    breakautoindent=false,
    breakatwhitespace=true,
    columns=flexible,
    keepspaces=false,
    frame=single,
    caption={T3, rest vs.~rust, sign forced toward rust: recovery and decay
    are both live, and a calibration sentence bounds even one week of decay
    above the cost of mild fatigue, so three months of decay dominates. Item
    \texttt{T3-002}, gold (a).},
    label={lst:prompt-T3-rust}
]
Question:
Last week, Sara learned to lay brick and practiced twice. She has a short brick garden wall to lay for her neighbor, and can do it either this Saturday or on a Saturday three months from now; she won't get a chance to lay brick in between. She knows that, this early on, even a one week break from practicing laying brick would slow her down more than being a little tired does. She is in the middle of a busy few weeks at work and will be a little tired this Saturday; she knows she won't be tired at all on the Saturday in three months. She is planning to lay the wall this Saturday.

If instead she waits and lays it on the Saturday in three months, will the wall take longer, shorter, or about the same?

(a) Longer
(b) Shorter
(c) About the same
(d) Could go either way

Answer guidelines:
Treat 'longer' and 'shorter' as including the possibility of equality. Answer 'about the same' only if it would make no difference at all. Answer in the following format, do not include anything else:
Answer: a single letter
Explanation: 1 to 2 sentences explaining why you chose that answer.
\end{lstlisting}

\begin{lstlisting}[
    basicstyle=\ttfamily\footnotesize,
    breaklines=true,
    breakindent=0pt,
    breakautoindent=false,
    breakatwhitespace=true,
    columns=flexible,
    keepspaces=false,
    frame=single,
    caption={T3, rest vs.~rust, open sign: recovery and decay are both live
    with no calibration between them, and fatigue is pinned only
    comparatively, so the description does not settle the sign. Item
    \texttt{T3-003}, gold (d).},
    label={lst:prompt-T3-open}
]
Question:
Last week, Tiffany learned to lay brick and practiced twice. She has a short brick garden wall to lay for her neighbor, and can do it either this Saturday or on a Saturday three months from now. She is in the middle of a busy few weeks at work and will be quite tired this Saturday; she knows she'll be less tired on the Saturday in three months. She is planning to lay the wall this Saturday.

If instead she waits and lays it on the Saturday in three months, will the wall take longer, shorter, or about the same?

(a) Longer
(b) Shorter
(c) About the same
(d) Could go either way

Answer guidelines:
Count any difference, however small, as longer or shorter. Answer 'about the same' only if it would make no difference at all. Answer in the following format, do not include anything else:
Answer: a single letter
Explanation: 1 to 2 sentences explaining why you chose that answer.
\end{lstlisting}

\begin{lstlisting}[
    basicstyle=\ttfamily\footnotesize,
    breaklines=true,
    breakindent=0pt,
    breakautoindent=false,
    breakatwhitespace=true,
    columns=flexible,
    keepspaces=false,
    frame=single,
    caption={T4, batching: alternating two tasks against batching them, with
    fatigue at the floor, a feasibility clause, and decay stipulated absent.
    Item \texttt{T4-001}, gold (b).},
    label={lst:prompt-T4}
]
Question:
This afternoon, Katrina needs to lay two rows of brick on her garden wall and solve two logic puzzles for a class she is taking. She isn't tired at all at the moment, and neither task is tiring for her, even if done twice in a row. She is planning to alternate: a row of brick, then a puzzle, then the other row, then the other puzzle.

If instead she lays both rows first and then solves both puzzles, will the afternoon's work take longer, shorter, or about the same? Assume that she doesn't get less skilled at either task just by spending some time on the other.

(a) Longer
(b) Shorter
(c) About the same
(d) Could go either way

Answer guidelines:
Count any difference, however small, as longer or shorter. Answer 'about the same' only if it would make no difference at all. Answer in the following format, do not include anything else:
Answer: a single letter
Explanation: 1 to 2 sentences explaining why you chose that answer.
\end{lstlisting}

\begin{lstlisting}[
    basicstyle=\ttfamily\footnotesize,
    breaklines=true,
    breakindent=0pt,
    breakautoindent=false,
    breakatwhitespace=true,
    columns=flexible,
    keepspaces=false,
    frame=single,
    caption={T5, overlap, fill wins: the fill fits inside the wait and the
    re-entry toll is stipulated folk-small, with the fill stipulated not
    tiring and decay during the wait stipulated absent. Item \texttt{T5-001},
    gold (b).},
    label={lst:prompt-T5-win}
]
Question:
This afternoon, Katrina is laying a brick garden wall and also needs to solve a logic puzzle for a class she is taking. The wall's lower rows are done, and their mortar needs an hour or so to firm up before it can take the weight of the final row. She isn't tired at all, and solving the puzzle isn't tiring for her. The puzzle would take her about twenty minutes, and it only takes her a moment to get back into the rhythm of laying. She is planning to sit and wait out the hour, lay the final row, and then solve the puzzle.

If instead she solves the puzzle during the hour's wait, will finishing everything take longer, shorter, or about the same? Assume that she doesn't get less skilled at laying brick just by spending an hour on something else.

(a) Longer
(b) Shorter
(c) About the same
(d) Could go either way

Answer guidelines:
Count any difference, however small, as longer or shorter. Answer 'about the same' only if it would make no difference at all. Answer in the following format, do not include anything else:
Answer: a single letter
Explanation: 1 to 2 sentences explaining why you chose that answer.
\end{lstlisting}

\begin{lstlisting}[
    basicstyle=\ttfamily\footnotesize,
    breaklines=true,
    breakindent=0pt,
    breakautoindent=false,
    breakatwhitespace=true,
    columns=flexible,
    keepspaces=false,
    frame=single,
    caption={T5, overlap, fill loses: the re-entry toll is pinned
    comparatively above the wait it would save, with the main task stipulated
    not tiring and decay stipulated absent. Item \texttt{T5-002}, gold (a).},
    label={lst:prompt-T5-lose}
]
Question:
This afternoon, Keith is laying a brick garden wall and also needs to reply to a message from his class group, which would take him a few minutes. He has just finished laying a row and has three rows left to lay, and the mortar he mixed for them needs to sit for ten minutes before it's ready to use. He isn't tired at all, and laying the three rows isn't tiring for him. But once he turns to something else, it takes him a while to find the rhythm of laying again, more than the ten minutes the mortar needs. He is planning to wait out the ten minutes, lay the three rows, and reply afterward.

If instead he replies to the message during the ten-minute wait, will finishing everything take longer, shorter, or about the same? Assume that he doesn't get less skilled at either task just by spending some time on something else.

(a) Longer
(b) Shorter
(c) About the same
(d) Could go either way

Answer guidelines:
Count any difference, however small, as longer or shorter. Answer 'about the same' only if it would make no difference at all. Answer in the following format, do not include anything else:
Answer: a single letter
Explanation: 1 to 2 sentences explaining why you chose that answer.
\end{lstlisting}

\begin{lstlisting}[
    basicstyle=\ttfamily\footnotesize,
    breaklines=true,
    breakindent=0pt,
    breakautoindent=false,
    breakatwhitespace=true,
    columns=flexible,
    keepspaces=false,
    frame=single,
    caption={T5, overlap, open sign: the re-entry toll is declared variable
    over a range that straddles the wait. Item \texttt{T5-003}, gold (d).},
    label={lst:prompt-T5-open}
]
Question:
This afternoon, Katrina is laying a brick garden wall and also needs to reply to a message from her class group, which would take her about five minutes. She has just finished laying a row and has three rows left to lay, and the mortar she mixed for them needs to sit for ten minutes before it's ready to use. She isn't tired at all. But once she turns to something else, it takes her some time to find the rhythm of laying again, a minute or two on some days and ten minutes or more on others. She is planning to wait out the ten minutes, lay the three rows, and reply afterward.

If instead she replies to the message during the ten-minute wait, will finishing everything take longer, shorter, or about the same?

(a) Longer
(b) Shorter
(c) About the same
(d) Could go either way

Answer guidelines:
Count any difference, however small, as longer or shorter. Answer 'about the same' only if it would make no difference at all. Answer in the following format, do not include anything else:
Answer: a single letter
Explanation: 1 to 2 sentences explaining why you chose that answer.
\end{lstlisting}

%% file: Appendices/exp1/order_shuffle_exp1.tex
\subsection{Option-order sensitivity in Experiment 1}
\label{app:shuffle}

In Experiment 1, our evaluation presents the four response options in a fixed order:
(a)~Longer, (b)~Shorter, (c)~About the same, (d)~Could go either way. This positioning makes the meaning of option (d) more salient, as ``either way'' now refers to options (a) and (b). However, a potential confounder is models' position bias, which has previously been documented in multiple choice tasks~\citep{wang2024large}, where a model favors a particular slot simply because of the label.

To address this possible confounder, we ran a pilot that permutes the options in a subset of the items, and compare results against the fixed order options under otherwise
identical conditions.

\subsubsection{Design}

\paragraph{Items.} We drew a 64-item subset covering all
nine axiom and theorem families. The subset deliberately over-samples the two
families with the highest share of ``(d) could go either way'' ground truths, because those are the items that, if shuffled, would likely be most sensitive to answer ordering. The remaining seven families
contribute two to four items each, yielding a ground truth balance of 16~(a), 16~(b), and
32~(d). As items appear in both orientations of a pairwise counterfactual structure, for the sake of balance a base item and its flip enter the subset together.

Each item was presented under two option orders:
\begin{itemize}[leftmargin=15pt, topsep=0pt, itemsep=2pt]
  \item \textbf{Fixed.} The canonical order used in the main experiment.
  \item \textbf{Shuffled.} The four options permuted uniformly at random, with
    the gold remapped to whichever display letter carries the gold's label.
\end{itemize}
\vspace{2pt}

Permutations are drawn from a seed keyed on the item's base identifier and run index. Keying on the base identifier makes an item and its flip share a permutation, so that the base/flip contrast is not itself confounded with a change of layout. Keying on the run index makes successive presentations of the same item differ. Across six repeated runs, each item drew a mean of 5.4 distinct permutations (minimum 3, maximum 6), and 14 of
384 draws happened to return the identity permutation. 

\paragraph{Sampling and decoding.} Every item was presented six times per model
in both conditions, resulting in giving $64 \times 6 \times 3 \times 2 = 2{,}304$ requests in total. Each request is a single turn with no system prompt, at each model's default reasoning level, matching
the main experiment.

\paragraph{Scoring.}
For each answer, we invert the permutation and recover the canonical label. We report two quantities:

\begin{itemize}[leftmargin=15pt, topsep=0pt, itemsep=2pt]
  \item \textbf{Pooled accuracy}, over all repeated runs.
  \item \textbf{Unanimity}, the share of items whose six replicates all returned the same canonical label.
\end{itemize}

Claude Opus~5 declined 30 of 384 rows in each arm with a content-classifier
refusal. The refusals fall on the same five items in both arms
(\texttt{C5-003}, \texttt{C5-003-R}, \texttt{INT-007}, \texttt{OVL-009},
\texttt{OVL-009-R}). Those rows are dropped and Claude's results are on the remaining subset of 59 items.

\subsubsection{Results}

Table~\ref{tab:shuffle-acc} gives accuracy and consistency for both conditions across three models tested. Permuting
the options moves pooled accuracy by less than a percentage point for GPT and Gemini and by $+2.2$ points for Claude, and every 95\% confidence interval on the
paired difference is consistent with no effect. Each interval is the 2.5th–97.5th percentile range of 10,000 two-stage cluster-bootstrap replicates. Each replicate resamples items with replacement, and within each item, resamples that item's response rows with replacement (independently for the two conditions); the difference in pooled accuracy is recomputed on each sample. 

\begin{table}[t]
\centering
\caption{Accuracy and consistency (\%) under fixed and shuffled option order. \emph{Unanimous} is the share of items answered with the same canonical label on all six shuffled reruns. $\Delta$ is shuffled minus fixed accuracy, reported with a 95\% paired cluster bootstrap interval.}
\vspace{8pt}
\small
\begin{tabular}{llrrrr}
\toprule
Model & Order & Scored & Accuracy & Unanimous & $\Delta$ accuracy (95\% CI) \\
\midrule
\multirow{2}{*}{Claude Opus 5}
  & Fixed    & 351 & 65.5 & 72.9  & \multirow{2}{*}{$+2.2$ \ ($-1.9$, $+6.5$)} \\
  & Shuffled & 353 & 67.7 & 74.6  & \\
\addlinespace
\multirow{2}{*}{GPT-5.6 Sol}
  & Fixed    & 384 & 93.5 & 98.4  & \multirow{2}{*}{$+0.3$ \ ($+0.0$, $+1.3$)} \\
  & Shuffled & 384 & 93.8 & 100.0 & \\
\addlinespace
\multirow{2}{*}{Gemini 3.7 Flash}
  & Fixed    & 384 & 94.5 & 96.9  & \multirow{2}{*}{$-0.8$ \ ($-2.3$, $+0.0$)} \\
  & Shuffled & 384 & 93.8 & 100.0 & \\
\bottomrule
\end{tabular}
\label{tab:shuffle-acc}
\end{table}

\begin{figure}[t]
\centering
\includegraphics[width=.8\textwidth]{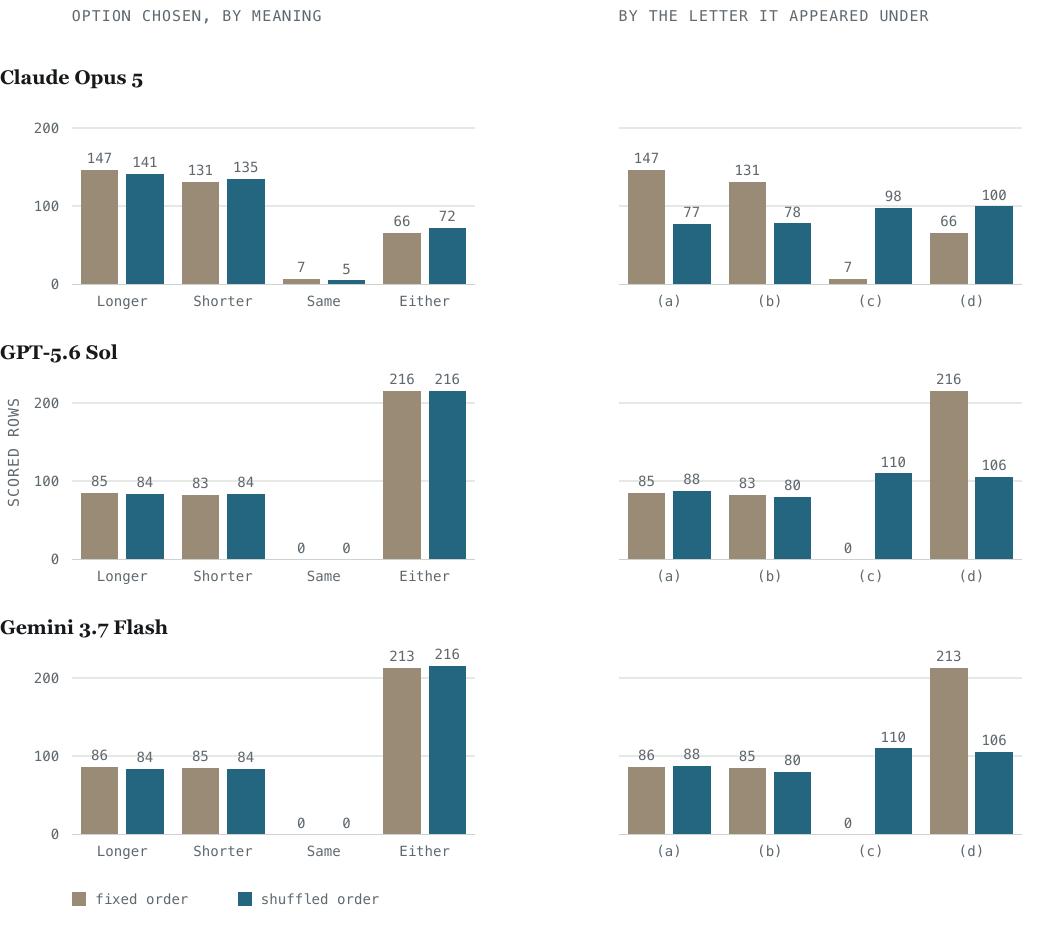}
\caption{Scored rows by the option a model chose (left) and by the letter that option appeared under (right). Under shuffling the label counts barely move, but the slot counts flatten, indicative of a model responding to option content rather than option position.}
\label{fig:shuffle-dist}
\end{figure}

Figure~\ref{fig:shuffle-dist} compares between what the models chose and which position they chose it. Under the shuffled order, the distribution over options flattens towards uniform, with $\chi^2$ values of $5.3$, $6.4$ and $6.4$ on three degrees of freedom for Claude, GPT and Gemini respectively. This suggests that models are not answering based on the display option. 

The same conclusion follows from splitting the shuffled arm by the slot the gold happened to land in (Table~\ref{tab:shuffle-slot}). A model with a slot
preference would score well when the gold coincided with its preferred letter and poorly otherwise; instead accuracy is flat across the four slots for all three models, suggesting no positional biases.

\begin{table}[t]
\centering
\caption{Shuffled-arm pooled accuracy (percent) by the display slot the gold
was permuted into, with 95\% cluster-bootstrap intervals. Row counts per cell
range from 73 to 116.}
\vspace{8pt}
\small
\begin{tabular}{lcccc}
\toprule
Model & Gold in (a) & Gold in (b) & Gold in (c) & Gold in (d) \\
\midrule
Claude Opus 5    & 62.0 \ (47.6, 75.9) & 69.9 \ (55.7, 82.7) & 67.6 \ (54.0, 79.6) & 71.0 \ (58.0, 82.4) \\
GPT-5.6 Sol      & 94.1 \ (86.4, 100)  & 91.1 \ (81.0, 98.8) & 93.1 \ (85.3, 99.1) & 96.2 \ (90.7, 100)  \\
Gemini 3.7 Flash & 94.1 \ (86.4, 100)  & 91.1 \ (81.0, 98.8) & 93.1 \ (85.3, 99.1) & 96.2 \ (90.8, 100)  \\
\bottomrule
\end{tabular}
\label{tab:shuffle-slot}
\end{table}

%% file: Appendices/exp1/token_limits_and_refusals.tex

\subsection{Token limits and refusals}
\label{app:coverage}

\paragraph{Token limits.} All prompts included an explicit cap on output and thinking tokens. 
GPT-5.6~Sol and Claude Opus~5 were run with a limit of 2{,}000 tokens and Gemini 3.7~Flash with a cap of 5{,}000. Claude occasionally exceeded the 2{,}000-token limit and returned an empty response. For those requests, we re-ran with a limit of 5{,}000 tokens and no truncations re-occurred. 

\paragraph{Refusals.} GPT-5.6~Sol and Gemini 3.7~Flash answered every request. Claude Opus~5 refused to answer a subset of requests due to a content-classifier refusal with category \texttt{cyber}, corresponding to a total of 14.2\% of the main evaluation. 
As our vignettes describe ordinary work tasks and are selected with the criteria to avoid topics that trigger safety flags, we consider the refusals as an API-side classifier failure. Refused rows are treated as missing data and are dropped from all analyses. 
Table~\ref{tab:refusals} provides counts broken down by axiom/theorem and by domain. 

\begin{table}[H]
\centering
\caption{Claude Opus~5 content-classifier refusals in the main run, by axiom or theorem and domain. Queries per cell is the total number of queries made in each domain per axiom/theorem row. Cells are refused responses out of runs in that row and domain. Domains: BS~=~brick steps, BW~=~brick wall, LQ~=~logic questions, MP~=~math problems, CR~=~customer replies, FC~=~fact-check claims, SP~=~social posts, SS~=~summarize sources. }
\vspace{8pt}
\label{tab:refusals}
\small
\setlength{\tabcolsep}{4pt}
\begin{tabular}{lrrrrrrrrrr}
\toprule
& & \multicolumn{8}{c}{Domain} & \\
\cmidrule(lr){3-10}
Axiom/theorem & Queries per cell & BS & BW & LQ & MP & CR & FC & SP & SS & Total \\
\midrule
learning           &  60 & 0 & 0 & 0 & 0 &   0 &   0 &  0 &   0 & 0/480 \\
decay              &  60 & 0 & 0 & 0 & 0 &   0 &   0 &  0 &   0 & 0/480 \\
screening          & 120 & 0 & 0 & 0 & 0 & 107 & 120 &  0 & 108 & 335/960 \\
loading            &  60 & 0 & 0 & 0 & 0 &   0 &  60 &  0 &   0 & 60/480 \\
recovery           &  60 & 0 & 0 & 0 & 0 &   0 &   0 &  0 &   6 & 6/480 \\
interruption count &  60 & 0 & 0 & 0 & 0 &  60 &  60 & 30 &  60 & 210/480 \\
rest-vs-rust       & 180 & 0 & 0 & 0 & 0 &   0 &   0 &  0 &   0 & 0/1440 \\
batching           &  60 & 0 & 0 & 0 & 0 &   0 &  30 &  0 &   0 & 30/480 \\
overlap            & 180 & 0 & 0 & 0 & 0 &  60 & 104 & 60 &  88 & 312/1440 \\
\midrule
Total                    & 840 & 0 & 0 & 0 & 0 & 227 & 374 & 90 & 262 & 953/6720 \\
\bottomrule
\end{tabular}
\end{table}

%% file: Appendices/exp2.tex
\section{Details for experiment 2: estimating completion time}
\subsection{Human task-duration estimation prompt}
\label{app:human-estimation-prompt}

The human task-duration experiment uses the following prompt. The placeholder \texttt{[TASK INSTRUCTIONS]} is replaced with the complete instructions for the corresponding HCAST task. Each request is independent and contains no system prompt.

\begin{lstlisting}[
    basicstyle=\ttfamily\footnotesize,
    breaklines=true,
    breakindent=0pt,
    breakautoindent=false,
    breakatwhitespace=true,
    columns=flexible,
    keepspaces=false,
    frame=single,
    caption={Prompt used to elicit human task-duration estimates.},
    label={lst:human-estimation-prompt}
]
Estimate how long a person from the following population would take to complete the task below. Estimate active wall-clock working time, including reading the instructions, investigation, implementation, testing, and debugging if applicable. Assume that the person does not use AI assistance and has not seen any previous attempts at the task, and that they work independently in the supplied task environment and can use its normal tools, documentation, terminal, and tests. Estimate time conditional on successful task completion.

Your primary estimate is the person's completion time of the task. Also provide 50%, 80%, and 95% uncertainty interval estimates. Do not search for or rely on remembered benchmark results.

POPULATION
Skilled professionals in software engineering, machine learning, and cybersecurity, with the majority having attended world top-100 universities. They have an average of about 5 years of relevant experience. These individuals are unfamiliar with this particular task but have the stated general background.

TASK INSTRUCTIONS
[TASK INSTRUCTIONS]

ESTIMATION INSTRUCTIONS
Return exactly these values in this format:
estimated completion time (minutes): number
50% interval: lower value, upper value
80% interval: lower value, upper value
95% interval: lower value, upper value
rationale: 1-2 sentences explaining the reasoning behind the estimates.
\end{lstlisting}
\subsection{Additional results for human task-duration estimation}
\label{app:human-time-estimation}

Table~\ref{tab:human-reference-containment} reports the complete reference-containment results for the human task-duration experiment. For each model and interval level, we calculate the proportion of tasks for which the HCAST geometric-mean human completion time falls within the model's median interval across 30 repetitions. Because models predict the completion time of an individual successful attempt, while the HCAST reference is an aggregate computed from a limited number of successful attempts, these values should be interpreted as a permissive reference-containment diagnostic rather than formal predictive coverage.

\begin{table}[h]
    \centering
    \caption{\textbf{Reference containment for human completion-time estimates.}
    Each cell reports the proportion and number of tasks for which the HCAST human geometric-mean reference falls within the model's median uncertainty interval across 30 repetitions.}
    \vspace{7pt}
    \label{tab:human-reference-containment}
    \begin{tabular}{lccc}
        \toprule
        \textbf{Model}
        & \textbf{50\% interval}
        & \textbf{80\% interval}
        & \textbf{95\% interval} \\
        \midrule
        Claude Opus 5
        & 43.75\% (7/16)
        & 56.25\% (9/16)
        & 93.75\% (15/16) \\
        
        GPT-5.6 Sol
        & 43.75\% (7/16)
        & 75.00\% (12/16)
        & 87.50\% (14/16) \\
        
        Gemini 3.7 Flash
        & 18.75\% (3/16)
        & 31.25\% (5/16)
        & 50.00\% (8/16) \\
        
        \midrule
        All model--task pairs
        & 35.42\% (17/48)
        & 54.17\% (26/48)
        & 77.08\% (37/48) \\
        \bottomrule
    \end{tabular}
\end{table}

Containment generally increases with the nominal interval width, but remains below the corresponding nominal level in nearly every condition. Gemini 3.7 Flash exhibits the largest shortfall, containing the human reference on only 50.00\% of tasks at the nominal 95\% level. GPT-5.6 Sol and Claude Opus 5 perform better at the widest interval, but neither model is consistently calibrated across all three levels. Aggregated across models, only 77.08\% of model--task pairs contain the human reference at the nominal 95\% level.

\subsection{Self-estimation prompts}
\label{app:self-estimation-prompts}

We use the same first-person prompt template for all three self-estimation conditions. Each API call contains no system prompt and is independent of all other calls. The placeholders \texttt{[AGENT CONFIGURATION]} and \texttt{[TASK INSTRUCTIONS]} are replaced with the corresponding agent profile and complete HCAST task instructions. The three rendered agent-configuration blocks are reported below.

\begin{lstlisting}[
    basicstyle=\ttfamily\footnotesize,
    breaklines=true,
    breakindent=0pt,
    breakautoindent=false,
    breakatwhitespace=true,
    columns=flexible,
    keepspaces=false,
    frame=single,
    caption={Shared prompt template for the self-estimation experiment.},
    label={lst:self-estimation-prompt}
]
Estimate how long you would take to complete the task below, operating with the agent scaffold specified in YOUR AGENT CONFIGURATION. Estimate elapsed wall-clock time from the beginning of your agent run until successful task completion, including model generation, scaffolding runtime, tool use, investigation, implementation, testing, and debugging if applicable. Assume that you do not have human assistance and do not have access to any previous attempts at the task, and that you run independently in the supplied task environment and can use its provided tools, documentation, terminal, and tests. Estimate time conditional on your successful task completion.

Your primary estimate is your completion time of the task. Also provide 50%, 80%, and 95% uncertainty interval estimates. Do not search for or rely on remembered benchmark results.

YOUR AGENT CONFIGURATION
[AGENT CONFIGURATION]

TASK INSTRUCTIONS
[TASK INSTRUCTIONS]

ESTIMATION INSTRUCTIONS
Return exactly these values in this format:
 estimated completion time (minutes): number
50% interval: lower value, upper value
80% interval: lower value, upper value
95% interval: lower value, upper value
rationale: 1-2 sentences explaining the reasoning behind the estimates.
\end{lstlisting}

\textbf{Claude Opus 4.6 configuration.} \\

\begin{lstlisting}[
    basicstyle=\ttfamily\footnotesize,
    breaklines=true,
    columns=fullflexible,
    frame=single
]
Model: Claude Opus 4.6
Reasoning configuration: not specified.
Agent scaffold: ReAct. The agent repeatedly selects an action, observes the result, and then selects its next action.
\end{lstlisting}

\textbf{GPT-5.2 configuration.} \\

\begin{lstlisting}[
    basicstyle=\ttfamily\footnotesize,
    breaklines=true,
    breakindent=0pt,
    breakautoindent=false,
    breakatwhitespace=true,
    columns=flexible,
    keepspaces=false,
    frame=single,
]
Model: GPT-5.2
Reasoning configuration: not specified.
Agent scaffold: Triframe. Before executing an action, the scaffold uses separate stages to provide strategy, generate candidate actions, evaluate the candidates, and select an action to execute.
\end{lstlisting}

\paragraph{Gemini proxy self-estimation configuration.}

Gemini 3 Pro was not callable when we conducted the experiment. We therefore send the shared first-person prompt to Gemini 3.1 Pro Preview while identifying the agent configuration as Gemini 3 Pro. The resulting estimates concern the released Gemini 3 Pro configuration and are not literal self-estimates by the queried model.

\begin{lstlisting}[
    basicstyle=\ttfamily\footnotesize,
    breaklines=true,
    breakindent=0pt,
    breakautoindent=false,
    breakatwhitespace=true,
    columns=flexible,
    keepspaces=false,
    frame=single,
]
Model: Gemini 3 Pro
Reasoning configuration: not specified.
Agent scaffold: ReAct. The agent repeatedly selects an action, observes the result, and then selects its next action.
\end{lstlisting}

\subsection{Additional self-estimation results}
\label{app:self-estimation-results}

Table~\ref{tab:self-reference-containment} reports reference containment for the self-estimation experiment. For each model and interval level, we calculate the proportion of evaluable tasks for which the geometric mean duration across successful METR runs falls within the model's median interval. Claude Opus 4.6 has ground-truth runtimes for 15 tasks because no successful released run is available for one task. GPT-5.2 and Gemini 3 Pro have ground-truth runtimes for all 16 tasks.

\begin{table}[h]
    \centering
    \caption{\textbf{Reference containment for agent-time estimates.}
    Each cell reports the proportion and number of evaluable tasks for which the geometric mean duration across successful METR runs falls within the model's median uncertainty interval across 30 repetitions. The Gemini condition is proxy self-estimation: Gemini 3.1 Pro Preview estimates Gemini 3 Pro.}
    \vspace{7pt}
    \label{tab:self-reference-containment}
    \begin{tabular}{lccc}
        \toprule
        \textbf{Estimator and target}
        & \textbf{50\% interval}
        & \textbf{80\% interval}
        & \textbf{95\% interval} \\
        \midrule
        Claude Opus 4.6 $\rightarrow$ self
        & 40.00\% (6/15)
        & 66.67\% (10/15)
        & 66.67\% (10/15) \\

        GPT-5.2 $\rightarrow$ self
        & 0.00\% (0/16)
        & 0.00\% (0/16)
        & 0.00\% (0/16) \\

        Gemini 3.1 Pro Preview $\rightarrow$ Gemini 3 Pro
        & 0.00\% (0/16)
        & 0.00\% (0/16)
        & 0.00\% (0/16) \\

        \midrule
        All evaluable model--task pairs
        & 12.77\% (6/47)
        & 21.28\% (10/47)
        & 21.28\% (10/47) \\
        \bottomrule
    \end{tabular}
\end{table}

Reference containment is substantially worse than in the human-duration experiment. Across all evaluable model--task pairs, only 21.28\% of the median 95\% intervals contain the observed agent duration. Moreover, increasing Claude Opus 4.6's interval width from 80\% to 95\% does not contain any additional references, while the GPT-5.2 and Gemini conditions remain at zero containment at every interval level.

\subsection{Robustness to additional task information}
\label{app:self-estimation-context}

The main self-estimation experiment provides each estimator with the complete task instructions and its target agent configuration. However, the agents whose runtimes we predict can also inspect files, services, and other details of their execution environments. We therefore repeated the experiment with as much task and environment information as we could reconstruct from METR's public task repository. We used the same 16 tasks, estimator and target configurations, elicitation format, and 30 independent repetitions per estimator--task pair, yielding 1,440 parsed estimates.

For each task, we added a \textsc{Task Environment and Available Materials} block describing the files and services available at the beginning of the run, corresponding public-repository materials, and relevant environment and variant details. Where feasible, we also included the complete contents of relevant text files. The prompt stated that the target agent did not begin with these materials already read or understood and instructed the estimator to include the time required to inspect and comprehend them.

The additional information varied by task family. For the three debugging tasks, we included the supplied Python source files and readable tests. For the two esoteric-language tasks, we included the variant-specific sample programs and described the interpreter interface and environment. For the six hypothesis-testing tasks, we specified the black-box service protocol, input arity, and environment constraints but did not expose the protected implementation. For the two local-research tasks, we included extracted text from the two supplied papers and two supplied web pages. For the three \texttt{local\_research\_tex} tasks, we described the complete resource collection and included the relevant public \LaTeX{} sources. We excluded protected graders, hidden tests, black-box implementations, answer keys, and official solutions. 

The additional-information condition supplies substantially more information than the original prompt but does not recreate interactive services, executable files, or every detail of the original execution environment.

\paragraph{Effect on forecasts.}
We test whether the additional information changes the models' forecasts. For each estimator--task pair, we aggregate the 30 repetitions using the median point estimate. We then log-transform the task-level estimates and compare the original and additional-information conditions using paired $t$-tests across the same 16 tasks. The log transformation reflects the multiplicative scale of completion-time estimates. We adjust the three model-specific $p$-values using Holm's procedure.

Table~\ref{tab:self-estimation-context-effect} reports the results. Additional information generally shifts forecasts downward. Claude Opus 4.6's geometric-mean forecast across tasks decreases from 8.93 to 7.01 minutes, but the change is not statistically significant ($p=.113$). GPT-5.2's forecast decreases significantly, from 18.19 to 11.22 minutes (Holm-adjusted $p<.001$). The Gemini proxy's forecast decreases from 8.55 to 6.31 minutes; the difference is significant before correction ($p=.026$) but not after Holm correction ($p=.053$). Thus, after correction, only GPT-5.2 shows a statistically significant change in forecasts.

\begin{table}[t]
    \centering
    \caption{\textbf{Effect of additional task information on self-estimates.}
Forecasts are geometric means, in minutes, of task-level median predictions. Factor is the additional-information forecast divided by the original forecast. Confidence intervals are unadjusted 95\% intervals from paired $t$-tests on log forecasts; $p$-values use Holm correction across the three estimators.}
    \vspace{7pt}
    \label{tab:self-estimation-context-effect}
    \small
    \setlength{\tabcolsep}{3.5pt}
    \begin{tabular}{lcccc}
        \toprule
        \textbf{Estimator}
        & \textbf{Original}
        & \textbf{Additional}
        & \textbf{Factor (95\% CI)}
        & \textbf{Adj. $p$} \\
        \midrule
        Claude Opus 4.6
        & 8.93
        & 7.01
        & 0.785 [0.577, 1.067]
        & .113 \\

        GPT-5.2
        & 18.19
        & 11.22
        & 0.617 [0.507, 0.751]
        & $<.001$ \\

        Gemini proxy
        & 8.55
        & 6.31
        & 0.738 [0.567, 0.960]
        & .053 \\
        \bottomrule
    \end{tabular}
\end{table}

The downward shifts do not correct the original underestimation. Descriptively, GPT-5.2's median estimate-to-reference ratio decreases from 0.0052 to 0.0041, while the Gemini proxy's changes from 0.0239 to 0.0226 and Claude Opus 4.6's remains 0.627. Rank correlations remain non-positive in both conditions. Claude's 95\% reference containment increases from 10/15 to 12/15 tasks, whereas GPT and Gemini contain no references at any interval level in either condition. Overall, the additional information does not consistently improve the estimates or explain the underestimation observed in the main experiment.

\subsection{Cross-model task-duration estimation prompt}
\label{app:cross-model-prompt}

The cross-model experiment uses the following prompt. The placeholders \texttt{[TARGET AGENT CONFIGURATION]} and \texttt{[TASK INSTRUCTIONS]} are replaced with the target agent's profile and the complete HCAST task instructions, respectively. Each request is independent and contains no system prompt.

\begin{lstlisting}[
    basicstyle=\ttfamily\footnotesize,
    breaklines=true,
    breakindent=0pt,
    breakautoindent=false,
    breakatwhitespace=true,
    columns=flexible,
    keepspaces=false,
    frame=single,
    caption={Prompt used to elicit completion-time estimates for other LLM agents.},
    label={lst:cross-model-estimation-prompt}
]
Estimate how long the specified LLM agent would take to complete the task below. Estimate elapsed wall-clock time from the beginning of the agent run until successful task completion, including model generation, scaffolding runtime, tool use, investigation, implementation, testing, and debugging if applicable. Assume that the agent does not have human assistance and does not have access to any previous attempts at the task, and that they run independently in the supplied task environment and can use its provided tools, documentation, terminal, and tests. Estimate time conditional on successful task completion.

Your primary estimate is the agent's completion time of the task. Also provide 50%, 80%, and 95% uncertainty interval estimates. Do not search for or rely on remembered benchmark results.

TARGET AGENT CONFIGURATION
[TARGET AGENT CONFIGURATION]

TASK INSTRUCTIONS
[TASK INSTRUCTIONS]

ESTIMATION INSTRUCTIONS
Return exactly these values in this format:
estimated completion time (minutes): number
50% interval: lower value, upper value
80% interval: lower value, upper value
95% interval: lower value, upper value
rationale: 1-2 sentences explaining the reasoning behind the estimates.
\end{lstlisting}

For each target agent, the configuration block is rendered in the following format:

\begin{lstlisting}[
    basicstyle=\ttfamily\footnotesize,
    breaklines=true,
    columns=fullflexible,
    frame=single
]
Model: [MODEL DESCRIPTION]
Reasoning configuration: [REASONING CONFIGURATION]
Agent scaffold: [SCAFFOLD NAME]. [SCAFFOLD DESCRIPTION]
\end{lstlisting}

The target profiles contain only information about the model and its execution configuration. We do not expose observed completion times, success rates, token consumption, previous trajectories, or other outcome information.

\subsection{Additional results for estimating other LLM agents}
\label{app:cross-agent-results}

Available METR data constrain our target set. We include a target-agent/task pair only when the released data contain at least one successful run, so the number of evaluable tasks varies by target. Among targets with configurable reasoning effort, the released metadata record the reasoning configuration only for Claude 3.7 Sonnet and o1; we describe it as unrecorded for the others.



Table~\ref{tab:agent-reference-coverage} reports point-estimate bias and reference containment separately for each target agent, together with the aggregate results summarized in Section~\ref{sec:exp2}. The aggregate estimate-to-reference ratios correspond to observed completion times approximately $9\times$, $13\times$, and $49\times$ longer than the estimates from Claude Opus 5, GPT-5.6 Sol, and Gemini 3.7 Flash, respectively.

Underestimation predominates across target agents. The median estimate-to-reference ratio is below one for nearly every estimator--target combination; the principal exception is GPT-4 0314, which all three estimators overestimate at the median. Claude Opus 5 also slightly overestimates Claude 3.5 Sonnet 0620 and Claude Opus 4.5. The most severe underestimation occurs for o1, o3, GPT-5, GPT-5.1-Codex-Max, GPT-5.2, GPT-5.3-Codex, and Gemini 3 Pro: for every estimator, the median ratio is below 0.03 and the median 95\% intervals contain none of the references for these targets.

\begin{sidewaystable*}[p]
\centering
\caption{\textbf{Point-estimate bias and reference containment for other LLM agents.}
For each target, $n$ is the number of tasks with at least one successful released METR run. Ratio is the median task-level estimate-to-reference ratio; values below one indicate underestimation. Containment entries report the percentage and count of tasks for which the geometric mean of successful METR run times falls within the median forecast interval. ``Unknown'' indicates that the reasoning configuration was not recorded in the available METR metadata or specified to the estimators; ``None'' denotes a model without a configurable reasoning-effort setting.}
\vspace{7pt}
\label{tab:agent-reference-coverage}
\tiny
\setlength{\tabcolsep}{1.4pt}
\renewcommand{\arraystretch}{1.08}
\resizebox{\textheight}{!}{%
\begin{tabular}{llll
rrrr
rrrr
rrrr}
\toprule
& & & &
\multicolumn{4}{c}{Claude Opus 5} &
\multicolumn{4}{c}{GPT-5.6 Sol} &
\multicolumn{4}{c}{Gemini 3.7 Flash} \\
\cmidrule(lr){5-8}
\cmidrule(lr){9-12}
\cmidrule(lr){13-16}
Target agent & Reasoning & Scaffold & $n$
& Ratio & 50\% & 80\% & 95\%
& Ratio & 50\% & 80\% & 95\%
& Ratio & 50\% & 80\% & 95\% \\
\midrule
Claude 3 Opus
& None & ReAct & 7
& 0.646 & 28.6 (2/7) & 57.1 (4/7) & 85.7 (6/7)
& 0.287 & 0.0 (0/7) & 28.6 (2/7) & 57.1 (4/7)
& 0.112 & 28.6 (2/7) & 28.6 (2/7) & 28.6 (2/7) \\

Claude 3.5 Sonnet 0620
& None & ReAct & 12
& 1.020 & 41.7 (5/12) & 50.0 (6/12) & 66.7 (8/12)
& 0.462 & 8.3 (1/12) & 25.0 (3/12) & 75.0 (9/12)
& 0.185 & 8.3 (1/12) & 25.0 (3/12) & 33.3 (4/12) \\

Claude 3.5 Sonnet 1022
& None & ReAct & 13
& 0.542 & 7.7 (1/13) & 46.2 (6/13) & 69.2 (9/13)
& 0.241 & 23.1 (3/13) & 38.5 (5/13) & 46.2 (6/13)
& 0.094 & 0.0 (0/13) & 0.0 (0/13) & 23.1 (3/13) \\

Claude 3.7 Sonnet
& Extended (32k) & ReAct & 16
& 0.415 & 6.2 (1/16) & 31.2 (5/16) & 75.0 (12/16)
& 0.194 & 6.2 (1/16) & 25.0 (4/16) & 31.2 (5/16)
& 0.086 & 0.0 (0/16) & 12.5 (2/16) & 25.0 (4/16) \\

Claude 4 Opus
& Unknown & ReAct & 16
& 0.679 & 43.8 (7/16) & 68.8 (11/16) & 81.2 (13/16)
& 0.344 & 18.8 (3/16) & 31.2 (5/16) & 68.8 (11/16)
& 0.179 & 6.2 (1/16) & 12.5 (2/16) & 25.0 (4/16) \\

Claude 4.1 Opus
& Unknown & ReAct & 16
& 0.053 & 12.5 (2/16) & 12.5 (2/16) & 12.5 (2/16)
& 0.024 & 6.2 (1/16) & 6.2 (1/16) & 25.0 (4/16)
& 0.010 & 0.0 (0/16) & 0.0 (0/16) & 0.0 (0/16) \\

Claude Opus 4.5
& Unknown & ReAct & 15
& 1.114 & 40.0 (6/15) & 46.7 (7/15) & 60.0 (9/15)
& 0.771 & 46.7 (7/15) & 53.3 (8/15) & 53.3 (8/15)
& 0.288 & 13.3 (2/15) & 26.7 (4/15) & 46.7 (7/15) \\

Claude Opus 4.6
& Unknown & ReAct & 15
& 0.503 & 20.0 (3/15) & 66.7 (10/15) & 100.0 (15/15)
& 0.335 & 0.0 (0/15) & 40.0 (6/15) & 86.7 (13/15)
& 0.132 & 13.3 (2/15) & 13.3 (2/15) & 26.7 (4/15) \\
\midrule

GPT-4 0314
& None & Modular & 8
& 9.220 & 12.5 (1/8) & 12.5 (1/8) & 12.5 (1/8)
& 4.432 & 12.5 (1/8) & 12.5 (1/8) & 25.0 (2/8)
& 1.740 & 25.0 (2/8) & 50.0 (4/8) & 50.0 (4/8) \\

GPT-4 1106
& None & ReAct & 9
& 0.823 & 55.6 (5/9) & 55.6 (5/9) & 66.7 (6/9)
& 0.398 & 11.1 (1/9) & 22.2 (2/9) & 66.7 (6/9)
& 0.158 & 0.0 (0/9) & 11.1 (1/9) & 22.2 (2/9) \\

GPT-4 Turbo
& None & ReAct & 7
& 0.186 & 14.3 (1/7) & 14.3 (1/7) & 28.6 (2/7)
& 0.077 & 14.3 (1/7) & 14.3 (1/7) & 14.3 (1/7)
& 0.031 & 0.0 (0/7) & 0.0 (0/7) & 0.0 (0/7) \\

GPT-4o
& None & ReAct & 10
& 0.737 & 20.0 (2/10) & 60.0 (6/10) & 80.0 (8/10)
& 0.333 & 0.0 (0/10) & 20.0 (2/10) & 70.0 (7/10)
& 0.136 & 0.0 (0/10) & 20.0 (2/10) & 40.0 (4/10) \\

o1-preview
& Unknown & Duet & 14
& 0.823 & 78.6 (11/14) & 92.9 (13/14) & 100.0 (14/14)
& 0.521 & 14.3 (2/14) & 78.6 (11/14) & 100.0 (14/14)
& 0.185 & 0.0 (0/14) & 0.0 (0/14) & 0.0 (0/14) \\

o1
& Medium (default) & Triframe & 15
& 0.013 & 0.0 (0/15) & 0.0 (0/15) & 0.0 (0/15)
& 0.006 & 0.0 (0/15) & 0.0 (0/15) & 0.0 (0/15)
& 0.003 & 0.0 (0/15) & 0.0 (0/15) & 0.0 (0/15) \\

o3
& Unknown & Triframe & 16
& 0.010 & 0.0 (0/16) & 0.0 (0/16) & 0.0 (0/16)
& 0.005 & 0.0 (0/16) & 0.0 (0/16) & 0.0 (0/16)
& 0.002 & 0.0 (0/16) & 0.0 (0/16) & 0.0 (0/16) \\

GPT-5
& Unknown & Triframe & 16
& 0.010 & 0.0 (0/16) & 0.0 (0/16) & 0.0 (0/16)
& 0.006 & 0.0 (0/16) & 0.0 (0/16) & 0.0 (0/16)
& 0.002 & 0.0 (0/16) & 0.0 (0/16) & 0.0 (0/16) \\

GPT-5.1-Codex-Max
& Unknown & Triframe & 16
& 0.013 & 0.0 (0/16) & 0.0 (0/16) & 0.0 (0/16)
& 0.008 & 0.0 (0/16) & 0.0 (0/16) & 0.0 (0/16)
& 0.003 & 0.0 (0/16) & 0.0 (0/16) & 0.0 (0/16) \\

GPT-5.2
& Unknown & Triframe & 16
& 0.004 & 0.0 (0/16) & 0.0 (0/16) & 0.0 (0/16)
& 0.002 & 0.0 (0/16) & 0.0 (0/16) & 0.0 (0/16)
& 0.001 & 0.0 (0/16) & 0.0 (0/16) & 0.0 (0/16) \\

GPT-5.3-Codex
& Unknown & Triframe & 16
& 0.014 & 0.0 (0/16) & 0.0 (0/16) & 0.0 (0/16)
& 0.008 & 0.0 (0/16) & 0.0 (0/16) & 0.0 (0/16)
& 0.003 & 0.0 (0/16) & 0.0 (0/16) & 0.0 (0/16) \\
\midrule

Gemini 3 Pro
& Unknown & ReAct & 16
& 0.026 & 0.0 (0/16) & 0.0 (0/16) & 0.0 (0/16)
& 0.022 & 0.0 (0/16) & 0.0 (0/16) & 0.0 (0/16)
& 0.006 & 0.0 (0/16) & 0.0 (0/16) & 0.0 (0/16) \\
\midrule

\textbf{All eligible pairs}
& --- & --- & 269
& \textbf{0.111} & \textbf{17.5} & \textbf{28.6} & \textbf{39.0}
& \textbf{0.075} & \textbf{7.8} & \textbf{19.0} & \textbf{33.5}
& \textbf{0.020} & \textbf{3.7} & \textbf{8.2} & \textbf{14.1} \\
\bottomrule
\end{tabular}%
}
\end{sidewaystable*}

These target-level comparisons are descriptive. The available agents differ jointly in model generation and scaffold, while METR's released metadata identify the reasoning configuration for only Claude 3.7 Sonnet and o1 among the reasoning-capable targets. Differences between targets therefore cannot be attributed solely to model identity or generation.

\subsection{WildChat filtering and sampling}
\label{app:wildchat-sampling}

We extracted the first user request from each conversation in
WildChat-1M~\citep{zhao2024wildchat}. We excluded conversations and first user messages that were flagged non-English, toxic, or redacted in WildChat's metadata. After trimming whitespace, we also only retained requests that contained at least three words and between 3 and 10,000 characters. These filters retained
441,109 eligible requests from the 837,989 conversations in the original dataset. We sampled 500 requests uniformly without
replacement using reservoir sampling with random seed 42.

\subsection{Wildchat response-time estimation prompt}
\label{app:wildchat-estimation-prompt}

The WildChat response-time experiment uses the following prompt. The placeholders
\texttt{[API NAME]} and \texttt{[USER QUERY]} are replaced with the corresponding
provider name and sampled first-user request. Each estimate request is independent,
begins a fresh conversation, and contains no system prompt.

\begin{lstlisting}[
    basicstyle=\ttfamily\footnotesize,
    breaklines=true,
    breakindent=0pt,
    breakautoindent=false,
    breakatwhitespace=true,
    columns=flexible,
    keepspaces=false,
    frame=single,
    caption={Prompt used to elicit response-time estimates.},
    label={lst:wildchat-estimation-prompt}
]
Estimate how many seconds it would take for you to respond to the following user query. The user query is sent through the [API NAME] API. In your estimate, include the latency from the user to the API and back. The user query is the start of a new conversation. Return exactly one number. Do not explain your answer.

User query:
[USER QUERY]
\end{lstlisting}

\subsection{Additional results for Wildchat response-time estimation}
\label{app:wildchat-results}

We uniformly sample 500 eligible first-user requests from WildChat-1M. After each
estimate call, we send the original request verbatim in a separate fresh conversation
and record wall-clock task-call latency. We repeat this paired procedure five times
for every model--request pair. For each pair $(M,i)$, we separately take the median
estimated and observed latency across valid repetitions, denoted $\hat t_{M,i}$ and
$t_{M,i}$. Defining $r_{M,i}=\hat t_{M,i}/t_{M,i}$, the bias factor is
$\exp\left(N_M^{-1}\sum_i\log r_{M,i}\right)$ and the absolute error factor is
$\exp\left(N_M^{-1}\sum_i|\log r_{M,i}|\right)$, where $N_M$ is the number of
requests with valid paired medians for model $M$. Table~\ref{tab:wildchat-time-estimation}
reports the complete results.

\begin{table}[t]
\centering
\small
\caption{\textbf{WildChat response-time estimation results.} A bias factor below one
indicates underestimation; the absolute error factor summarizes error magnitude
irrespective of direction. Within-$2\times$ and within-$4\times$ report the
proportions of valid requests in the corresponding multiplicative ranges.}
\label{tab:wildchat-time-estimation}
\begin{tabular}{lrrrrrr}
\toprule
Model & $N_M$ & Bias factor & Abs. error factor & Within $2\times$ & Within $4\times$ & $\rho$ \\
\midrule
Claude Opus 5    & 497 & 0.94 & 1.96 & 52.5\% & 95.6\% & 0.61 \\
GPT-5.6 Sol      & 500 & 0.70 & 1.61 & 73.4\% & 96.8\% & 0.86 \\
Gemini 3.7 Flash & 498 & 0.50 & 2.04 & 46.2\% & 99.4\% & 0.82 \\
\bottomrule
\end{tabular}
\end{table}

\textbf{Failures and exclusions.}
Among 7,500 scheduled trials, 7,465 produce ordinary completed responses, 34 end in
deterministic policy outcomes, and one produces a blank final response. Policy
outcomes and the blank response are excluded from the accuracy calculations.

%% file: Appendices/exp4.tex
\section{Details for experiment 3: duration judgments}
\label{app:duration}

\begin{table}[t]
\centering
\caption{Duration judgments on 5000 UDS-T events. $\rho$: Spearman correlation with the mean human label. Bias: mean model label minus mean human label, in scale steps. Gain: sensitivity to sentence features relative to people ($g$; defined in Section~5), over all features and over the predicate and context groups. Brackets give 95\% bootstrap intervals. Qwen2.5-32B was run on the 1000-event subset only.}
\vspace{7pt}
\label{tab:exp4-duration}
\small
\setlength{\tabcolsep}{4pt}
\begin{tabular}{lccccc}
\toprule
& & & \multicolumn{3}{c}{Sensitivity gain} \\
\cmidrule(lr){4-6}
Judge & $\rho$ & Bias & All features & Predicate & Context \\
\midrule
One annotator (reference) & 0.49 & -- & 1 & 1 & 1 \\
Claude Opus~5 & 0.688~{\scriptsize[0.67, 0.71]} & +0.49 & 1.35~{\scriptsize[1.22, 1.47]} & 1.62~{\scriptsize[1.50, 1.76]} & 0.89~{\scriptsize[0.69, 1.11]} \\
GPT-5.6 Sol & 0.619~{\scriptsize[0.60, 0.64]} & +0.01 & 1.27~{\scriptsize[1.13, 1.39]} & 1.57~{\scriptsize[1.40, 1.72]} & 0.76~{\scriptsize[0.52, 0.99]} \\
Gemini~3.7 Flash & 0.632~{\scriptsize[0.61, 0.65]} & +0.50 & 1.30~{\scriptsize[1.15, 1.42]} & 1.66~{\scriptsize[1.49, 1.83]} & 0.69~{\scriptsize[0.47, 0.92]} \\
Qwen2.5-32B & 0.430 & -0.91 & 0.67~{\scriptsize[0.30, 1.03]} & 1.13~{\scriptsize[0.75, 1.51]} & 0.37~{\scriptsize[-0.17, 0.90]} \\
\bottomrule
\end{tabular}
\end{table}

\subsection{Prompts}
\label{app:duration-prompts}
Each prompt reproduces the instructions given to the human participants. The API models returned structured JSON through each provider's schema-constrained output mode. Qwen2.5-32B, served through vLLM, received the same prompt followed by an explicit instruction to answer in JSON with the corresponding fields; answers were parsed with one retry on a malformed response.

\begin{lstlisting}[
    basicstyle=\ttfamily\footnotesize,
    breaklines=true,
    breakindent=0pt,
    breakautoindent=false,
    breakatwhitespace=true,
    columns=flexible,
    keepspaces=false,
    frame=single,
    caption={UDS-T, base prompt.}
]
You will read a sentence from English web text (for example a blog, review, email, or forum post). One word in the sentence is marked with [[double brackets]]. It refers to an event or state.

What is the likely duration of the event or state referred to by the marked word?

Choose exactly one of: instant, seconds, minutes, hours, days, weeks, months, years, decades, centuries, forever.

Then say how confident you are in your duration answer:
0 = not at all confident
1 = not very confident
2 = somewhat confident
3 = very confident
4 = totally confident

Sentence: {sentence}
\end{lstlisting}

\paragraph{Prompt variants.}
\emph{Reworded} rephrases every instruction while keeping the scale. \emph{Reversed} lists the categories from \textit{forever} to \textit{instant}. \emph{Numeric} presents the scale as numbered points and asks for a number from 0 to 10, which is mapped back to its category. The reworded and numeric prompts are:
\begin{lstlisting}[
    basicstyle=\ttfamily\footnotesize,
    breaklines=true,
    breakindent=0pt,
    breakautoindent=false,
    breakatwhitespace=true,
    columns=flexible,
    keepspaces=false,
    frame=single,
    caption={}
]
Below is a sentence taken from text people wrote online, such as reviews, emails, blog
posts, or forum threads. The word inside [[double brackets]] names an event or a state.

How long does that event or state most likely last?

Pick one answer from this list: instant, seconds, minutes, hours, days, weeks, months,
years, decades, centuries, forever.

Also rate your confidence in that answer, from 0 to 4:
0 = not at all confident
1 = not very confident
2 = somewhat confident
3 = very confident
4 = totally confident

Sentence: {sentence}
\end{lstlisting}

\begin{lstlisting}[
    basicstyle=\ttfamily\footnotesize,
    breaklines=true,
    breakindent=0pt,
    breakautoindent=false,
    breakatwhitespace=true,
    columns=flexible,
    keepspaces=false,
    frame=single,
    caption={}
]
You will read a sentence from English web text (for example a blog, review, email, or forum
post). One word in the sentence is marked with [[double brackets]]. It refers to an event or
state.

What is the likely duration of the event or state referred to by the marked word? Answer
with a number on this scale:
0 = instant
1 = seconds
2 = minutes
3 = hours
4 = days
5 = weeks
6 = months
7 = years
8 = decades
9 = centuries
10 = forever

Then say how confident you are in your duration answer:
0 = not at all confident
1 = not very confident
2 = somewhat confident
3 = very confident
4 = totally confident

Sentence: {sentence}
\end{lstlisting}

\paragraph{Wittenberg \& Levy, Experiment 1a.}
The instructions are those given to participants in the original study.
\begin{lstlisting}[
    basicstyle=\ttfamily\footnotesize,
    breaklines=true,
    breakindent=0pt,
    breakautoindent=false,
    breakatwhitespace=true,
    columns=flexible,
    keepspaces=false,
    frame=single,
    caption={}
]
In this survey, we would like your opinion on how long you think events take. For example,
the event "Frida is eating an apple" might take about seven minutes; "Frida is driving 150
miles" might take about 2 hours 15 minutes; or "Frida sneezed" might take about one second.

Please read the following sentence carefully and estimate how long the described event took
as best and accurately as you can. Give your estimate as hours, minutes, and seconds.

Sentence: {sentence}
\end{lstlisting}

\subsection{Models, settings and cost}
We use default settings for all models. Neither Claude Opus~5 nor GPT-5.6 Sol accepts a temperature parameter, and we use temperature of 0 for Gemini. Qwen2.5-32B-Instruct (4-bit AWQ) was served with vLLM in fp16 on two NVIDIA TITAN RTX GPUs using the server's default sampling. Refusal fallbacks were disabled so that every answer comes from the named model. Table~\ref{tab:duration-usage} lists calls, tokens and cost for all runs.

\begin{table}[h]
\centering
\caption{API usage for all duration runs: the full sample, four additional runs and three prompt variants on the 1000-event subset, and pilot calls.}
\vspace{7pt}
\label{tab:duration-usage}
\small
\begin{tabular}{lrrr}
\toprule
Model & Calls & Tokens (in / out) & Cost \\
\midrule
Claude Opus~5 & 14315 & 7.13M / 0.21M & \$40.99 \\
GPT-5.6 Sol & 12202 & 2.87M / 0.69M & \$25.17 \\
Gemini~3.7 Flash & 12211 & 2.01M / 1.72M & \$7.95 \\
Qwen2.5-32B & 1042 & -- & local \\
\midrule
Total & & & \$74.11 \\
\bottomrule
\end{tabular}
\end{table}

\subsection{Item sampling and feature definitions}
\label{app:exp3_sampling}
The 5000 events are a uniform random sample without replacement from the 25192 UDS-T predicates with at least three duration labels, pooling the train, development and test splits. They were drawn in two stages: a first sample of 1000 events (seed 0), which forms the subset used for repeated runs and prompt variants, and 4000 further events from the remaining pool (seed 1). Each event is shown as its full sentence with the predicate word marked by double brackets. Features were read off the Universal Dependencies parse of the UD English Web Treebank (v1.2) on which UDS-T is annotated:
\begin{itemize}
\item \emph{Predicate form}: a non-verbal predicate (adjective, noun or copular construction); otherwise the verb form, determined by its tag and auxiliaries (simple past as the reference, simple present, perfect \textit{have}+participle, progressive \textit{be}+gerund, imperative mood). Modal, future, infinitival and participial forms are included as further levels of the same factor but are not reported.
\item \emph{Context}: the subject type (first- or second-person pronoun as the reference; third-person pronoun, singular noun, plural noun, proper noun, or no subject); a habitual adverb modifying the predicate (\textit{always}, \textit{usually}, \textit{often}, \textit{never}, \ldots); an \textit{if} marker on the predicate's clause; passive voice. Object type, negation, clause type (main, subordinate, conjunct) and question status are included as controls.
\end{itemize}
Duration phrases (\textit{for years}), \textit{every}-phrases and point-in-time adverbials were too rare to estimate reliably and were dropped. All regressions use ordinary least squares with heteroskedasticity-robust (HC1) standard errors; the human regression uses the mean label per event.

\subsection{Per-feature effects}
\label{app:duration-detail}
The two largest divergences in Figure~\ref{fig:exp4-duration}a are the predicate features. Describing an event as a state rather than an action raises the human estimate by 1.38 steps and the models' by 2.69--3.00; conversely a habitual adverb (\textit{always}, \textit{usually}) raises the human estimate by 0.90 steps but the models' by at most 0.36. The predicate gain exceeds 1 for all three API models, and the context gain falls below 1 for GPT-5.6 Sol and Gemini~3.7 Flash (Claude Opus~5: 0.89~[0.69, 1.11]). Because the predicate and context groups were defined after inspecting the per-feature effects, we treat that split as exploratory.

Table~\ref{tab:duration-cues} gives every feature effect underlying Figure~\ref{fig:exp4-duration}a. Standard errors for the models are bootstrap standard deviations. Qwen2.5-32B was estimated on the 1000-event subset, so the last column gives the human effects on that subset for comparison.

\begin{figure}[h]
\centering
\includegraphics[width=.82\linewidth]{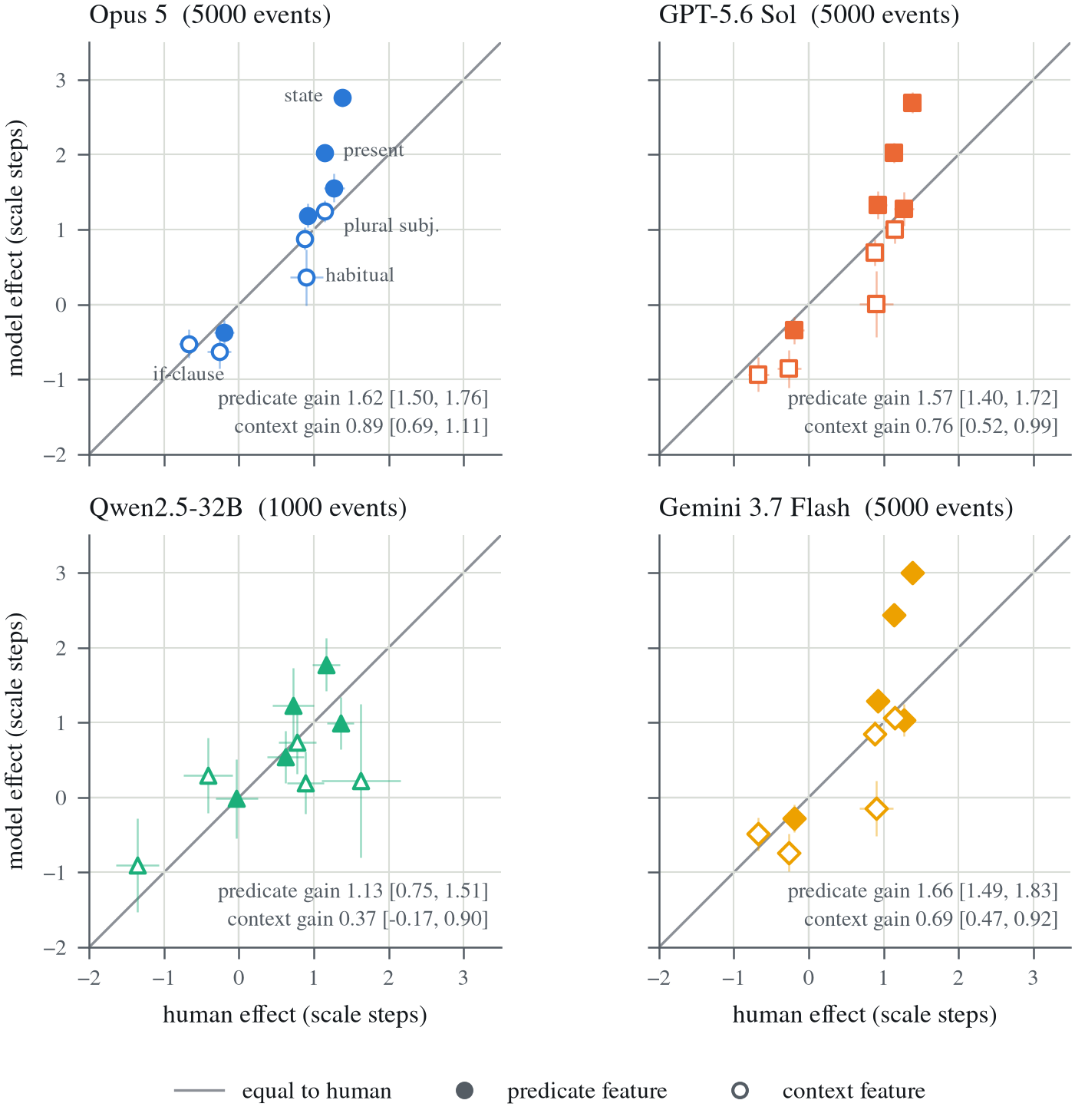}
\vspace{-5pt}
\caption{Per-model version of Figure~\ref{fig:exp4-duration}a: human effect ($x$) against model effect ($y$) for each of the ten features, one panel per model, with $\pm 1$ SE. Filled markers are predicate features, open markers context features. Qwen2.5-32B is estimated on the 1000-event subset.}
\vspace{-3pt}
\label{fig:duration-cues-per-model}
\end{figure}

\begin{table}[h]
\centering
\caption{Effect of each feature on the duration label, in scale steps (standard error in parentheses). Upper block: predicate features; lower block: context features.}
\vspace{7pt}
\label{tab:duration-cues}
\small
\setlength{\tabcolsep}{3pt}
\begin{tabular}{lcccccc}
\toprule
Feature & Humans (5k) & Claude Opus~5 & GPT-5.6 Sol & Gemini 3.7 & Qwen 2.5 & Humans (1k) \\
\midrule
Non-verbal pred. & +1.38 (0.09) & +2.76 (0.12) & +2.69 (0.14) & +3.00 (0.13) & +1.77 (0.35) & +1.16 (0.18) \\
Simple present & +1.14 (0.08) & +2.03 (0.11) & +2.02 (0.13) & +2.44 (0.12) & +0.99 (0.35) & +1.36 (0.18) \\
Perfect & +1.27 (0.14) & +1.56 (0.19) & +1.28 (0.23) & +1.03 (0.22) & +1.23 (0.50) & +0.72 (0.28) \\
Progressive & +0.92 (0.12) & +1.19 (0.16) & +1.33 (0.18) & +1.29 (0.16) & +0.54 (0.35) & +0.62 (0.25) \\
Imperative & -0.20 (0.14) & -0.37 (0.16) & -0.35 (0.18) & -0.28 (0.18) & -0.02 (0.53) & -0.03 (0.28) \\
\midrule
Plural-noun subj. & +1.15 (0.11) & +1.25 (0.14) & +1.00 (0.18) & +1.06 (0.17) & +0.20 (0.42) & +0.89 (0.24) \\
Named subject & +0.88 (0.11) & +0.88 (0.15) & +0.69 (0.17) & +0.85 (0.17) & +0.74 (0.43) & +0.78 (0.25) \\
Habitual adverb & +0.90 (0.22) & +0.36 (0.37) & +0.00 (0.44) & -0.15 (0.37) & +0.22 (1.02) & +1.63 (0.53) \\
\textit{If}-clause & -0.67 (0.13) & -0.52 (0.19) & -0.93 (0.23) & -0.49 (0.22) & -0.90 (0.63) & -1.36 (0.29) \\
Passive & -0.26 (0.16) & -0.63 (0.23) & -0.86 (0.25) & -0.74 (0.25) & +0.29 (0.50) & -0.42 (0.33) \\
\bottomrule
\end{tabular}
\end{table}

\subsection{Robustness to the prompt}
\label{app:exp3_robustness}
Across the five runs on the 1000-event subset, the pooled sensitivity gain varies by at most 0.08 (SD). Under the three prompt variants, predicate gains remain within 1.61--1.99 and context gains within 0.45--0.79 (Table~\ref{tab:duration-prompts}).

\begin{table}[h]
\centering
\caption{Agreement and sensitivity gain under each prompt, on the 1000-event subset. Base values average five runs; each variant is a single run.}
\vspace{6pt}
\label{tab:duration-prompts}
\small
\begin{tabular}{llcccc}
\toprule
Model & Prompt & $\rho$ & Gain (all) & Gain (predicate) & Gain (context) \\
\midrule
Claude Opus~5 & base (mean of 5) & 0.676 & 1.16 & 1.82 & 0.72 \\
 & reworded & 0.651 & 1.09 & 1.85 & 0.57 \\
 & reversed & 0.679 & 1.14 & 1.87 & 0.64 \\
 & numeric & 0.653 & 1.12 & 1.85 & 0.62 \\
\midrule
GPT-5.6 Sol & base (mean of 5) & 0.591 & 1.05 & 1.73 & 0.58 \\
 & reworded & 0.608 & 0.97 & 1.64 & 0.52 \\
 & reversed & 0.600 & 1.12 & 1.61 & 0.79 \\
 & numeric & 0.613 & 0.98 & 1.76 & 0.45 \\
\midrule
Gemini~3.7 Flash & base (mean of 5) & 0.634 & 1.15 & 1.94 & 0.62 \\
 & reworded & 0.625 & 1.13 & 1.95 & 0.57 \\
 & reversed & 0.636 & 1.19 & 1.99 & 0.65 \\
 & numeric & 0.627 & 1.04 & 1.91 & 0.45 \\
\bottomrule
\end{tabular}
\end{table}

\begin{figure}[h]
\centering
\includegraphics[width=\linewidth]{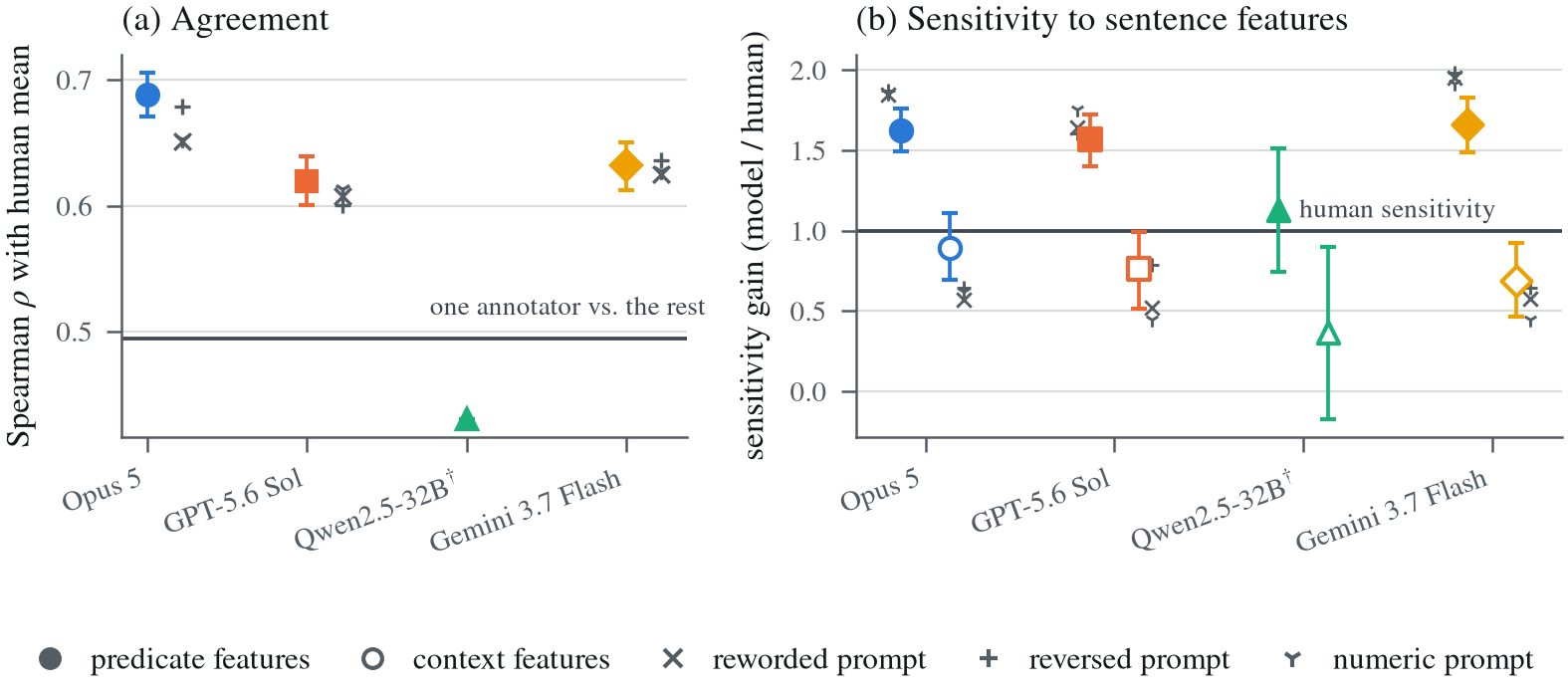}
\vspace{6pt}
\caption{(a) Correlation between each model's labels and the mean human label, with the single-annotator reference. (b) Sensitivity gain by feature group with 95\% intervals; small gray markers give the three prompt variants. Both panels use the 1000-event subset, where the repeated runs and prompt variants were collected.}
\label{fig:duration-agreement}
\end{figure}

\subsection{Run stability and self-consistency}
We report the run stability in Table~\ref{tab:duration-stability}. The left column states the model or subject, and we report same category agreement, mean difference, and standard deviations of $\rho$ and the pooled sensitivity gain. 

\begin{table}[h]
\centering
\caption{Agreement between two annotators on the same event, and between two runs of the same model, on the 1000-event subset. The last two columns give the standard deviation of $\rho$ and of the pooled sensitivity gain across five runs.}
\vspace{6pt}
\label{tab:duration-stability}
\small
\begin{tabular}{lcccc}
\toprule
& Same category & Mean difference (steps) & SD of $\rho$ & SD of gain \\
\midrule
Two annotators & 29.2\% & 1.82 & -- & -- \\
Claude Opus~5, two runs & 92.2\% & 0.15 & 0.004 & 0.058 \\
GPT-5.6 Sol, two runs & 80.4\% & 0.45 & 0.007 & 0.080 \\
Gemini~3.7 Flash, two runs & 89.4\% & 0.20 & 0.008 & 0.035 \\
\bottomrule
\end{tabular}
\end{table}

\subsection{Use of the duration scale}
Models also differ from people in which parts of the scale they use (Table~\ref{tab:duration-shares}). Claude Opus~5 assigns \textit{years} to 26\% of events against 13\% for people, and Qwen2.5-32B assigns \textit{instant} to 27\% against 6\%.

\begin{table}[h]
\centering
\caption{Share of labels (\%) in each duration category. Humans and the API models: 5000 events. Qwen2.5-32B$^\dagger$: 1000-event subset.}
\vspace{6pt}
\label{tab:duration-shares}
\scriptsize
\begin{tabular}{lccccccccccc}
\toprule
& inst. & sec. & min. & hours & days & weeks & months & years & dec. & cent. & forever \\
\midrule
Humans & 6.0 & 8.6 & 27.7 & 11.5 & 11.1 & 6.5 & 10.2 & 12.5 & 2.6 & 0.7 & 2.6 \\
Claude Opus~5 & 2.4 & 6.6 & 27.2 & 11.5 & 8.8 & 4.8 & 8.1 & 26.3 & 2.0 & 0.4 & 2.0 \\
GPT-5.6 Sol & 9.8 & 13.1 & 21.0 & 10.3 & 7.7 & 2.9 & 8.5 & 22.7 & 1.0 & 0.4 & 2.6 \\
Gemini~3.7 Flash & 5.1 & 10.5 & 19.1 & 11.3 & 10.0 & 4.2 & 11.7 & 20.6 & 2.7 & 0.9 & 4.0 \\
Qwen2.5-32B$^\dagger$ & 27.2 & 5.1 & 34.5 & 4.6 & 7.7 & 1.0 & 3.2 & 9.7 & 0.4 & 0.5 & 6.1 \\
\bottomrule
\end{tabular}
\end{table}

\subsection{Human judgments on the full corpus}
To characterize the human judgments independently of the model comparison, we fit the same kind of regression to all 163338 UDS-T labels with annotator fixed effects and standard errors clustered by sentence. Annotator identity alone explains $R^2=0.107$ of the variance; adding the grammatical features raises it to 0.184, adding verb identity (lemmas with at least 30 events) instead raises it to 0.181, and adding both to 0.229. Holding grammar fixed, the verbs judged shortest denote brief acts (\textit{thank}~$-1.6$, \textit{sign}~$-1.5$, \textit{attach}~$-1.5$, \textit{guess}~$-1.5$), and those judged longest denote states and gradual changes (\textit{live}~$+2.9$, \textit{grow}~$+2.5$, \textit{remain}~$+2.2$, \textit{support}~$+2.1$).

\subsection{Wittenberg \& Levy items}
\label{app:duration-wl}
On the minimal pairs of \citet{wittenberg2017quick}, people judge the light-verb version as shorter for all three event classes: 0.46$\times$ for punctive events, 0.75$\times$ in mass syntax and 0.85$\times$ in count syntax. The API models shorten punctive (0.50--0.61$\times$) and mass-syntax events (0.25--0.64$\times$), but lengthen count-syntax events (1.26--1.46$\times$), the class for which human estimates barely moved. Because these items hold the event fixed, the divergence cannot be attributed to different beliefs about the events themselves.

Light-verb sentences were built from each item's simple-verb sentence (Appendix~A of \citet{wittenberg2017quick}) using the noun forms in their Table~1, following their example (\textit{After their first date, Douglas kissed Mary} $\rightarrow$ \textit{\ldots Douglas gave a kiss to Mary}). Four of their 24 events (\textit{condoling}, \textit{punching}, \textit{respecting}, \textit{sermoning}) have no sentence in the published appendix and were excluded. Human values are medians of the released per-participant estimates; model values are medians over five runs (Qwen2.5-32B: one run). Ratios below 1 mean the light-verb version was judged shorter.

\begin{figure}[h]
\centering
\includegraphics[width=0.95\linewidth]{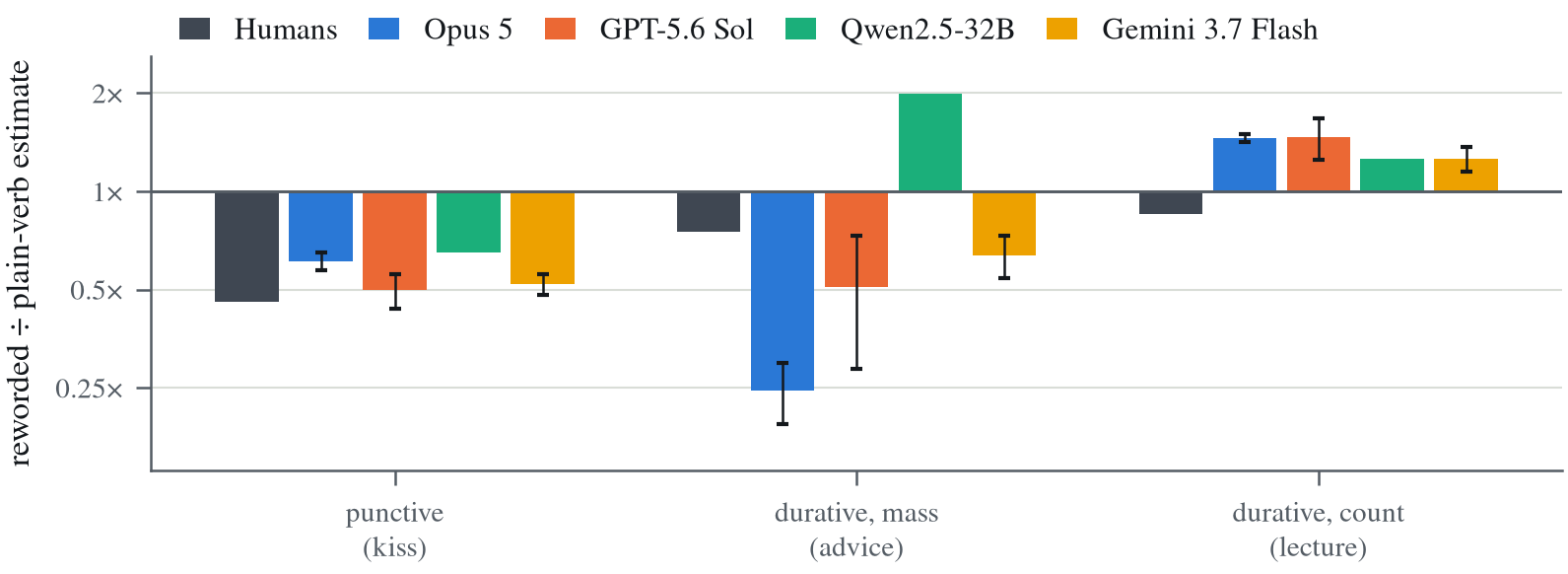}
\caption{Per-model version of Figure~\ref{fig:exp4-duration}c, including Qwen2.5-32B: light-verb estimate divided by simple-verb estimate, as a geometric mean over the events of each class (log scale). Error bars: SD across five runs.}
\label{fig:duration-wl}
\end{figure}

\begin{table}[h]
\centering
\caption{Light-verb / simple-verb duration ratio for each Wittenberg \& Levy event.}
\vspace{7pt}
\label{tab:duration-wl}
\scriptsize
\begin{tabular}{llccccc}
\toprule
Event & Class & Humans & Claude Opus~5 & GPT-5.6 Sol & Gemini~3.7 Flash & Qwen2.5-32B \\
\midrule
embracing & punctive & 0.50 & 0.80 & 1.00 & 0.83 & 1.67 \\
hugging & punctive & 1.21 & 1.00 & 1.00 & 1.00 & 1.00 \\
cuddling & punctive & 0.10 & 0.07 & 0.03 & 0.03 & 0.30 \\
kicking & punctive & 0.67 & 1.00 & 0.67 & 1.00 & 1.00 \\
shaking & punctive & 0.41 & 0.67 & 0.50 & 1.00 & 0.17 \\
poking & punctive & 0.40 & 1.00 & 1.00 & 1.00 & 1.00 \\
kissing & punctive & 0.67 & 1.00 & 0.40 & 0.60 & 0.60 \\
\midrule
advising & durative, mass & 0.57 & 0.67 & 0.50 & 1.00 & 0.50 \\
recognizing & durative, mass & 0.67 & 0.17 & 1.00 & 0.67 & 1.00 \\
encouraging & durative, mass & 0.67 & 1.00 & 0.50 & 1.00 & 1.00 \\
assuring & durative, mass & 0.77 & 1.00 & 3.00 & 2.00 & 1.00 \\
assisting & durative, mass & 0.69 & 0.33 & 0.12 & 0.06 & 0.50 \\
thanking & durative, mass & 1.00 & 1.67 & 4.00 & 1.00 & 1.67 \\
supporting & durative, mass & 1.00 & 0.00 & 0.33 & 0.49 & 300.00 \\
\midrule
addressing & durative, count & 1.20 & 1.00 & 1.00 & 1.00 & 2.00 \\
lecturing & durative, count & 1.02 & 1.25 & 1.00 & 1.00 & 0.50 \\
scolding & durative, count & 1.00 & 1.00 & 1.00 & 1.00 & 2.00 \\
presenting & durative, count & 0.30 & 1.67 & 3.00 & 3.00 & 1.00 \\
speaking & durative, count & 1.42 & 1.20 & 1.00 & 1.17 & 2.00 \\
talking & durative, count & 0.75 & 3.00 & 1.50 & 1.00 & 1.00 \\
\bottomrule
\end{tabular}
\end{table}

\subsection{Refusals and failures}
Claude Opus~5 declined one UDS-T sentence, which mentions the Ames strain of anthrax, in every run (\texttt{stop\_reason=refusal}); it is excluded from that model's statistics.

%% file: Appendices/exp3.tex
\clearpage
\section{Details for Experiment 4: decisions involving time}

\subsection{Prompts}
\label{app:exp3_prompts}
Placeholders in braces are substituted at run time. \texttt{\{question\}} is replaced with the task text; \texttt{\{tool\_description\}} is replaced with the tool block in Appendix~\ref{app:exp4-tools}; \texttt{\{tool\_delay\_seconds\}} is replaced with the stated time cost $\Tilde{t}$; \texttt{\{reasoning\_level\}} is replaced with the model's default reasoning level; \texttt{\{calculator\_query\}} and \texttt{\{calculator\_results\}} are replaced with the model's query and the tool output; and \texttt{\{time\_without\_tool\}} and \texttt{\{time\_with\_tool\}} are replaced with the completion times provided in Section~6.2 (Appendix~\ref{app:exp4-provided-times}). Each request is independent and contains no system prompt. We only show the reasoning-condition prompts for the mathematics task. The no-reasoning templates are identical except that the \texttt{Reason:} field is removed, ``at the \{reasoning\_level\} reasoning level'' is replaced with ``without any reasoning'', and ``You will be reasoning at the \{reasoning\_level\} reasoning level.'' is replaced with ``You will not be allowed to reason.'' The factual question-answering templates are identical except that the decision stage asks for a \texttt{Query:} field containing a free-text search query in place of \texttt{Expressions:}, answers are requested as ``your answer'' rather than ``all roots, comma separated'', and Stage~4 refers to the search tool, its query, and its results.

\subsubsection{Decision prompts (Section~6.1)}
\label{app:exp4-prompts-decision}
Stages~1 and~2 elicit confidence, Stage~3 elicits the decision, and Stage~4 elicits the final answer after a tool query. In the no stated delay baseline, the sentence ``The tool takes \{tool\_delay\_seconds\} seconds to finish.'' is removed from Stages~2 and~3; the prompts are otherwise identical.

\begin{lstlisting}[
    basicstyle=\ttfamily\footnotesize, breaklines=true, columns=fullflexible,
    frame=single, title={Stage 1: confidence without the tool.}, label={lst:exp3-no-tool-confidence}
]
Question:
{question}

No tool is available to you yet.
Do not solve or answer the question yet.

Estimate the probability from 0 to 1 that you would answer this question correctly WITHOUT the help of any tools, at the {reasoning_level} reasoning level. Do not solve the question in order to provide the estimate.

Respond in this exact format, do not output anything else:
Estimate: your estimate as a single number from 0 to 1
Reason: briefly provide your reasoning
\end{lstlisting}

\begin{lstlisting}[
    basicstyle=\ttfamily\footnotesize, breaklines=true, columns=fullflexible,
    frame=single, title={Stage 2: confidence with the tool.}, label={lst:exp3-with-tool-confidence}
]
You now have access to one query for the following tool.

{tool_description}

The tool takes {tool_delay_seconds} seconds to finish.
Do not solve or answer the question yet.

Estimate the probability from 0 to 1 that you would answer this question correctly WITH the help of this tool, at the {reasoning_level} reasoning level. Do not solve the question in order to provide the estimate.

Respond in this exact format, do not output anything else:
Estimate: your estimate as a single number from 0 to 1
Reason: briefly provide your reasoning
\end{lstlisting}
\newpage
\begin{lstlisting}[
    basicstyle=\ttfamily\footnotesize, breaklines=true, columns=fullflexible,
    frame=single, title={Stage 3: decision.}, label={lst:exp3-action-choice}
]
You need to answer the following question.

Question:
{question}

You have access to one query for the following tool:

{tool_description}

The tool takes {tool_delay_seconds} seconds to finish.

You must choose exactly one of two mutually exclusive actions:
1. Query the tool.
2. Answer directly without querying the tool.

You will only be allowed to query the tool once.

You will be reasoning at the {reasoning_level} reasoning level. Once you begin your response, do not revise, replace, or append another action.

If you choose to query the tool, output exactly one 'Expressions:' field. Do not output an 'Answer:' field. Respond in this exact format, do not output anything else:
Expressions: [expression 1, expression 2, expression 3, ...]
Reason: briefly provide your reasoning for why you chose to query the tool

If you choose to answer directly, output exactly one 'Answer:' field. Do not output an 'Expressions:' field. Respond in this exact format, do not output anything else:
Answer: all roots, comma separated
Reason: briefly provide your reasoning for why you chose not to query the tool
\end{lstlisting}

\begin{lstlisting}[
    basicstyle=\ttfamily\footnotesize, breaklines=true, columns=fullflexible,
    frame=single, title={Stage 4: answer after a tool query.}, label={lst:exp3-final-after-calculation}
]
You chose to query the calculator tool.

Your query was:
{calculator_query}

Tool results:
{calculator_results}

You have used up your one query for the calculator tool. Next, provide your final answer.

Respond in this exact format, do not output anything else:
Final Answer: all roots, comma separated
Reason: briefly provide your reasoning for your final answer
\end{lstlisting}

\subsubsection{Time-estimate prompts (Section~6.1)}
\label{app:exp4-prompts-time}
To measure how long models expect each action to take, we ask each model to estimate its total response time, first without and then with the tool, in one conversation at each stated time cost $\Tilde{t}$ ($n=5$ per task and $\Tilde{t}$; 12{,}600 trials). The no-reasoning templates also drop ``the time taken for your reasoning'' from the list of what to count. The corresponding time-estimation results are reported
in Tables~\ref{tab:time-estimates-no-tool}
and~\ref{tab:time-estimates-with-tool}.
\newpage
\begin{lstlisting}[
    basicstyle=\ttfamily\footnotesize, breaklines=true, columns=fullflexible,
    frame=single, title={Time estimate without the tool.}, label={lst:exp3-no-tool-time}
]
Question:
{question}

No tool is available to you yet.
Do not solve or answer the question yet.

Estimate how many seconds it would take you to answer this question WITHOUT the help of any tools, at the {reasoning_level} reasoning level. Count the total time from when the user sends the question until they receive your final answer, including the time taken for your reasoning, the time taken to generate your response, and the latency from the user to the API and back. Do not solve the question in order to provide the estimate.

Respond in this exact format, do not output anything else:
Time: your estimate as a single number in seconds
Reason: briefly provide your reasoning
\end{lstlisting}

\begin{lstlisting}[
    basicstyle=\ttfamily\footnotesize, breaklines=true, columns=fullflexible,
    frame=single, title={Time estimate with the tool.}, label={lst:exp3-with-tool-time}
]
You now have access to one query for the following tool.

{tool_description}

The tool takes {tool_delay_seconds} seconds to finish.
Do not solve or answer the question yet.

Estimate how many seconds it would take you to answer this question WITH the help of this tool, using your one query, at the {reasoning_level} reasoning level. Count the total time from when the user sends the question until they receive your final answer, including the time taken for your reasoning, the time taken to generate your response, the latency from the user to the API and back, and the time the tool takes to run. Do not solve the question in order to provide the estimate.

Respond in this exact format, do not output anything else:
Time: your estimate as a single number in seconds
Reason: briefly provide your reasoning
\end{lstlisting}

\subsubsection{Decision prompt with provided completion times (Section~6.2)}
\label{app:exp4-prompts-provided}
Section~6.2 uses the Stage~3 decision prompt with the block below inserted after the sentence on the reasoning level. \texttt{\{time\_without\_tool\}} is the provided no-tool time rounded to whole seconds (at least 1), and \texttt{\{time\_with\_tool\}} is the provided tool-route time plus $\Tilde{t}$, rounded. The underlying completion times are summarized in Table~\ref{tab:provided-times}. In the no-reasoning template, the first line reads ``where you were also not allowed to reason''. Trials end once the model decides: the tool is not run and no final answer is requested.

\begin{lstlisting}[
    basicstyle=\ttfamily\footnotesize, breaklines=true, columns=fullflexible,
    frame=single, title={Provided completion times (inserted into Stage 3).}, label={lst:exp3-provided-times}
]
From your other attempts at this question, where you were also at the {reasoning_level} reasoning level, we know that:
- Answering without the tool takes you about {time_without_tool} seconds in total.
- Answering with the tool takes you about {time_with_tool} seconds in total, which includes the {tool_delay_seconds} seconds the tool takes to finish.
\end{lstlisting}

\subsubsection{Forced-run prompts (Section~6.2)}
\label{app:exp4-prompts-forced}
Where a model never took an action cleanly on a question in Section~6.1, we time that action with one forced run (Appendix~\ref{app:exp4-provided-times}; Table~\ref{tab:provided-times}). Forced runs use the Stage~3 prompt with the choice removed and the stated time cost left out, so that only the model's own time is recorded. The two numbered actions are replaced with ``For this question, answer directly without querying the tool.'' or ``For this question, query the tool before giving your final answer.'', and only the corresponding output format is kept. A forced tool run then continues with Stage~4.

\subsubsection{Tool descriptions}
\label{app:exp4-tools}

\begin{lstlisting}[basicstyle=\ttfamily\footnotesize, breaklines=true, columns=fullflexible, frame=single, title={Calculator Tool description.}]
CALCULATOR
Evaluates fully numerical mathematical expressions.
It cannot solve equations or perform symbolic algebra.
Real and complex arithmetic are supported.
Each expression is evaluated independently and returns a numerical value.
Returns an array of decimal values in the same order as the given expressions.

A list of self-contained mathematical expressions to evaluate.
Each expression is evaluated independently and returns a single number.

Supported operations include:
    - +, -, *, /, and ^ or ** for powers
    - parentheses
    - sqrt(), cbrt(), abs(), log(), exp()
    - sin(), cos(), tan(), asin(), acos(), atan()
    - scientific notation such as 1e10 or 2.5e-3
    - the constants pi and E, where E is Euler's number
    - complex arithmetic using I as the imaginary unit

Not supported; these return an error rather than a value:
    - equation solving. There is no solve(). Send an expression to evaluate, not an equation to solve.
    - "=" anywhere, including assignments such as "p = b^2 - 3*c".
    - variables. Expressions cannot contain free variables.

Complex results are returned, written like "1.0 + 1.73205*I".

Undefined results such as 1/0 are reported as errors.

Example:
    ["125 / 8", "sqrt(72)", "3^4 - 10", "2*sqrt(19/3)*cos(acos(-0.5)/3)"]
\end{lstlisting}

\begin{lstlisting}[basicstyle=\ttfamily\footnotesize, breaklines=true, columns=fullflexible, frame=single, title={Search Tool description.}]
WEB SEARCH
Searches the web for up-to-date information relevant to the query. Returns a set of search results with titles, URLs, snippets, and source metadata.
\end{lstlisting}

\subsection{Tasks}
\label{app:exp4-tasks}

\subsubsection{Task Generation}
\label{app:exp3_task_generation}
The 30 cubic equations were generated with a wrapper around DeepMind's Mathematics Dataset \citep{saxton2019analysing} that restricts its \texttt{algebra\_\_polynomial\_roots} module to degree-3, solve-for-the-variable problems; upstream samples degrees 2--5 and asks for a factorization about half the time. The three difficulty levels are the upstream entropy schedule, which increases the magnitude of the coefficients. Every equation is built from rational roots and factors exactly, and ten were drawn at each difficulty level.

\subsubsection{Example Tasks}
\label{app:exp3_example_tasks}
\begin{lstlisting}[basicstyle=\ttfamily\footnotesize, breaklines=true, columns=fullflexible, frame=single, title={Example tasks.}]
EASY:   Suppose 2*q**3 - 18*q**2 - 20*q + 336 = 0. What is q?
MEDIUM:   Let -5*b**3 - 125*b**2 + 440*b + 560 = 0. What is b?
HARD:   What is v in -4*v**3 + 1099208*v**2 - 75512841472*v - 453116621184 = 0?

FRESHQA: What's the current Tomatometer score of the most recent season of The Lord of the Rings: The Rings of Power on Rotten Tomatoes?
FRESHQA: Who is the latest winner of The Voice US?
FRESHQA: Who won the most recent Super Bowl?
\end{lstlisting}

\subsection{Data Exclusions}
\label{app:exp4-exclusions}

\textbf{Section~6.1.} We exclude 289 of 28{,}799 trials whose response visibly exposes problem-solving or reasoning (mathematical work, verification, rejection of an answer candidate, self-evaluation, or reasoning about the required format) anywhere other than the requested \texttt{Reason:} field, including inside an answer or within a tool query. Because such text can serve as extra computational space the prompt protocol did not provide other trials, these trials are treated as protocol-noncompliant. The exclusion is concentrated in Claude Opus 5 without reasoning on cubic equations, which includes 221 of the 289.

\textbf{Section~6.2.} We apply the same rule and exclude 127 of 25{,}200 trials (0.5\%) whose response visibly exposes reasoning, all on cubics: 61 by Claude Opus 5 without reasoning and 42 and 24 by Gemini 3.7 Flash with and without reasoning.

\subsection{Additional Results for Section~6.1}
\label{app:exp4-estimated-results}

\subsubsection{Results Summary}
\label{app:exp3_summary}
Table~\ref{tab:app-summary} summarizes each condition. \emph{Tool} and \emph{Acc.} are unweighted means over the 30 task-level rates, with accuracy under the model's own chosen action, and $N$ is the number of trials kept after exclusions. $\Delta p$ is the median perceived benefit in percentage points. For $\rho$, we compare each task's median reported $\Delta p$ with its tool-use rate across all $\Tilde{t}$ and compute the rank correlation across the 30 tasks; $\rho$ is undefined when either quantity does not vary across tasks. \emph{pp} and \emph{OR} give the within-task change in tool use per $10\times$ increase in $\Tilde{t}$, from linear and logistic models fitted to individual trials, relating each trial's choice (tool or no tool) to $\log_{10}\Tilde{t}$ with a separate baseline rate of tool use for each task; their 95\% intervals are computed over tasks rather than trials. Because both models estimate the slope within each task, a task on which the model always or never used the tool provides no evidence of sensitivity to $\Tilde{t}$. \emph{Var.} therefore counts the tasks on which the model both used and declined the tool. The logistic fit is identified only from these tasks, so OR is unstable when \emph{Var.} is small. Table~\ref{tab:app-baseline} reports the same quantities for the baseline, in which no time cost is stated.
\begin{table}[!htbp]
\centering
\caption{\textbf{Results summary.} Tool use, accuracy, perceived benefit, and within-task sensitivity to $\Tilde{t}$ for each condition. Quantities are defined in Appendix~\ref{app:exp3_summary}.}
\vspace{7pt}
\label{tab:app-summary}
\small
\setlength{\tabcolsep}{3pt}
\begin{tabular}{@{}llrrrrr r@{\;}l r@{\;}l r@{}}
\toprule
Model & Reasoning & $N$ & Tool & Acc. & $\Delta p$ & $\rho$ & \multicolumn{2}{c}{pp [95\% CI]} & \multicolumn{2}{c}{OR [95\% CI]} & Var. \\
\midrule
\multicolumn{12}{@{}l}{\emph{Cubic equations}} \\
Claude Opus 5    & Default & 2,100 & 0.0\%  & 100.0\% & 3  & --   & --      &                   & --     &              & 0/30 \\
                 & None & 1,912 & 72.5\% & 42.4\%  & 10 & 0.79 & $-7.1$  & [$-10.8$, $-3.4$] & $0.26$ & [0.14, 0.48] & 18/30 \\
GPT-5.6 Sol      & Default & 2,100 & 13.4\% & 99.7\%  & 7  & 0.86 & $+0.1$  & [$-1.0$, $+1.2$]  & $1.02$ & [0.88, 1.18] & 16/30 \\
                 & None & 2,100 & 53.0\% & 12.7\%  & 19 & 0.91 & $+0.5$  & [$-1.0$, $+2.1$]  & $1.08$ & [0.85, 1.37] & 13/30 \\
Gemini 3.7 Flash & Default & 2,068 & 26.1\% & 97.4\%  & 18 & 0.81 & $-3.1$  & [$-5.2$, $-0.9$]  & $0.56$ & [0.42, 0.74] & 13/30 \\
                 & None & 2,076 & 38.4\% & 87.8\%  & 15 & 0.83 & $-4.2$  & [$-6.3$, $-2.0$]  & $0.55$ & [0.43, 0.69] & 15/30 \\
\midrule
\multicolumn{12}{@{}l}{\emph{FreshQA search}} \\
Claude Opus 5    & Default & 2,100 & 94.9\% & 71.0\%  & 38 & 0.58 & $-2.4$  & [$-5.2$, $+0.5$]  & $0.05$ & [0.03, 0.09] & 4/30 \\
                 & None & 2,098 & 98.1\% & 66.6\%  & 52 & 0.31 & $-1.3$  & [$-3.9$, $+1.3$]  & --     &              & 1/30 \\
GPT-5.6 Sol      & Default & 2,100 & 94.8\% & 62.6\%  & 40 & 0.68 & $0.0$   & [$-0.6$, $+0.5$]  & $0.98$ & [0.74, 1.30] & 6/30 \\
                 & None & 2,100 & 99.5\% & 64.1\%  & 44 & 0.50 & $0.0$   & [$-0.1$, $+0.1$]  & $1.01$ & [0.85, 1.19] & 3/30 \\
Gemini 3.7 Flash & Default & 2,100 & 88.0\% & 56.0\%  & 30 & 0.76 & $-10.5$ & [$-15.3$, $-5.8$] & $0.06$ & [0.02, 0.13] & 15/30 \\
                 & None & 2,099 & 86.2\% & 44.0\%  & 29 & 0.71 & $-10.0$ & [$-14.5$, $-5.5$] & $0.11$ & [0.05, 0.21] & 20/30 \\
\bottomrule
\end{tabular}
\end{table}

\begin{table}[!htbp]
\centering
\caption{\textbf{No-delay baseline summary.} Same quantities as Table~\ref{tab:app-summary}, for trials in which no time cost is stated.}
\vspace{7pt}
\label{tab:app-baseline}
\small
\setlength{\tabcolsep}{3.5pt}
\begin{tabular}{@{}llrrrrr@{}}
\toprule
Model & Reasoning & $N$ & Tool & Acc. & $\Delta p$ & $\rho$ \\
\midrule
\multicolumn{7}{@{}l}{\emph{Cubic equations}} \\
Claude Opus 5    & Default  & 300 & 0.0\%   & 100.0\% & 4  & --   \\
                 & None  & 267 & 81.4\%  & 40.2\%  & 13 & 0.59 \\
GPT-5.6 Sol      & Default  & 300 & 15.3\%  & 99.7\%  & 7  & 0.61 \\
                 & None  & 300 & 61.0\%  & 10.3\%  & 18 & 0.89 \\
Gemini 3.7 Flash & Default  & 295 & 32.5\%  & 98.6\%  & 18 & 0.80 \\
                 & None  & 295 & 46.6\%  & 83.4\%  & 15 & 0.70 \\
\midrule
\multicolumn{7}{@{}l}{\emph{FreshQA search}} \\
Claude Opus 5    & Default  & 300 & 96.7\%  & 76.0\%  & 36 & 0.29 \\
                 & None  & 300 & 100.0\% & 66.3\%  & 53 & --   \\
GPT-5.6 Sol      & Default  & 300 & 97.7\%  & 58.0\%  & 40 & 0.38 \\
                 & None  & 300 & 100.0\% & 61.0\%  & 49 & --   \\
Gemini 3.7 Flash & Default  & 300 & 99.3\%  & 41.3\%  & 30 & 0.37 \\
                 & None  & 300 & 97.0\%  & 39.0\%  & 30 & 0.43 \\
\bottomrule
\end{tabular}
\end{table}
\subsubsection{Tool Use, Accuracy, and Confidence}
\label{app:exp3_accuracy}

\begin{figure}[!htbp]
\centering
\includegraphics[width=.97\textwidth]{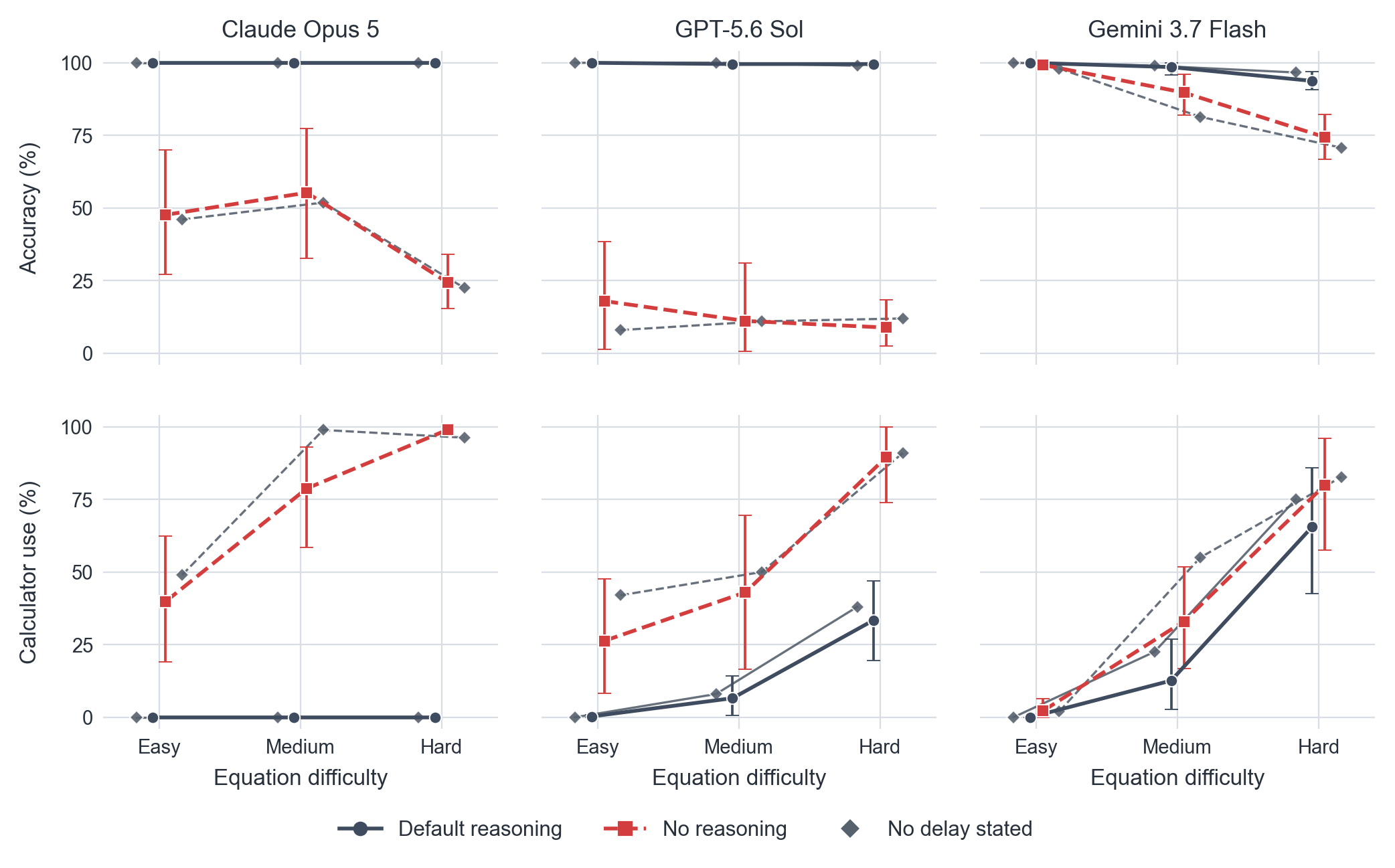}
\caption{\textbf{Accuracy and calculator use by equation difficulty.} Each point averages the 10 tasks at that level, pooled over all seven $\Tilde{t}$; bars are 95\% bootstrap intervals over tasks. Gray diamonds beside each series mark the no-delay baseline.}
\label{fig:app-difficulty}
\end{figure}

Figure~\ref{fig:app-difficulty} reports the accuracy and calculator use in the three cubic difficulty levels, along with the no-delay baseline. Across models, overall accuracy ranges from 97.4\% to 100\% with reasoning and from 12.7\% to 87.8\% without reasoning (these averages pool all three difficulty levels). Claude with reasoning answers all trials correctly without ever querying the calculator. In the other conditions, calculator use increases with equation difficulty. This does not by itself show an accuracy benefit, since harder equations are both solved with the calculator more often and answered incorrectly more often. The no-delay baseline shows the same pattern, where accuracy ranges from 98.6\% to 100\% with reasoning and from 10.3\% to 83.4\% without, and calculator use is higher on harder questions.

Tables~\ref{tab:app-toolbydelay}, \ref{tab:app-bydifficulty}, and~\ref{tab:app-confidence} report tool use by stated time cost, calculator use and accuracy by equation difficulty, and the elicited confidence estimates.
\input{Appendices/exp3_descriptive}

\subsubsection{Perceived Benefit and Tool Choice}
\label{app:benefit-choice}
\begin{figure}[!htbp]
\centering
\includegraphics[width=.9\textwidth]{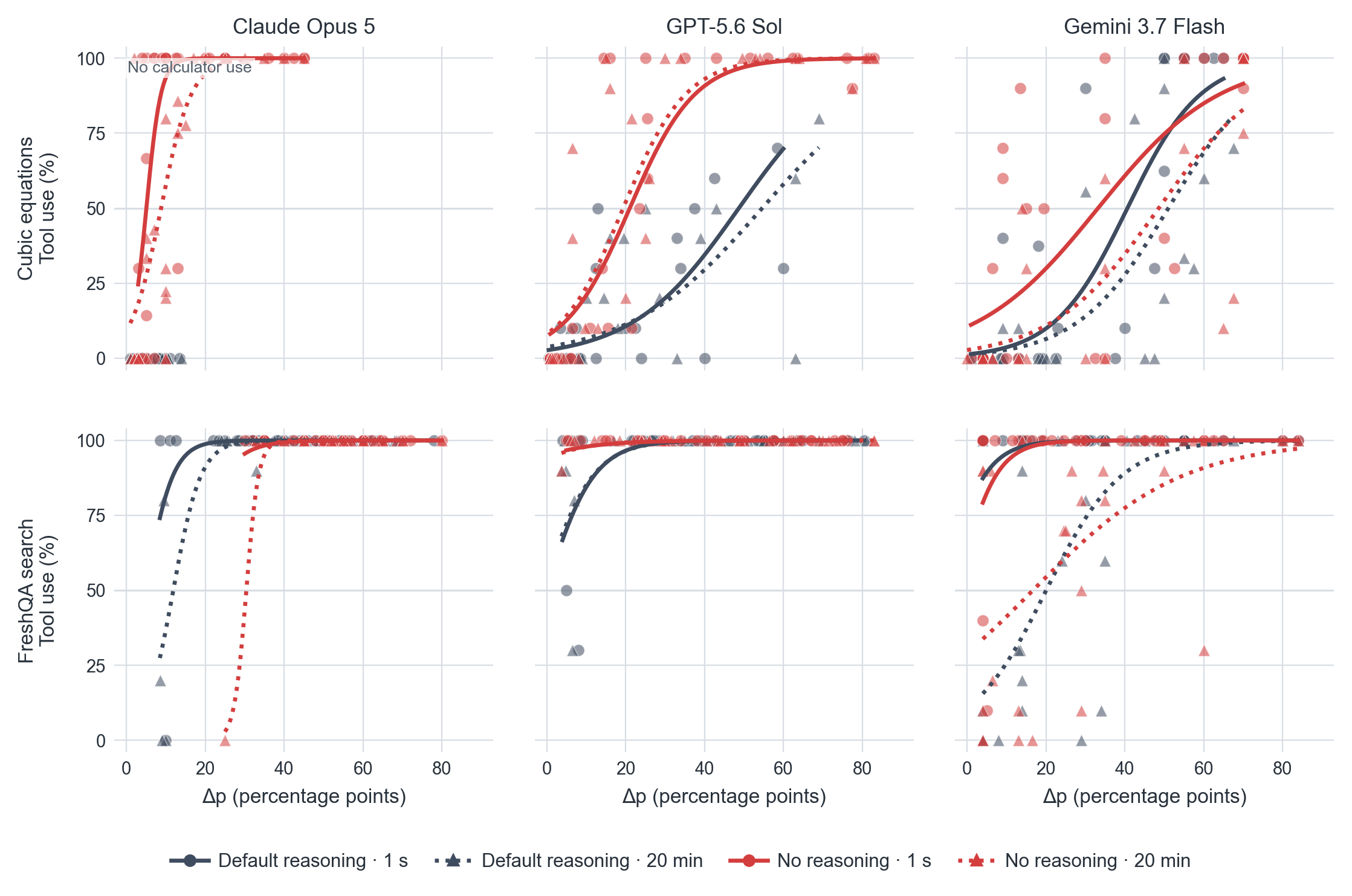}
\caption{\textbf{Tool use increases with the perceived accuracy benefit from using the tool, demonstrated by the positive relationship between the two axes.} Points are task--$\Tilde{t}$ cells, with median reported benefit $\Delta p$ on the horizontal axis and tool-use rate on the vertical axis. Color denotes reasoning condition; marker and line style denote $\Tilde{t}$. Curves are descriptive logistic fits, each drawn over the $\Delta p$ range observed at that $\Tilde{t}$. For simplicity, only the shortest and longest delay values ($\Tilde{t}$) are shown.}
\label{fig:benefit-choice}
\end{figure}

The curves in Figure~\ref{fig:benefit-choice} come from a descriptive fit, estimated separately for each condition over all seven $\Tilde{t}$: a binomial GLM on task--delay cell rates, weighted by retained trials,
\begin{equation}
\operatorname{logit}\pi_c = \beta_0 + \beta_1 z^{p}_c + \beta_2 z^{t}_c + \beta_3\, z^{p}_c z^{t}_c ,
\end{equation}
where $\pi_c$ is the probability of tool use in cell $c$, $z^{p}_c$ is the standardized median $\Delta p$ of the cell, and $z^{t}_c$ is the standardized $\log_{10}\Tilde{t}$. The elicited confidence estimates $\Delta p$ are summarized in Table~\ref{tab:app-confidence}. The fit has no task fixed effects. Each curve spans only the $\Delta p$ range observed at that $\Tilde{t}$, so curves should be compared where those ranges overlap. Several search panels sit near ceiling with few tasks varying (Table~\ref{tab:app-summary}), and their curves should be read accordingly. A rightward shift from 1~s to 20~min is consistent with a larger perceived accuracy gain being needed to offset the longer stated $\Tilde{t}$.

\subsubsection{Model-Call Durations}
\label{app:reasoning-duration}
Table~\ref{tab:reasoning-duration} reports median differences in model-call duration between matched tool-use and no-tool trials. Tool-use trials take longer in all five estimable comparisons, by 1.3 to 6.7~s. This is consistent with API overhead contributing to the differences, but each call also includes model computation and response generation. These measurements therefore cannot isolate internal reasoning time or establish that reasoning durations are equivalent.
\begin{table}[!htbp]
\centering
\small
\caption{\textbf{Model-call durations.}
All results use default reasoning. For each task and stated time cost where both tool and no-tool trials occur, we subtract the mean no-tool duration from the mean tool duration. Then, the differences are averaged equally across matched delays within each task, and the median across matched tasks is reported. Duration is the sum of recorded model-call latencies in the task-answering conversation, excluding confidence estimation and tool execution. Positive values indicate longer model-call durations with tool use.}
\vspace{6pt}
\label{tab:reasoning-duration}
\begin{tabular}{llrr}
\toprule
Domain & Model & Matched tasks & Median difference (s) \\
\midrule
Cubics & Claude Opus 5    &  0 & --      \\
       & GPT-5.6 Sol      & 16 & $+6.69$ \\
       & Gemini 3.7 Flash & 13 & $+2.77$ \\
\midrule
Search & Claude Opus 5    &  4 & $+2.06$ \\
       & GPT-5.6 Sol      &  6 & $+2.46$ \\
       & Gemini 3.7 Flash & 14 & $+1.29$ \\
\bottomrule
\end{tabular}
\end{table}

\subsubsection{Time Estimates}
\label{app:exp4-time-estimates}
Tables~\ref{tab:time-estimates-no-tool} and~\ref{tab:time-estimates-with-tool} compare the time estimates with the times recorded in Section~6.1 (Appendix~\ref{app:exp4-provided-times}). Models overestimate their no-tool time, most on cubics, while their tool-route estimates net of $\Tilde{t}$ are closer to the recorded times. Their with-tool estimates therefore fall below their no-tool estimates at short delays, although the tool route is recorded as slower for nearly every cubic. This belief disappears once $\Tilde{t}$ exceeds the expected saving (Table~\ref{tab:believed-faster}).

\begin{table}[!htbp]
\centering
\small
\caption{\textbf{Estimated and recorded response time without the tool.} \emph{Estimated} is the median of the model's estimates of its total response time without the tool; \emph{Recorded} is the median over questions of the model's recorded time answering directly in Section~6.1 (Appendix~\ref{app:exp4-provided-times}). \emph{Ratio} is the median over questions of estimated divided by recorded time, with the 10th--90th percentile range; values above one indicate overestimation.}
\vspace{8pt}
\label{tab:time-estimates-no-tool}
\begin{tabular}{llrrl}
\toprule
Model & Reasoning & Estimated (s) & Recorded (s) & Ratio ($\times$) \\
\midrule
\multicolumn{5}{l}{\emph{Cubic equations}} \\
Claude Opus 5 & Default & $100.0$ & $8.2$ & $12.0$ [$7.7$--$17.3$] \\
 & None & $120.0$ & $2.6$ & $43.3$ [$15.3$--$102.6$] \\
GPT-5.6 Sol & Default & $25.0$ & $6.1$ & $4.8$ [$2.7$--$7.9$] \\
 & None & $25.0$ & $2.0$ & $12.2$ [$4.7$--$21.6$] \\
Gemini 3.7 Flash & Default & $18.0$ & $4.1$ & $4.1$ [$2.3$--$6.7$] \\
 & None & $4.5$ & $2.9$ & $1.5$ [$1.1$--$3.2$] \\
\midrule
\multicolumn{5}{l}{\emph{FreshQA search}} \\
Claude Opus 5 & Default & $15.0$ & $7.5$ & $1.7$ [$0.9$--$3.6$] \\
 & None & $5.0$ & $3.1$ & $1.8$ [$1.2$--$3.4$] \\
GPT-5.6 Sol & Default & $8.0$ & $4.3$ & $2.1$ [$0.7$--$4.3$] \\
 & None & $3.0$ & $2.2$ & $1.4$ [$0.9$--$5.0$] \\
Gemini 3.7 Flash & Default & $4.0$ & $2.3$ & $2.0$ [$1.2$--$2.6$] \\
 & None & $1.2$ & $1.9$ & $0.7$ [$0.4$--$1.0$] \\
\bottomrule
\end{tabular}
\end{table}

\begin{table}[!htbp]
\centering
\small
\caption{\textbf{Estimated and recorded response time with the tool.} \emph{Estimated} is the median of the model's with-tool estimate minus the stated time cost $\Tilde{t}$, i.e., its own time on the tool route; \emph{Recorded} is the median over questions of the recorded model-call time on that route, excluding the tool's running time. \emph{Ratio} is the median over questions of estimated divided by recorded time.}
\vspace{6pt}
\label{tab:time-estimates-with-tool}
\begin{tabular}{llrrr}
\toprule
Model & Reasoning & Estimated (s) & Recorded (s) & Ratio ($\times$) \\
\midrule
\multicolumn{5}{l}{\emph{Cubic equations}} \\
Claude Opus 5 & Default & $85.0$ & $12.8$ & $6.2$ \\
 & None & $79.0$ & $8.1$ & $9.4$ \\
GPT-5.6 Sol & Default & $15.0$ & $13.8$ & $1.1$ \\
 & None & $7.0$ & $5.9$ & $1.1$ \\
Gemini 3.7 Flash & Default & $10.0$ & $7.2$ & $1.5$ \\
 & None & $3.0$ & $4.9$ & $0.7$ \\
\midrule
\multicolumn{5}{l}{\emph{FreshQA search}} \\
Claude Opus 5 & Default & $15.0$ & $9.7$ & $1.5$ \\
 & None & $5.5$ & $5.3$ & $1.0$ \\
GPT-5.6 Sol & Default & $8.0$ & $4.8$ & $1.7$ \\
 & None & $3.0$ & $3.0$ & $1.1$ \\
Gemini 3.7 Flash & Default & $5.0$ & $2.6$ & $2.1$ \\
 & None & $1.5$ & $3.2$ & $0.5$ \\
\bottomrule
\end{tabular}
\end{table}

\begin{table}[!htbp]
\centering
\small
\caption{\textbf{Tool route estimated to be faster, by stated time cost.} Share of estimates (\%) in which the model's estimated total time with the tool is below its estimated time without it, by $\Tilde{t}$. The last column gives the share of questions (\%) on which the recorded tool route plus 1\,s is in fact faster than answering directly.}
\vspace{6pt}
\label{tab:believed-faster}
\begin{tabular}{llrrrrrrrr}
\toprule
 & & \multicolumn{7}{c}{Believed faster, by $\Tilde{t}$} & Actual \\
\cmidrule(lr){3-9}
Model & Reasoning & 1\,s & 5\,s & 10\,s & 20\,s & 30\,s & 3\,min & 20\,min & at 1\,s \\
\midrule
\multicolumn{10}{l}{\emph{Cubic equations}} \\
Claude Opus 5 & Default & $89$ & $67$ & $61$ & $45$ & $31$ & $2$ & $0$ & $0$ \\
 & None & $99$ & $92$ & $71$ & $57$ & $25$ & $0$ & $0$ & $0$ \\
GPT-5.6 Sol & Default & $97$ & $83$ & $66$ & $40$ & $23$ & $0$ & $0$ & $3$ \\
 & None & $97$ & $85$ & $78$ & $45$ & $30$ & $0$ & $0$ & $0$ \\
Gemini 3.7 Flash & Default & $86$ & $70$ & $44$ & $9$ & $0$ & $0$ & $0$ & $3$ \\
 & None & $59$ & $16$ & $5$ & $1$ & $0$ & $0$ & $0$ & $3$ \\
\midrule
\multicolumn{10}{l}{\emph{FreshQA search}} \\
Claude Opus 5 & Default & $47$ & $9$ & $2$ & $0$ & $0$ & $0$ & $0$ & $27$ \\
 & None & $9$ & $0$ & $0$ & $0$ & $0$ & $0$ & $0$ & $3$ \\
GPT-5.6 Sol & Default & $73$ & $21$ & $13$ & $0$ & $1$ & $0$ & $0$ & $30$ \\
 & None & $23$ & $6$ & $2$ & $0$ & $0$ & $0$ & $0$ & $0$ \\
Gemini 3.7 Flash & Default & $9$ & $1$ & $0$ & $0$ & $0$ & $0$ & $0$ & $13$ \\
 & None & $0$ & $0$ & $0$ & $0$ & $0$ & $0$ & $0$ & $0$ \\
\bottomrule
\end{tabular}
\end{table}

\subsection{Additional Results for Section~6.2}
\label{app:exp4-provided-results}

\subsubsection{Provided Completion Times}
\label{app:exp4-provided-times}
For each model, reasoning setting, question and action, the provided time is the median recorded duration of the model's calls in clean Section~6.1 trials that took that action, pooled over $\Tilde{t}$. It counts the decision call and, for the tool route, the final-answer call, but not the tool's own running time; the stated $\Tilde{t}$ is added instead, so the tool takes exactly as long as stated. Where a model never took an action cleanly on a question, we record one forced run of that action (Appendix~\ref{app:exp4-prompts-forced}). Table~\ref{tab:provided-times} summarizes the provided times.

\begin{table}[!htbp]
\centering
\small
\caption{\textbf{Completion times provided in Section~6.2.} \emph{Sec.~6.1} and \emph{Forced} count the 60 question--action times (30 questions, two actions) taken from Section~6.1 trials and from forced runs. \emph{No tool} and \emph{Tool} are medians over questions of the provided times, the latter before adding $\Tilde{t}$.}
\vspace{6pt}
\label{tab:provided-times}
\begin{tabular}{llrrrr}
\toprule
 & & \multicolumn{2}{c}{Source} & \multicolumn{2}{c}{Median time (s)} \\
\cmidrule(lr){3-4}\cmidrule(lr){5-6}
Model & Reasoning & Sec.~6.1 & Forced & No tool & Tool \\
\midrule
\multicolumn{6}{l}{\emph{Cubic equations}} \\
Claude Opus 5 & Default & 30/60 & 30/60 & $8.0$ & $12.8$ \\
 & None & 48/60 & 12/60 & $3.0$ & $8.1$ \\
GPT-5.6 Sol & Default & 46/60 & 14/60 & $6.0$ & $13.8$ \\
 & None & 43/60 & 17/60 & $2.0$ & $5.9$ \\
Gemini 3.7 Flash & Default & 43/60 & 17/60 & $4.0$ & $7.2$ \\
 & None & 45/60 & 15/60 & $3.0$ & $4.9$ \\
\midrule
\multicolumn{6}{l}{\emph{FreshQA search}} \\
Claude Opus 5 & Default & 34/60 & 26/60 & $7.5$ & $9.7$ \\
 & None & 31/60 & 29/60 & $3.0$ & $5.3$ \\
GPT-5.6 Sol & Default & 36/60 & 24/60 & $4.0$ & $4.8$ \\
 & None & 33/60 & 27/60 & $2.0$ & $3.0$ \\
Gemini 3.7 Flash & Default & 45/60 & 15/60 & $2.0$ & $2.6$ \\
 & None & 50/60 & 10/60 & $2.0$ & $3.2$ \\
\bottomrule
\end{tabular}
\end{table}

\subsubsection{Results Summary}
\label{app:exp4-provided-summary}
Table~\ref{tab:results-provided} gives the Section~6.2 counterpart of Table~\ref{tab:app-summary}, estimated the same way (Appendix~\ref{app:exp3_summary}). Section~6.2 trials end at the decision, so accuracy is not measured and $\Delta p$ is not elicited again; $\rho$ instead uses each task's median $\Delta p$ from Section~6.1. OR is omitted for Claude without reasoning on search, where every varying task switches from always to never using the tool at a single $\Tilde{t}$, so the logistic fit separates. Table~\ref{tab:tool-by-delay-provided} gives the tool-use rates behind Figure~\ref{fig:tool-use-delay2}, the counterpart of Table~\ref{tab:app-toolbydelay}.

\begin{table}[H]
\centering
\small
\caption{\textbf{Results summary when durations are provided (Section~6.2).} Quantities as in Table~\ref{tab:app-summary}, defined in Appendix~\ref{app:exp3_summary}.}
\vspace{6pt}
\label{tab:results-provided}
\begin{tabular}{llrrrllc}
\toprule
Model & Reasoning & $N$ & Tool & $\rho$ & pp [95\% CI] & OR [95\% CI] & Var. \\
\midrule
\multicolumn{8}{l}{\emph{Cubic equations}} \\
Claude Opus 5 & Default & 2,098 & $0.0$\% & -- & -- & -- & 0/30 \\
 & None & 2,039 & $46.4$\% & $0.83$ & $-11.5$ [$-17.1$, $-6.0$] & $0.15$ [$0.06$, $0.38$] & 13/30 \\
GPT-5.6 Sol & Default & 2,100 & $2.8$\% & $0.75$ & $-1.9$ [$-3.8$, $0.0$] & $0.37$ [$0.20$, $0.70$] & 9/30 \\
 & None & 2,100 & $9.1$\% & $0.78$ & $-5.7$ [$-8.9$, $-2.4$] & $0.32$ [$0.21$, $0.47$] & 13/30 \\
Gemini 3.7 Flash & Default & 2,058 & $11.3$\% & $0.79$ & $-6.7$ [$-10.9$, $-2.5$] & $0.15$ [$0.07$, $0.34$] & 11/30 \\
 & None & 2,075 & $26.1$\% & $0.83$ & $-9.5$ [$-13.4$, $-5.7$] & $0.26$ [$0.19$, $0.34$] & 21/30 \\
\midrule
\multicolumn{8}{l}{\emph{FreshQA search}} \\
Claude Opus 5 & Default & 2,100 & $91.5$\% & $0.61$ & $-5.3$ [$-8.9$, $-1.6$] & $0.02$ [$<0.01$, $0.14$] & 9/30 \\
 & None & 2,100 & $93.3$\% & $0.72$ & $-6.0$ [$-10.4$, $-1.6$] & -- & 7/30 \\
GPT-5.6 Sol & Default & 2,100 & $81.0$\% & $0.74$ & $-13.5$ [$-17.8$, $-9.2$] & $0.12$ [$0.07$, $0.23$] & 22/30 \\
 & None & 2,100 & $90.8$\% & $0.69$ & $-7.6$ [$-12.5$, $-2.7$] & $0.08$ [$0.03$, $0.21$] & 8/30 \\
Gemini 3.7 Flash & Default & 2,100 & $72.3$\% & $0.83$ & $-22.0$ [$-27.2$, $-16.9$] & $0.01$ [$<0.01$, $0.04$] & 23/30 \\
 & None & 2,100 & $74.2$\% & $0.81$ & $-20.4$ [$-25.3$, $-15.5$] & $0.08$ [$0.05$, $0.14$] & 26/30 \\
\bottomrule
\end{tabular}
\end{table}

\begin{table}[H]
\centering
\small
\caption{\textbf{Tool use by stated time cost when durations are provided.} Each entry is the unweighted mean of the 30 task-level rates at that $\Tilde{t}$, matching the filled points in Figure~\ref{fig:tool-use-delay2}; Table~\ref{tab:app-toolbydelay} gives the same quantities for Section~6.1.}
\vspace{6pt}
\label{tab:tool-by-delay-provided}
\begin{tabular}{llrrrrrrr}
\toprule
Model & Reasoning & 1\,s & 5\,s & 10\,s & 20\,s & 30\,s & 3\,min & 20\,min \\
\midrule
\multicolumn{9}{l}{\emph{Cubic equations}} \\
Claude Opus 5 & Default & $0.0$ & $0.0$ & $0.0$ & $0.0$ & $0.0$ & $0.0$ & $0.0$ \\
 & None & $50.7$ & $56.9$ & $53.8$ & $53.6$ & $54.3$ & $33.2$ & $20.3$ \\
GPT-5.6 Sol & Default & $6.7$ & $2.3$ & $2.3$ & $4.0$ & $3.3$ & $0.7$ & $0.0$ \\
 & None & $17.0$ & $12.7$ & $10.7$ & $13.7$ & $7.7$ & $0.7$ & $1.3$ \\
Gemini 3.7 Flash & Default & $21.2$ & $20.3$ & $13.0$ & $11.3$ & $11.9$ & $3.0$ & $2.0$ \\
 & None & $38.8$ & $31.9$ & $30.0$ & $29.5$ & $28.6$ & $14.2$ & $10.9$ \\
\midrule
\multicolumn{9}{l}{\emph{FreshQA search}} \\
Claude Opus 5 & Default & $96.3$ & $94.3$ & $93.7$ & $93.7$ & $93.7$ & $90.0$ & $78.7$ \\
 & None & $96.7$ & $96.7$ & $96.7$ & $96.7$ & $96.7$ & $93.3$ & $76.3$ \\
GPT-5.6 Sol & Default & $91.0$ & $89.0$ & $90.0$ & $87.3$ & $87.3$ & $73.3$ & $49.0$ \\
 & None & $96.3$ & $95.0$ & $98.7$ & $93.7$ & $94.0$ & $82.0$ & $75.7$ \\
Gemini 3.7 Flash & Default & $92.0$ & $86.3$ & $86.7$ & $81.7$ & $77.0$ & $57.7$ & $24.7$ \\
 & None & $91.7$ & $91.0$ & $89.3$ & $77.7$ & $77.7$ & $61.0$ & $31.0$ \\
\bottomrule
\end{tabular}
\end{table}

\subsubsection{Perceived Benefit and Tool Choice}
\label{app:exp4-provided-benefit}
\begin{figure}[H]
\centering
\includegraphics[width=.9\textwidth]{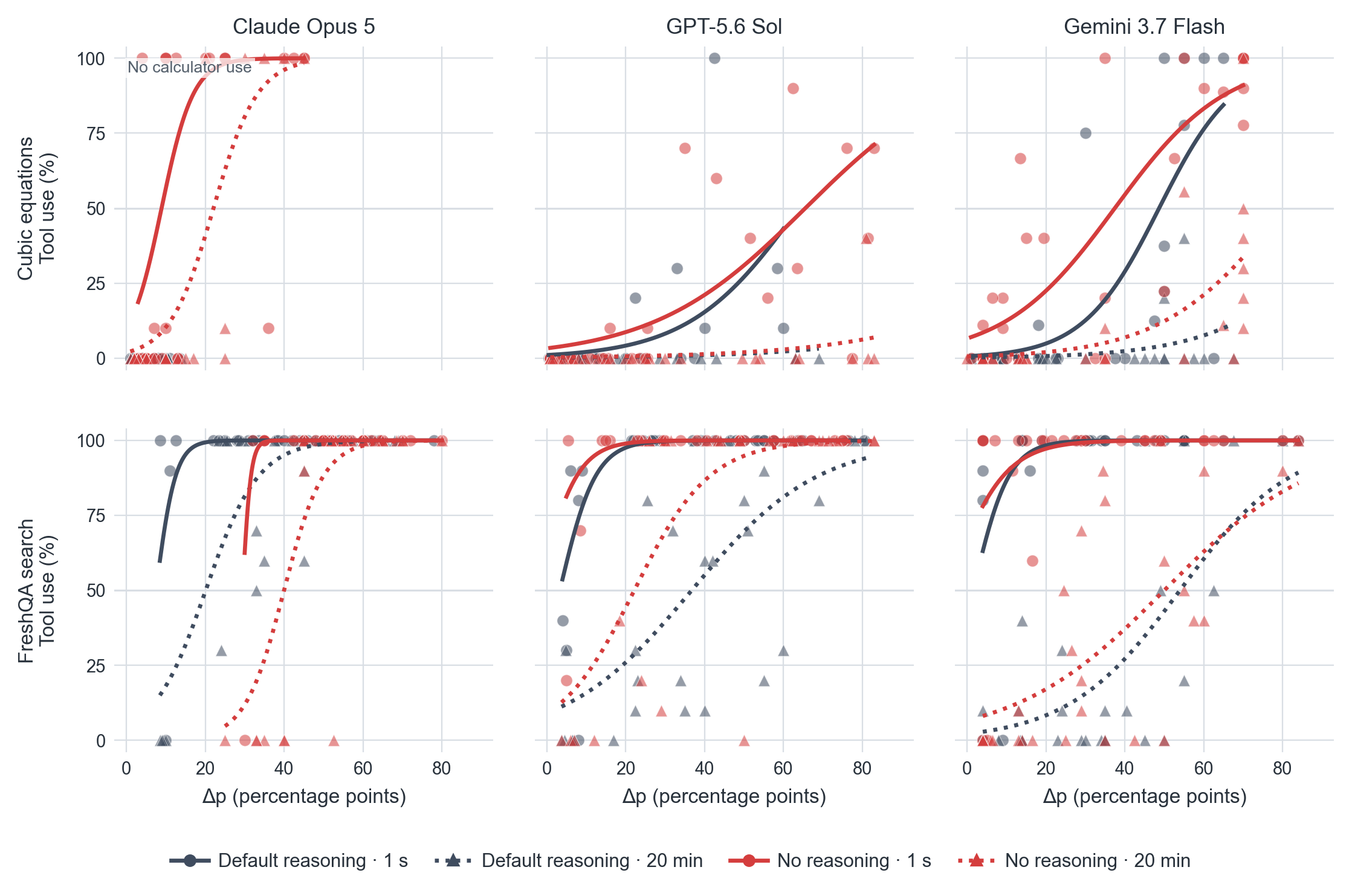}
\caption{\textbf{Perceived accuracy benefit and tool choice when durations are provided.} Same layout and fits as Figure~\ref{fig:benefit-choice}, for Section~6.2. Points are task--$\Tilde{t}$ cells, with each task's median reported benefit $\Delta p$ from Section~6.1 on the horizontal axis and its tool-use rate on the vertical axis. Color denotes reasoning condition; marker and line style denote the $\Tilde{t}$. Each curve is drawn only over the $\Delta p$ range observed at that $\Tilde{t}$.}
\label{fig:benefit-choice-provided}
\end{figure}

Figure~\ref{fig:benefit-choice-provided} repeats the descriptive fits of Figure~\ref{fig:benefit-choice} (Appendix~\ref{app:benefit-choice}) for Section~6.2. Each task's $\Delta p$ is its median from Section~6.1 at the same $\Tilde{t}$.

\subsubsection{Results by Difficulty}
\label{app:exp4-provided-difficulty}
\begin{figure}[!htbp]
\centering
\includegraphics[width=\textwidth]{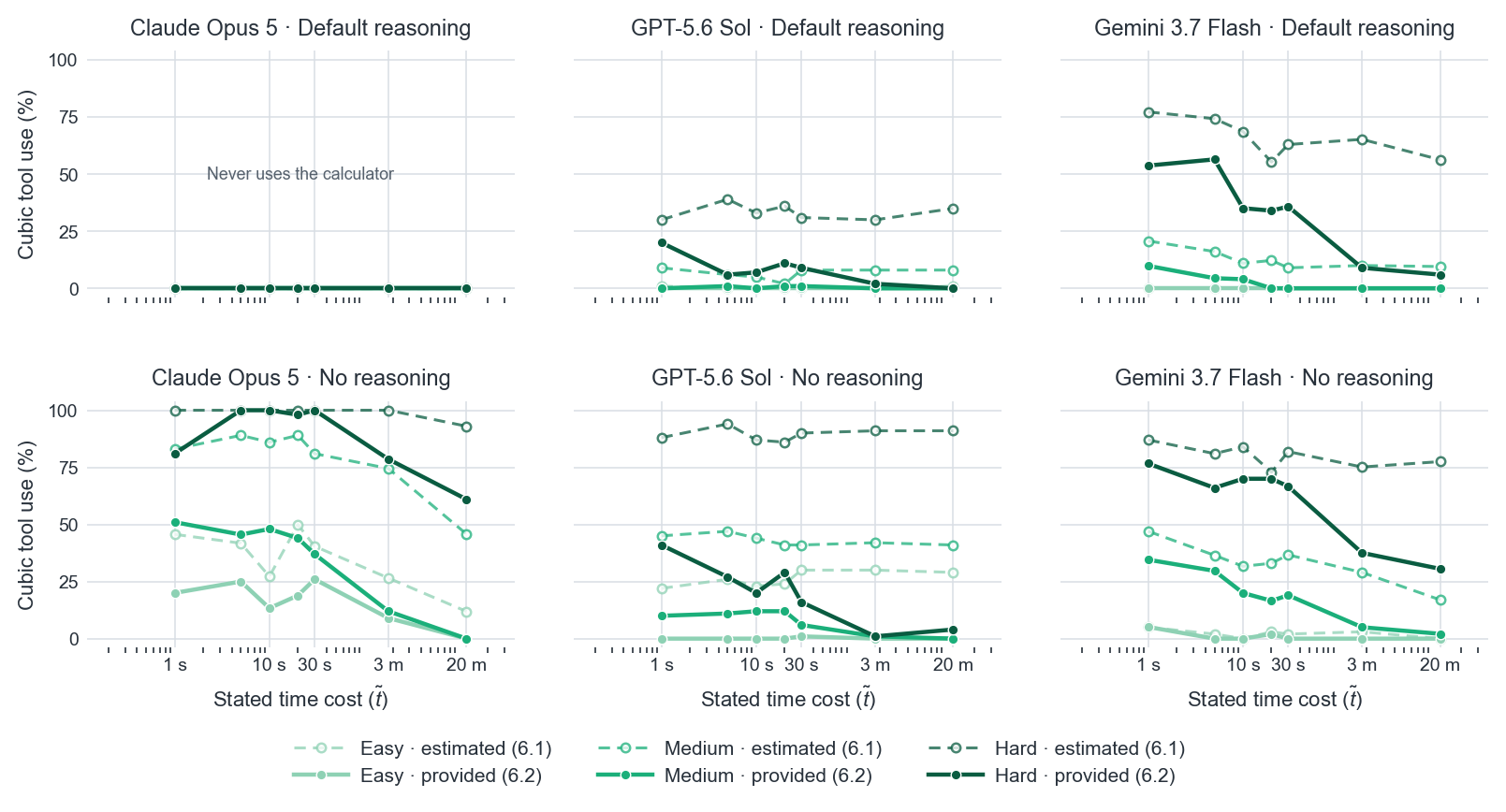}
\caption{\textbf{Cubic tool use by difficulty when durations are estimated and provided.} Points show mean tool use across the 10 cubics of each difficulty. Hollow points with dashed lines show durations estimated by the model (Section~6.1); filled points with solid lines show durations provided in the prompt (Section~6.2).}
\label{fig:difficulty-provided}
\end{figure}

Figure~\ref{fig:difficulty-provided} and Table~\ref{tab:difficulty-provided} break the cubic results down by difficulty, for both sections; Figure~\ref{fig:app-difficulty} and Table~\ref{tab:app-bydifficulty} give the corresponding Section~6.1 rates pooled over $\Tilde{t}$.

\begin{table}[!htbp]
\centering
\small
\caption{\textbf{Change in time sensitivity by equation difficulty.} Slope of tool use per $10\times$ increase in $\Tilde{t}$ (percentage points), from least-squares fits in $\log_{10}\Tilde{t}$ to the mean tool use of the ten cubics at each level, as estimated $\rightarrow$ provided (Sections~6.1 and 6.2). Dashes mark no calculator use in either section.}
\vspace{6pt}
\label{tab:difficulty-provided}
\begin{tabular}{ll r@{\;$\rightarrow$\;}l r@{\;$\rightarrow$\;}l r@{\;$\rightarrow$\;}l}
\toprule
Model & Reasoning & \multicolumn{2}{c}{Easy} & \multicolumn{2}{c}{Medium} & \multicolumn{2}{c}{Hard} \\
\midrule
Claude Opus 5 & Default & \multicolumn{2}{c}{--} & \multicolumn{2}{c}{--} & \multicolumn{2}{c}{--} \\
 & None & $-10.3$ & $-7.0$ & $-12.3$ & $-18.5$ & $-1.9$ & $-9.2$ \\
GPT-5.6 Sol & Default & $0.0$ & $0.0$ & $+0.3$ & $-0.1$ & $+0.1$ & $-5.5$ \\
 & None & $+2.6$ & $0.0$ & $-1.6$ & $-4.3$ & $+0.7$ & $-12.7$ \\
Gemini 3.7 Flash & Default & \multicolumn{2}{c}{--} & $-3.4$ & $-3.0$ & $-6.3$ & $-18.1$ \\
 & None & $-1.0$ & $-1.2$ & $-8.3$ & $-11.3$ & $-3.2$ & $-16.3$ \\
\bottomrule
\end{tabular}
\end{table}

%% file: Appendices/exp3_descriptive.tex
\begin{table}[h]
\centering
\caption{\textbf{Tool use by stated time cost.} Each entry is the unweighted mean of the 30
task-level rates at that $\tilde{t}$, matching Figure~\ref{fig:tool-use-delay1}. The first column is
the no-delay baseline.}
\vspace{7pt}
\label{tab:app-toolbydelay}
\small
\setlength{\tabcolsep}{4pt}
\begin{tabular}{llrrrrrrrr}
\toprule
Model & Reasoning & No $\tilde{t}$ & 1\,s & 5\,s & 10\,s & 20\,s & 30\,s & 3\,min & 20\,min \\
\midrule
\multicolumn{10}{l}{\emph{Cubic equations}} \\
Claude Opus 5 & Default & 0.0 & 0.0 & 0.0 & 0.0 & 0.0 & 0.0 & 0.0 & 0.0 \\
 & None & 81.4 & 77.3 & 76.9 & 71.1 & 79.6 & 73.8 & 67.0 & 50.2 \\
GPT-5.6 Sol & Default & 15.3 & 13.3 & 15.0 & 12.7 & 12.7 & 13.0 & 12.7 & 14.7 \\
 & None & 61.0 & 51.7 & 55.7 & 51.3 & 50.3 & 53.7 & 54.3 & 53.7 \\
Gemini 3.7 Flash & Default & 32.5 & 32.7 & 30.1 & 26.6 & 22.5 & 24.0 & 25.1 & 22.0 \\
 & None & 46.6 & 46.3 & 39.8 & 38.6 & 36.3 & 40.2 & 35.7 & 31.5 \\
\midrule
\multicolumn{10}{l}{\emph{FreshQA search}} \\
Claude Opus 5 & Default & 96.7 & 96.7 & 96.3 & 96.7 & 95.7 & 96.3 & 93.0 & 89.7 \\
 & None & 100.0 & 100.0 & 100.0 & 99.3 & 97.7 & 96.7 & 96.7 & 96.7 \\
GPT-5.6 Sol & Default & 97.7 & 96.0 & 94.7 & 94.7 & 94.3 & 94.0 & 93.3 & 96.3 \\
 & None & 100.0 & 100.0 & 99.7 & 99.3 & 98.7 & 99.3 & 100.0 & 99.7 \\
Gemini 3.7 Flash & Default & 99.3 & 100.0 & 95.7 & 90.3 & 91.3 & 90.7 & 81.0 & 66.7 \\
 & None & 97.0 & 95.0 & 94.3 & 93.0 & 88.0 & 88.7 & 78.7 & 65.7 \\
\bottomrule
\end{tabular}
\end{table}

\begin{table}[!htbp]
\centering
\caption{\textbf{Calculator use and accuracy by equation difficulty.} Cubic tasks only, ten
per level. Each entry is the unweighted mean of the ten task-level rates, pooled over all seven
$\tilde{t}$.}
\vspace{7pt}
\label{tab:app-bydifficulty}
\small
\begin{tabular}{llrrrrrr}
\toprule
& & \multicolumn{3}{c}{Calculator use (\%)} & \multicolumn{3}{c}{Accuracy (\%)} \\
\cmidrule(lr){3-5}\cmidrule(lr){6-8}
Model & Reasoning & Easy & Medium & Hard & Easy & Medium & Hard \\
\midrule
Claude Opus 5 & Default & 0.0 & 0.0 & 0.0 & 100.0 & 100.0 & 100.0 \\
 & None & 39.8 & 78.8 & 99.0 & 47.7 & 55.3 & 24.3 \\
GPT-5.6 Sol & Default & 0.3 & 6.6 & 33.4 & 100.0 & 99.6 & 99.6 \\
 & None & 26.3 & 43.0 & 89.6 & 18.0 & 11.1 & 8.9 \\
Gemini 3.7 Flash & Default & 0.0 & 12.6 & 65.6 & 100.0 & 98.5 & 93.8 \\
 & None & 2.1 & 33.0 & 79.9 & 99.3 & 89.7 & 74.4 \\
\bottomrule
\end{tabular}
\end{table}

\begin{table}[!htbp]
\centering
\caption{\textbf{Elicited confidence estimates.} Medians over retained trials.
$p_{\mathrm{no\text{-}tool}}$ and $p_{\mathrm{tool}}$ are the model's stated probabilities of
answering correctly without and with the tool; $\Delta p$ is their difference. Cubic entries are
broken out by difficulty.}
\vspace{7pt}
\label{tab:app-confidence}
\small
\setlength{\tabcolsep}{4pt}
\begin{tabular}{llrrrrrr}
\toprule
& & \multicolumn{3}{c}{Median $p_{\mathrm{no\text{-}tool}}$ (\%)} & \multicolumn{3}{c}{Median $\Delta p$ (pp)} \\
\cmidrule(lr){3-5}\cmidrule(lr){6-8}
Model & Reasoning & Easy & Med. & Hard & Easy & Med. & Hard \\
\midrule
\multicolumn{8}{l}{\emph{Cubic equations}} \\
Claude Opus 5 & Default & 95 & 91 & 85 & 2 & 4 & 8 \\
 & None & 85 & 75 & 35 & 5 & 10 & 35 \\
GPT-5.6 Sol & Default & 97 & 95 & 65 & 3 & 4 & 33 \\
 & None & 93 & 85 & 35 & 6 & 13 & 63 \\
Gemini 3.7 Flash & Default & 95 & 85 & 35 & 4 & 14 & 50 \\
 & None & 95 & 80 & 20 & 4 & 15 & 70 \\
\midrule
\multicolumn{2}{l}{\emph{FreshQA search}} & \multicolumn{3}{c}{(no difficulty levels)} & & & \\
Claude Opus 5 & Default & \multicolumn{3}{c}{55} & \multicolumn{3}{c}{38} \\
 & None & \multicolumn{3}{c}{35} & \multicolumn{3}{c}{52} \\
GPT-5.6 Sol & Default & \multicolumn{3}{c}{55} & \multicolumn{3}{c}{40} \\
 & None & \multicolumn{3}{c}{55} & \multicolumn{3}{c}{44} \\
Gemini 3.7 Flash & Default & \multicolumn{3}{c}{65} & \multicolumn{3}{c}{30} \\
 & None & \multicolumn{3}{c}{70} & \multicolumn{3}{c}{29} \\
\bottomrule
\end{tabular}
\end{table}